\documentclass[12pt,letterpaper]{article}
\usepackage[letterpaper]{geometry} 

\usepackage[utf8]{inputenc} 

 \usepackage{setspace} 
\usepackage{amsmath,amssymb,amsfonts}
\usepackage{enumerate} 
\usepackage{xfrac} 
\usepackage{tabularx}
\usepackage{mdwlist} 
\usepackage{diagbox} 
\usepackage{booktabs} 
\usepackage[usenames,dvipsnames,svgnames,table]{xcolor}
\usepackage[linesnumbered,lined,boxed]{algorithm2e} 
\graphicspath{{./images/}} 
\usepackage{subcaption} 
\usepackage{float} 
\usepackage{soul} 
\usepackage[table]{xcolor}
\usepackage{bm} 
\usepackage{tikz}
  \usetikzlibrary{decorations.pathreplacing}
  \usetikzlibrary{calc,patterns,angles,quotes}
  \usetikzlibrary{shapes}
\usepackage{verbatim}
\usepackage[normalem]{ulem} 
\usepackage[hidelinks]{hyperref} 
\usepackage{multirow}
\usepackage{array} 
   \newcolumntype{L}[1]{>{\raggedright\let\newline\\\arraybackslash\hspace{0pt}}m{#1}}
   \newcolumntype{C}[1]{>{\centering\let\newline\\\arraybackslash\hspace{0pt}}m{#1}}
   \newcolumntype{R}[1]{>{\raggedleft\let\newline\\\arraybackslash\hspace{0pt}}m{#1}}
\usepackage{lipsum}
\usepackage{pdflscape} 
\usepackage[toc]{appendix}
\usepackage{enumitem}

\setstcolor{red}
\newcommand{\aad}[1]{{\color{YellowOrange}{#1}}}
\newcommand{\maad}[1]{{\color{ForestGreen}{#1}}}

\title{Detecting Biological Locomotion in Video: \\ A Computational Approach}
\author{Soo Min Kang\\
\\
Richard P. Wildes \\ 
\\
Department of Electrical Engineering and Computer Science \\
Centre for Vision Research \\
York University \\
Toronto, Ontario, Canada \\
}

\begin{document}
\maketitle

\pagenumbering{roman}

\newpage

\begin{abstract}
	\thispagestyle{plain}
	\pagenumbering{roman}
	\setcounter{page}{2}
		
	Animals locomote for various reasons: to search for food, to find suitable habitat, to pursue prey, to escape from predators, or to seek a mate. The grand scale of biodiversity contributes to the great locomotory design and mode diversity. In this report, the locomotion of general biological species is referred to as \textbf{biolocomotion}. The goal of this report is to develop a computational approach to detect biolocomotion in any unprocessed video. \\

The ways biological entities locomote through an environment are extremely diverse. Various creatures make use of legs, wings, fins, and other means to move through the world. Significantly, the motion exhibited by the body parts to navigate through an environment can be modelled by a combination of an overall positional advance with an overlaid asymmetric oscillatory pattern, a distinctive signature that tends to be absent in non-biological objects in locomotion. In this report, this key trait of positional advance with asymmetric oscillation along with differences in an object's common motion (extrinsic motion) and localized motion of its parts (intrinsic motion) is exploited to detect biolocomotion. In particular, a computational algorithm is developed to measure the presence of these traits in tracked objects to determine if they correspond to a biological entity in locomotion. An alternative algorithm, based on generic handcrafted features combined with learning is assembled out of components from allied areas of investigation, also is presented as a basis of comparison to the main proposed algorithm. \\

A novel biolocomotion dataset encompassing a wide range of moving biological and non-biological objects in natural settings is provided. Also, biolocomotion annotations to an extant camouflage animals dataset is provided. Quantitative results indicate that the proposed algorithm considerably outperforms the alternative approach, supporting the hypothesis that biolocomotion can be detected reliably based on its distinct signature of positional advance with asymmetric oscillation and extrinsic/intrinsic motion dissimilarity.
\end{abstract}

\newpage


\setcounter{page}{6}

\begin{section}{Introduction}
	\pagenumbering{arabic}
	\begin{subsection}{Motivation}
Videos have become a vital component of our lives as they contain important information about the world. 
This information has served humans in various domains from security to robotics to entertainment and many more. 
Not only are videos important sources of information, their sheer quantity is becoming overwhelming. 
Given the potential quality of information available from videos and their vast quantity, automated systems for processing and analyzing videos are of great importance. \\

A wide range of video analysis tasks have been considered in computer vision (e.g., segmentation \cite{Caelles19, Yao19}, tracking \cite{Cannons08,Fiaz18,Kristan16}, and action recognition \cite{Heilbron15,Kang16,Kong18}). Curiously, however, a task that appears to not have yet been addressed is the detection of biological entities as they locomote through their environment. In this report, general biological objects in locomotion will be referred to as \textbf{biolocomotion} and its detection will refer to determining its spatiotemporal loci in the video. \\

Be it a natural or artificial intelligent system, the ability to detect biological entities as they locomote would provide the system with powerful information about what subsequent actions to take. 
In the realm of humans, the detection of biolocomotion can be used in monitoring systems to focus on regions of interest, 
assistive robots to adapt its behaviour to assist its person of interest \cite{Leo17}, sports analysis for broadcast, coaching, or training \cite{Moeslund14, Thomas17}, and autonomous vehicle technology for safe navigation. Similarly, its application can be extended to other animals to monitor wildlife (e.g., to help preserve biodiversity). 
In addition, a concrete algorithm for biolocomotion detection could provide the basis for a model of how natural systems perform this task. \\

Biological evidence suggests that natural systems have the ability to detect biolocomotion on the basis of very limited visual data \cite{Johnson06,Troje06}. This ability is especially striking given the wide range of species and modes of locomotion; see Figure~\ref{fig:bioloco}. Perhaps the very wide range of intra-class variations has discouraged previous researchers in artificial systems from delving into biolocomotion detection. Nevertheless, the ability of biological systems to detect biolocomotion from limited data raises the possibility that there may be a distinctive signature to biolocomotion independent of species type and mode of locomotion that could be leveraged by a computer vision system. Given that all biological systems are governed by biomechanical principles, a possible basis for such a signature comes from biomechanics. Fortunately, there is a rich literature on the biomechanics of animal locomotion from which to draw \cite{Alexander03,Biewener03}. \\

\begin{figure}[!h]
	\begin{tabular}{c}
		\includegraphics[width=0.98\textwidth]{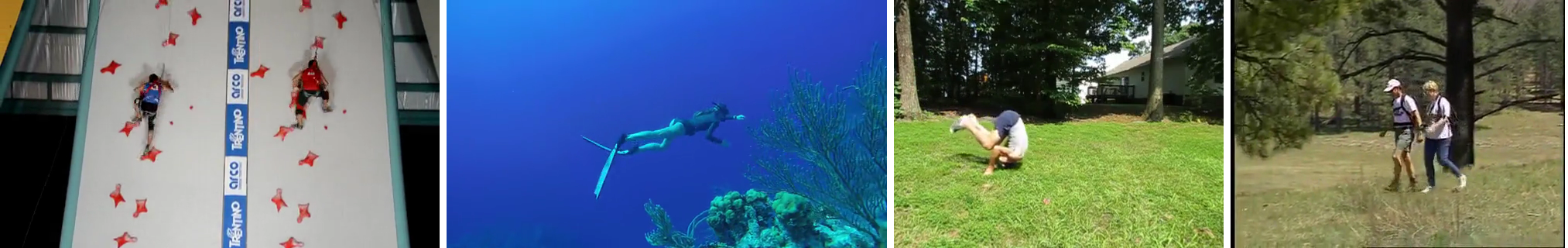} \\
		(a) Humans can locomote in various forms (e.g., climb, swim, roll, and walk). \\
		\includegraphics[width=0.98\textwidth]{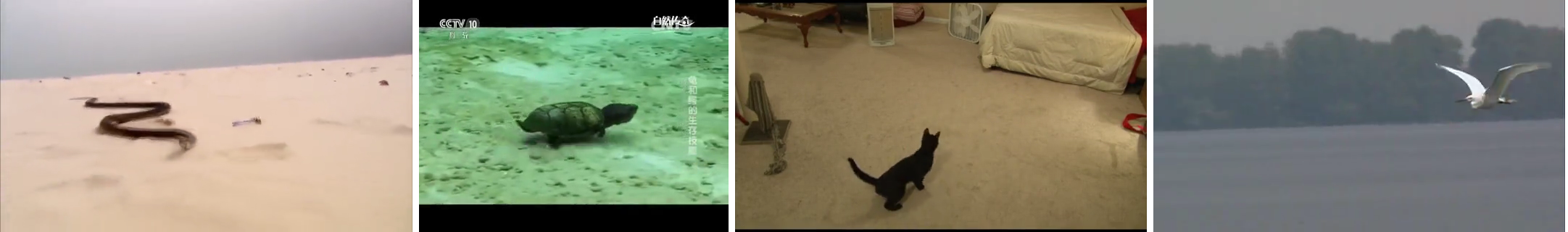} \\
		(b) Variations in biological species (e.g., snake, turtle, terrestrial quadruped, and bird).
	\end{tabular}
	\caption[Variations in modes of locomotion and species]{Variations in modes of locomotion and species. (a) Humans can locomote in various ways (e.g., climb, swim, roll, and walk) and (b) biodiversity encompasses a wide range of species (e.g., snake, turtle, cat, and bird). \label{fig:bioloco}}
\end{figure}

The apparent ability of natural systems to detect biolocomotion from visual data combined with its potential usefulness provides a strong motivation for studying biolocomotion from a computational perspective. 

\end{subsection}

\begin{subsection}{Challenges}
	Like any image- or video-based recognition or detection task, biolocomotion detection must be robust to variable acquisition scenarios (e.g., illumination, clutter, intrinsic and extrinsic camera parameters). Beyond these usual concerns, two additional outstanding challenges must be considered in biolocomotion detection in videos. First, there is extreme diversity in how biological entities move through the environment (e.g., mammals use legs, birds use wings, and fish use fins); even within species, wide variations are present (e.g., humans can walk, run, skip, or swim). Thus, a general biolocomotion detection must be robust to such variations. Second, prior to the work described in this report, there were no extant datasets suitable for developing and evaluating biolocomotion detection algorithms. Correspondingly, it was necessary to construct a novel dataset for the task.
\end{subsection}

\begin{subsection}{Contributions}
The contributions of this report are as follows.
\begin{itemize}
	\item Biolocomotion detection in videos is introduced as a new research topic in computer vision. Despite the strong motivation for this area of study, it appears that no previous computer vision research has addressed this topic.
	\item A novel algorithm capable of spatiotemporally detecting biolocomotion in videos is proposed. The proposed algorithm is motivated by biomechanical properties of animals in locomotion and psychophysical studies on biological motion perception; thus, it benefits from not having to learn the within class variations of biological and non-biological objects in motion.
	\item An alternative algorithm for biolocomotion detection in videos is presented to provide the main proposed algorithm with a basis for comparison in evaluation. This algorithm is assembled out of components from allied areas of investigation (i.e., action proposals \cite{vanGemert15} and recognition \cite{Wang13}) and,  unlike the main proposed algorithm, relies on training rather than biomechanical modelling.
	\item A novel biolocomotion video dataset is introduced. The dataset is extremely diverse in capturing terrestrial, aquatic, and aerial biological entities as well as non-biological objects moving in various ways. Biolocomotion groundtruth labels are provided for this dataset as well as an extant camouflage animals dataset from allied area of motion segmentation \cite{Bideau16}. These datasets are used to evaluate both of the developed algorithms. 
\end{itemize}

\end{subsection}

\begin{subsection}{Outline}
This report unfolds in five sections. 
This initial section has served to motivate the importance of the proposed problem, biolocomotion detection in videos.
Section~\ref{ch:relatedWork} covers related work from various fields ranging across biomechanics, psychophysics, and computer vision. 
Section~\ref{ch:techApproach} provides a unified computational algorithm inspired by biomechanics and psychophysics to detect biolocomotion in videos.
Section~\ref{ch:evaluation} provides empirical evaluation of the approach, including introduction of a novel biolocomotion dataset as well as an alternative baseline algorithm. 
Finally, Section~\ref{ch:conclusion} provides an overall summary of the presented research, as well as suggestions for future research. 
\end{subsection}
\end{section}

\begin{section}{Related research}\label{ch:relatedWork}
	\begin{subsection}{Overview}\label{sec:related_intro}
Animal locomotion has been studied extensively in ethology 
for some time. Animals locomote for various reasons: to search for food, to find suitable habitat, to pursue prey, to escape from predators, or to seek a mate. The grand scale of biodiversity (ranging from mammals, lizards, birds, fish, insects, and many more) contributes to the great locomotory design and mode diversity. Fortunately, there are common principles that underlie most of these components. Thus, understanding these physical principles would provide a general understanding of why certain biological structures evolved for movement.
In complement, biological motion can be detected with very limited visual data by natural visual systems \cite{Johnson06,Troje06}. The goal of this report is to use common traits found in biomechanics of moving biological species and biological motion perception in psychophysics 
to build a unified computational approach to detect arbitrary locomoting biological species in videos. \\

This section unfolds in five subsections. 
This first subsection has served to identify related fields for the development of a biolocomotion detection algorithm in video (i.e., biomechanics and psychophysics). 
Subsection~\ref{sec:biomechanics} covers the biomechanical properties that underlie locomotion in biological species. 
Subsection~\ref{sec:psychophysics} describes characteristic motions that induce the perception of biological motion. 
Subsection~\ref{sec:comp_models} describes previous computational work developed for biological motion analysis. Finally, Subsection~\ref{sec:related_summary} provides an overall summary of the work from various fields that are necessary to build a unified approach for the detection of biolocomotion in videos.
Note that extensive reviews of animal locomotion using biomechanical properties can be found in \cite{Alexander03, Biewener03} and surveys on the perception of biological motion in psychophysics can be found in \cite{Troje08,Troje12a}.
\end{subsection}

\begin{subsection}{Biomechanics}\label{sec:biomechanics}
While the means by which animals traverse their environment is extremely varied, common locomotory mechanisms have emerged as a result of biomechanical constraints. In particular, animal locomotion is typically accompanied by the overall positional advance with an asymmetric oscillatory trace of the body parts to provide a propulsive force, as will be detailed in this section. \\

Land, air, and water constitute the type of environments animals move through. 
The properties of these media, such as density and viscosity, 
can influence the locomotory mechanisms evolved by the animals \cite{Biewener03}. 
Air has lower density than water. 
Thus, aerial animals must exert sufficient forces to support their weight in air, while most aquatic animals are neutrally buoyant since their body density is nearly the same as water \cite{Biewener03}. 
Air also has lower viscosity than water. This lower viscosity imposes less, but non-absent, drag forces on flying and terrestrial animals compared to aquatic animals. In essence: Aerial animals must generate enough force to lift their bodies as well as thrust to overcome the aerodynamic drag forces associated with moving forward; 
aquatic animals must swim strategically to reduce drag forces induced by high viscosity of water; and
terrestrial animals must overcome gravitational forces as they move. \\
For each medium, animals can locomote in various forms. Movement in air can be achieved by gliding or flapping, where flight by \textit{glide} generates lift by keeping the wing fixed and exploits the airflow for movement and \textit{flapping} generates lift and thrust forces simultaneously by continuous wing oscillation \cite{Alexander03,Calisti17}. 
Movement in water can be achieved by lift-powered swimming, undulation, drag-powered swimming, or jet propulsion.
\textit{Lift-powered swimming} involves flapping fins or tail to propel forward, \textit{undulation} refers to the oscillation of the entire body, \textit{drag-powered swimming} pushes water backwards by using fins or limbs as oars that move back and forth, and \textit{jet-propulsion} involves sequential ingestion and expulsion of finite mass of water \cite{Alexander03,Calisti17}.
Movement on land can be achieved by crawling, walking, running, hopping, or jumping \cite{Alexander03,Calisti17}.
A common trait across these different means of powered locomotion is the activation of muscles to lengthen and shorten at constantly changing speeds to accelerate and decelerate moving body parts. An exception is glide, as glide is an unpowered flight as it generates no mechanical power with its flight muscles \cite{Biewener03,Calisti17}.
These powered actions apply forces in an oscillatory manner such that a structure with mass (body and/or its part) oscillates in its environment \cite{Alexander03, Biewener03}. \\

A generalized model to understand the dynamics of legged terrestrial locomotion (e.g., trot, run, and hop) often builds on the \textbf{bouncing spring-mass model} \cite{Blickhan93,McMahon90}. The spring-mass model consists of a massless spring attached to a point mass, where the leg is represented by the spring and the body of an animal is represented by the point mass. A point-mass spring has been used to model uniformly a wide range of species (e.g., humans, dogs, kangaroos, land birds, crabs, and cockroaches) across different locomotory designs (e.g., number, length, shape, position, and skeleton type of legs), as the relative vertical ground-reaction force and the relative compression of the leg are the essential, yet common, components required to move an animal's centre of mass \cite{Blickhan87}. 
Indeed, several walking \cite{Collins05,McGeer90}, running, and hopping \cite{Raibert86b} robots and computational models for tracking  \cite{Brubaker07} have emerged through the understanding of legged locomotion via the bouncing spring-mass model 
\cite{Calisti17, McGeer90}. Overall, the bouncing spring-mass model further underlines that oscillation can be an important component of biological locomotion.\\


The use of rotating systems as a means of transport has brought tremendous efficiency in artificial locomotory devices (e.g., wheels and propellers). However, very few biological species have adapted rotating systems as a means for locomotion; notable exceptions include bacterial flagellum, rolling spiders, caterpiller-tred stomatopods, and pangolins \cite{LaBarbera83}. The lack of rotating systems in natural systems can be attributed to environmental constraints or the level of efficiency. That is, consider wheel-based transportation on terrestrial surfaces. These are efficient modes on flat rigid terrains (e.g., asphalt and concrete), but far less on irregular terrains - a very common characteristic of natural terrains. In addition, wheel-based transportation require corners to be wide enough and not too sharp for manoeuvrability, which would be a severe disadvantage for species in cluttered terrains. As a result, natural selection favours the evolution of limbs capable of travelling on irregular surfaces and manoeuvring around obstacles \cite{Biewener03, LaBarbera83}. 
As another example, thrusting in aerial or aquatic media by propellers (as done by artificial locomotory devices) is far less energetically efficient than oscillating flexible foils 
(as in caudal fin of fish and bird wings) \cite{Katz79}. 
Consequently, natural systems favour the oscillation of the body and/or its appendages for its energetic savings \cite{LaBarbera83}. \\

Moving efficiently is a very important aspect of biological motion. 
In terrestrial locomotion, specifically in walking, electromyography (EMG) data revealed that once the leg muscles are activated to set the foot into motion during the stance phase, its muscles are almost inactive during the swing phase, such that the foot moves entirely under the influence of gravity \cite{Crowninshield81}. 
In aerial locomotion, the wings rotate to move down and in front of its body during the downstroke and move up and slightly backwards during the upstroke to generate lift and thrust forces simultaneously \cite{Berg10, Biewener03,Yu05}. 
In aquatic locomotion, the body and/or appendages accelerate to produce a propulsive force, then decelerate before initiating its subsequent propulsive stroke to maintain steady speed in its viscous environment \cite{Biewener03, Fish16}.
In general, the effective means of locomotion in various media results in an asymmetric path traced by body parts of animals as they traverse their environments. This asymmetry results because a steeper slope is observed during the lift/stance phase of a walk compared to the swing phase, downstroke compared to upstroke during flight, and initiation compared to completion of a propulsive stroke in swimming; see Figure~\ref{fig:teaser_biomech}. \\

\begin{figure}[!h]
	\begin{center}	
		\resizebox{0.9\textwidth}{!}{
		\begin{tabular}{c c c}
			\includegraphics[width=0.2\linewidth]{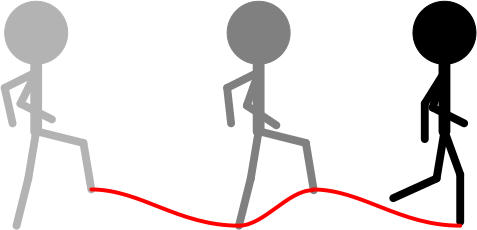}
			& \includegraphics[width=0.35\linewidth]{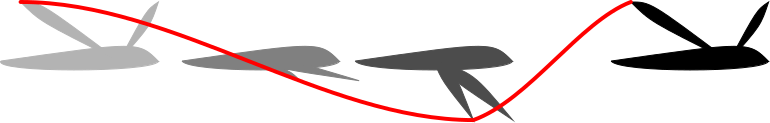}
			& \includegraphics[width=0.35\linewidth]{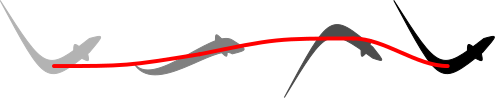}
		\end{tabular}
		}
	\end{center}
	\caption[Illustration of motion patterns exhibited by locomoting biological species]{Illustration of characteristic motion patterns exhibited by locomoting biological objects in various media (human on land, bird in air, and fish in water). As objects move across time (indicated by varying grey levels, with lighter-to-darker moving forward in time), the means that biological objects use to locomote (legs for humans, wings for birds, and body/fins for fish) exhibit an asymmetric sinusoidal pattern (marked in red). \label{fig:teaser_biomech}}
\end{figure}

Overall, consideration of the biomechanics of animal locomotion shows that regardless of the wide range of locomotory designs of different animals in various media, animal locomotion involves the use of its body and/or appendages to generate propulsion. Specifically, the acceleration and deceleration in an oscillatory fashion of the body and/or its parts to move in its environment is commonly observed. Moreover, the oscillation typically unfolds in an asymmetric fashion across time. Provided such patterns can be visually observed, there is potential for the development of a principled approach for biolocomotion detection in videos.
\end{subsection}

\begin{subsection}{Psychophysics}\label{sec:psychophysics}

Psychophysical studies have shown that human observers are able to perceive a set of dynamic dots as a coherent figure representative of a person or other animals in motion, provided the dots are located near major joints and their motions are consistent with their representative figures \cite{Johansson73,Mather93}. These visual stimuli are referred to as \textbf{point-light displays}; see Figure~\ref{fig:point_light_display}. Studies on point-light displays show that biological motion can be perceived in the absence of any relevant appearance information (e.g., body silhouette, texture, or colour). It also has been shown that other animals can perceive biological motion in such displays (e.g., cats \cite{Blake93}, pigeons \cite{Dittrich98}, and chicks \cite{Vallortigara06}). Indeed, not only can gross motion patterns be discriminated (e.g., walk, run, and stair climb), but more subtle differences can be perceived (e.g., gender \cite{Barclay78, Kozlowski77, Troje02} and emotion \cite{Dittrich96,Pollick01}). It is argued that in making such inferences, biological visual systems decompose motion in terms of common motion, referred to as \textbf{extrinsic motion}, and relative motion between parts, referred to as \textbf{intrinsic motion} \cite{Johansson73}. Interestingly, neuroimaging studies have been able to localize biological motion processing in the brain \cite{Chang18}. \\

\begin{figure}[!h]
	\begin{center}
		\begin{tabular}{c c c}

	\begin{tikzpicture}[scale=3.5]
		
		\draw[fill=black] (0,0) rectangle (0.9,1.4);
		
	         \draw[fill=white, draw=none] (0.46,1.17) circle (0.8pt);
	         \draw[fill=white, draw=none] (0.46,1.02) circle (0.8pt);
	         \draw[fill=white, draw=none] (0.44,0.86) circle (0.8pt);
	         \draw[fill=white, draw=none] (0.52,0.85) circle (0.8pt);
	         \draw[fill=white, draw=none] (0.35,0.77) circle (0.8pt);
	         \draw[fill=white, draw=none] (0.44,0.71) circle (0.8pt);
	         \draw[fill=white, draw=none] (0.50,0.68) circle (0.8pt);
	         \draw[fill=white, draw=none] (0.39,0.46) circle (0.8pt);
	         \draw[fill=white, draw=none] (0.51,0.47) circle (0.8pt);
	         \draw[fill=white, draw=none] (0.35,0.20) circle (0.8pt);
	         \draw[fill=white, draw=none] (0.61,0.27) circle (0.8pt);
	\end{tikzpicture}
	
	& 
	
	\begin{tikzpicture}[scale=3.5]
		
		\draw[fill=black] (0,0) rectangle (0.9,1.4);
	
	         \draw[fill=white, draw=none] (0.46,1.18) circle (0.8pt);
	         \draw[fill=white, draw=none] (0.46,1.03) circle (0.8pt);
	         \draw[fill=white, draw=none] (0.46,0.87) circle (0.8pt);
	         \draw[fill=white, draw=none] (0.50,0.86) circle (0.8pt);
	         \draw[fill=white, draw=none] (0.38,0.74) circle (0.8pt);
	         \draw[fill=white, draw=none] (0.43,0.72) circle (0.8pt);
	         \draw[fill=white, draw=none] (0.48,0.69) circle (0.8pt);
	         \draw[fill=white, draw=none] (0.47,0.47) circle (0.8pt);
	         \draw[fill=white, draw=none] (0.40,0.47) circle (0.8pt);
	         \draw[fill=white, draw=none] (0.41,0.21) circle (0.8pt);
	         \draw[fill=white, draw=none] (0.60,0.32) circle (0.8pt);
	         
	\end{tikzpicture}
	
	&
	
	\begin{tikzpicture}[scale=3.5]
	
		\draw[fill=black] (0,0) rectangle (0.9,1.4);
	
	         \draw[fill=white, draw=none] (0.47,1.20) circle (0.8pt);
	         \draw[fill=white, draw=none] (0.46,1.05) circle (0.8pt);
	         \draw[fill=white, draw=none] (0.51,0.89) circle (0.8pt);
	         \draw[fill=white, draw=none] (0.46,0.87) circle (0.8pt);
	         \draw[fill=white, draw=none] (0.39,0.73) circle (0.8pt);
	         \draw[fill=white, draw=none] (0.45,0.73) circle (0.8pt);
	         \draw[fill=white, draw=none] (0.47,0.72) circle (0.8pt);
	         \draw[fill=white, draw=none] (0.48,0.48) circle (0.8pt);
	         \draw[fill=white, draw=none] (0.40,0.48) circle (0.8pt);
	         \draw[fill=white, draw=none] (0.43,0.26) circle (0.8pt);
	         \draw[fill=white, draw=none] (0.52,0.20) circle (0.8pt);
	
	\end{tikzpicture}
\end{tabular}
	\end{center}
	\caption[Illustration of a point-light display]{Illustration of a point-light display. A set of 11 markers are used to represent the major joints of the human body (the head, shoulder, hip, two elbows, two wrists, two knees, and two ankles). Three frames from a sequence of such markers animated in accord with the motion of a walking person. Human observers perceive a temporal sequence of such frames as resulting from a walking human. Figure redrawn from \cite{Johansson73}. \label{fig:point_light_display}}
\end{figure}
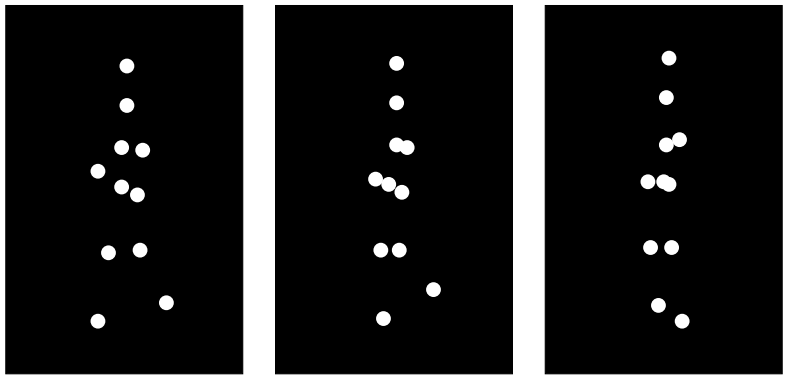

The \textbf{direction discrimination task} is a task that asks the observers to indicate \textit{which} direction (left or right) the figures depicted in point-light displays are facing. Such experiments have shown that direction discrimination accuracies are highly correlated with the amount the displays appeared animate when variations of the display (coherent vs. scrambled and upright vs. inverted) were presented to the observers \cite{Chang08}. Consequently, the direction discrimination task has often been used to study the perception of biological motion. Further studies on the direction discrimination task have indicated that the local motion, of feet in particular, play a vital role in the accuracy of direction identification \cite{Johnson06,Troje06}. 
Experiments comparing the displays of naturally accelerating foot motions with those containing constant speeds revealed that the acceleration contained in the foot motion plays a significant role in direction discrimination accuracy \cite{Chang09, Hirai11}. It has been shown that other animals (e.g., newly hatched chicks) show similar sensitivity to vertical acceleration, as they respond to upright point-light displays (of a hen) by aligning their bodies in the apparent direction of motion but not to inverted displays \cite{Vallortigara06}. 
These findings suggest that biological motion perception is based on the vertical acceleration patterns the foot exhibits as an animal moves through the environment. \\

Psychophysical evidence suggests that humans can make inferences of animacy from the trajectory that an object traces \cite{Bingham95}. 
Further evidence shows that humans can discriminate between symmetric and asymmetric trajectories \cite{Muchisky02}. Combining these pieces of evidence with the above reviewed work on point-light displays suggests that motion information similar to those found in biomechanics that indicate locomoting biological species (e.g., asymmetric oscillatory traces) may also be exploited in biological vision systems. Moreover, they suggest the potential efficacy of decomposing motion into intrinsic and extrinsic components in making such inferences. 

\end{subsection}

\begin{subsection}{Computational Vision}\label{sec:comp_models}
Some computational vision work was inspired by the ability of humans to perceive biological structure and motion in point-light displays viewed across time. Early work along these lines reconstructed the 3D structure and motion of animals using anatomical constraints by observing that animal limbs are (i) rigid, (ii) have a fixed length, and (iii) typically move in a single plane for extended periods of time \cite{Hoffman82}. While these anatomical constraints are generally true of legged terrestrial and aerial animals, they are not true for undulating animals. Furthermore, estimating the species-invariant animal pose in a non-intrusive way with a limited set of training data is a challenging task that further limits the exploitation of the proposed method from working on point-light displays.\\

Other work made use of 3D periodicity constraints \cite{Zhang07}. The goal is to reconstruct a 3D structure from motion capture data of humans, rather than to detect general biological species in locomotion. Nevertheless, a walking human is described as a Fourier representation of (i) average posture, (ii) characteristic postures of the fundamental frequency, (iii) second harmonic of a discrete Fourier expansion, (iv) fundamental frequency, which does not explicitly model vertical acceleration that is present. While the notion of asymmetry is mentioned, it is in the context of left and right limb asymmetry and not within a stride as prevalent in animal locomotion. \\

Taking further inspiration from biology regarding how the mammalian visual system appears to process information in two parallel streams for form and motion \cite{Goodale92}, a corresponding algorithm was developed to infer biological shape and motion models from sparse point displays \cite{Giese03}. Specifically, the form pathway that analyzes the body shapes was modelled using Gabor-like filters \cite{Jones87} to obtain orientation details, max-like pooling \cite{Riesenhuber99} to provide position and scale-invariance, then Gaussian radial basis functions \cite{Poggio90} to support selectivity towards complex shapes. 
The motion pathway was modelled using optical flow \cite{Giese02} patterns to mimic the direction sensitive and motion sensitive patterns in our brain. 
While their model provided a computational demonstration that the motion (dorsal) pathway is predominantly active (and that the form (ventral) pathway tends not to be activated) in the recognition of the point-light displays, it did not address how the motion model can be used to detect general biolocomotion nor the specific motion patterns that were learned for the categorization. 

Yet another method considered decomposing motion exhibited by an animation (e.g., a person walking or strutting, a kangaroo or a rabbit hopping) into global and local components \cite{Shavit93}, akin to extrinsic and intrinsic motions used to describe perceptual organization in the biological visual system \cite{Johansson73}. The global component is responsible for measuring the motion of the object's centre of mass and the local component measures the rate of dispersion of the object about its centre of mass. 
Similar to other computational work to date, this algorithm did not address how a general biological species can be detected nor how a non-biolocomoting object can be rejected. \\


Periodic motion has been used in previous work to detect, track, and classify objects in videos (e.g.,  humans and dogs)  \cite{Aggarwal99,Briassouli07,Cutler00,Niyogi94,Polana97,Ran07,Seitz97,Tsai94}. 
Across this research direction, various approaches have been proposed for periodic motion detection, including time-frequency analysis \cite{Briassouli07,Cutler00,Runia18,Tsai94}, period trace \cite{Seitz97}, hypothesis testing on periodograms \cite{Ran07}, and convolutional neural networks (CNNs) \cite{Levy15}. A limitation of these approaches is that they rely on non-trivial preprocessing of their input videos, including extraction of points corresponding to the major joints of the human body \cite{Tsai94}, conversion to figure-centric volumes \cite{Cutler00,Polana97,Ran07}, background-subtraction \cite{Briassouli07}, require a precise slicing of a video along the XT-axis \cite{Niyogi94}, or require static cameras \cite{Levy15} to obtain data indicative of periodic motion. 
Furthermore, the computational vision literature often has modelled a person's walk via an inverted pendulum \cite{Aggarwal99,Srinivasan06}, while a spring-mass system is a more accurate representation, as it accounts for the vertical ground-reaction force \cite{Geyer06}. 
Significantly, none of these approaches have been applied to the challenge of general biolocomotion detection of species on land, in air, and in water. Indeed, analysis of oscillation alone does not suffice for detection of biolocomotion, as not all oscillating objects are locomoting biological entities (e.g., person using a jump rope or a pendulum). \\


There has been growing interest in applications of computer vision to species classification and detection for wildlife. 
Correspondingly, several datasets that concentrate on imagery of wildlife in natural habitats (e.g., CUB200 \cite{Welinder10}, NABirds700 \cite{vanHorn15},  iNat2017 \cite{vanHorn18}, Snapshot Serengeti \cite{Swanson15}, Missouri Camera-Trap \cite{Zhang16}, and CCT-20 \cite{Beery18}) have been made available to the research community.
CUB200, NABirds700,  iNat2017, and CCT-20 are image datasets; 
and the Snapshot Serengeti dataset consists of image sequences collected from a camera trap, which are heat- or motion-activated cameras that capture a single image or a short sequences of images (1-5 frames with a frame rate of approximately 1 frame per second) at each trigger \cite{Swann11}. 
The interest of the current report, however, lies in common videos rather than specialized ones, which are typically of longer duration and have higher frame rate (typically between 25-30 fps).
Furthermore, while datasets extracted from camera traps (e.g., Snapshot Serengeti, Missouri Camera-Trap, and CCT-20) pose various challenges similar to those in-the-wild, such as dynamic background, illumination changes, cluttered and dynamic scenes, they lack camera motion. Thus, some detection algorithms developed to perform reasonably on camera trap data are limited to videos with limited camera motion \cite{Beery18,Lin14}. 
Moreover, CUB200 and NABirds700 focus on fine-grained species classification, thus only consist of bird images, while the goal of the present report is to detect a wide range of species in locomotion. 
Thus, current algorithms developed to perform reasonably on wildlife datasets are either constrained to images (i.e., not video) \cite{Chen14}, images that have been manually cropped to delineate the animals of interest \cite{Chen14,Yu13}, species-specific \cite{Miguel16,Wilber13}, or background-specific \cite{Zhang15,Zhang16} requiring sufficient training data to model those backgrounds in the test set \cite{Chen14}. \\

Overall, while considerable computational work has addressed biological motion analysis or detection and classification of biological species in images, none has considered the detection of general biolocomotion in videos. 
\end{subsection}

\begin{subsection}{Summary}\label{sec:related_summary}

Despite the diverse locomotory designs that exist in different animals, there are significant mechanical and energetic similarities in the body and/or its parts for various types of locomotion. These underlying biomechanical constraints have been incorporated into minimalist models of legged animal locomotion in terms of a bouncing mass-spring model \cite{Blickhan93,McMahon85,Raibert86} that have been successfully applied to a wide range of terrestrial animals. Significantly, the implied overall positional advance of the body along with asymmetric oscillatory motion of its parts is present not just in terrestrial legged locomotion, but also extends to non-legged terrestrial \cite{Marvi12},  aerial \cite{Portugal13}, and aquatic \cite{Lauder15,Triantafyllou00,Wen13} creatures. \\

Psychophysical studies suggest that the vertical acceleration pattern induced by gravitational and biomechanical constraints in terrestrial creatures play a significant role in the perception of biological motion \cite{Chang09, Hirai11}. Vertical acceleration exhibited by these creatures as they push off against gravity causes their trajectories to trace asymmetric oscillatory patterns, while non-biological objects that locomote with oscillation display symmetric cycloidal patterns during their advance (e.g., rolling objects). 
Moreover, other studies from psychophysics have revealed that the human visual system decomposes the kinematics of an object into common translatory and residual motion (i.e., extrinsic and intrinsic motion) to understand the mechanics of a scene \cite{Johansson73, Johansson74}. The overall direction of an object (e.g., translation of a walker) and the local cues of the body (e.g., trajectory of the feet) tend to be different in locomoting biological entities as compared to non-biological objects. \\

While previous computational work has addressed animal motion analysis, none has addressed the detection of general biolocomotion in videos. Building on work from the biomechanics and perceptual psychophysics of biological motion, the remainder of this report presents the first algorithm capable of spatiotemporally detecting biolocomotion in videos. Notably, the biomechanical and psychophysical principles motivate the definition of species-invariant biolocomotion signatures so that the approach benefits from not needing to model the wide range of within class variations of biological and non-biological objects in motion.

\end{subsection}	
\end{section}

\begin{section}{Technical approach}\label{ch:techApproach}
	\begin{subsection}{Overview}

The goal of a biolocomotion detection algorithm is to take an unprocessed video and output the spatiotemporal coordinates of biolocomotion; see Figure~\ref{fig:general_framework}. To achieve this goal, distinctive properties of biolocomotion as seen in videos must be defined. 
A key trait that can be observed from biomechanics of animal locomotion and perception of biological motion is a directional trajectory modulated by asymmetric oscillation along with differences in its overall motion and its local cues.
This observation is used as the basis for biolocomotion detection. In particular, a collection of trajectories across an image sequence that show an overall advance in spatial position of their tracked elements (i.e., \textit{locomotion}) that exhibit \textit{asymmetric oscillation} and overall motion (\textit{extrinsic motion}) difference from its local cues (\textit{intrinsic motion}) is detected; e.g., Figure~\ref{fig:teaser}.\\ 

\begin{figure}[H]
	\begin{center}
	\resizebox{\textwidth}{!}{
		\input{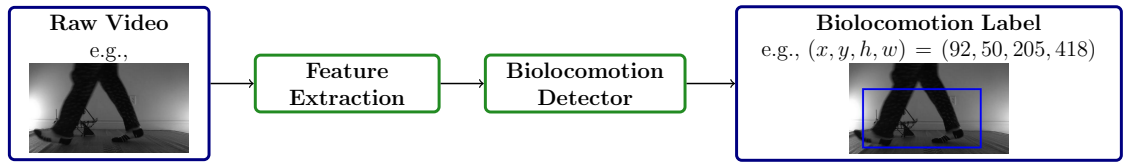}
	}
	\end{center}
	\caption[General framework of the biolocomotion detection algorithm]{General framework of the biolocomotion detection algorithm. The input and output of the proposed biolocomotion algorithm are marked in blue and the intermediate components that are involved in obtaining the spatiotemporal coordinates of locomoting biological objects in a video are marked in green.} \label{fig:general_framework}
\end{figure}

This section unfolds in five subsections. 
This first subsection has served to define the problem under consideration and outline the components necessary for biolocomotion detection.
Subsection~\ref{sec:tech_proc} describes the extraction of primitive features that can encapsulate critical information for modelling biolocomotion in terms of image point trajectories traced during biolocomotion. Subsection~\ref{sec:tech_algorithm} presents algorithmic measures that map the extracted features to various components of the developed biolocomotion detector. Subsection~\ref{sec:tech_aSTW} presents a sliding window realization of the approach for continuous video processing. Finally, subsection~\ref{sec:tech_summary} provides an overall summary of the approach.

\begin{landscape}
\begin{figure}
	\begin{center}	
		\resizebox{1.2\textwidth}{!}{
		\begin{tabular}{c c c c}
			\includegraphics[width=0.2\linewidth]{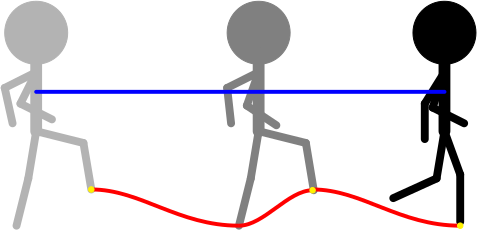}
			& \includegraphics[width=0.35\linewidth]{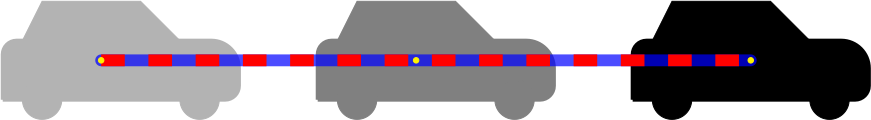} 
			& \multirow{3}{*}[\normalbaselineskip]{\includegraphics[width=0.3\textwidth]{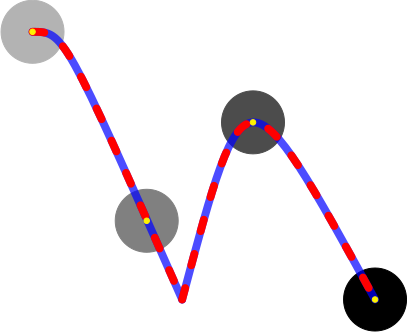}}
			& \multirow{3}{*}{\includegraphics[width=0.1\linewidth]{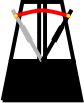}} \\
			\includegraphics[width=0.35\linewidth]{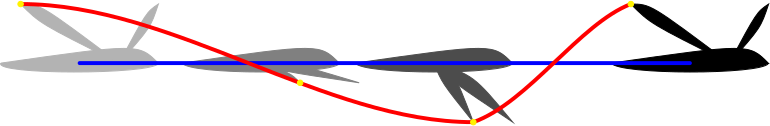}
			& \includegraphics[width=0.35\linewidth]{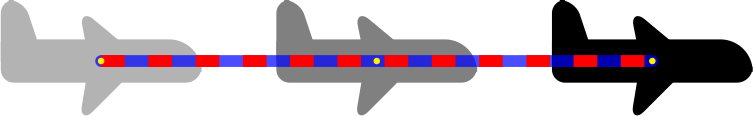} & \\
			\includegraphics[width=0.35\linewidth]{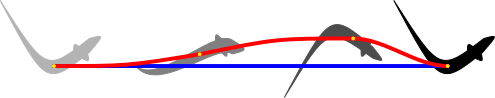} 
			& \includegraphics[width=0.35\textwidth]{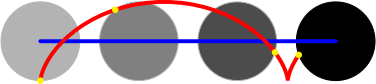} & \\
			(a) Biological objects in locomotion
			& \multicolumn{2}{c}{(b) Non-biological objects in locomotion}
			& (c) Metronome
		\end{tabular}
		}
	\end{center}
	\caption[Illustration of characteristic motion patterns exhibited by biological and non-biological objects]{Illustration of characteristic motion patterns exhibited by (a) biological objects in locomotion, (b) non-biological objects in locomotion, and (c) a non-biological object in oscillation. As objects move across time (indicated by varying grey levels, with lighter-to-darker moving forward in time), the means that biological objects use to locomote (e.g., legs for humans, wings for birds, body/fins for fish) exhibit an \textit{asymmetric} sinusoidal pattern (with a point in yellow generating the red trace). Non-biological objects in locomotion, on the other hand, tend to exhibit either trajectories that lack oscillation as they move (e.g., car and plane), trajectories with \textit{symmetric} oscillation (e.g., rolling ball), or the extrinsic motion (i.e., overall direction of motion) (blue) and its intrinsic motion (i.e., local cues) (red) tend to coincide (e.g., bouncing ball), while an oscillating non-biological object (e.g., metronome) does not have an accompanying overall positional advance. Note that the alignment of the extrinsic and intrinsic curves yield an alternating red and blue traces for locomoting non-biological objects, except those where the intrinsic component traces a cycloid.
\label{fig:teaser}}
\end{figure}
\end{landscape}

\end{subsection}

\begin{subsection}{Feature extraction}\label{sec:tech_proc}
	In this report, a collection of tracked point trajectories serve as inputs to the biolocomotion detector. Thus, in this section, a definition of point trajectories and subsequent post-processing that make the trajectories more amenable to further biolocomotion processing are provided.
	
\begin{subsubsection}{Point trajectories}
	Given an input video, point trajectories provide the path in which the tracked point travelled across time. In particular, point trajectories support quantitative measurement of key components that distinguish biolocomotion: spatial advance in overall position, asymmetric oscillatory traces, and the difference between extrinsic and intrinsic motions. Consequently, a collection of tracked point trajectories are used as inputs to the biolocomotion detector. There are a variety of approaches to obtaining point trajectories available in the field of computer vision (e.g., KLT Trajectories \cite{Messing09}, SIFT Trajectories \cite{Sun09}, and Dense Trajectories \cite{Wang11,Wang13b}). In this report, the recovered trajectories are built on improved Dense Trajectories (iDTs) \cite{Wang13} that previously supported state-of-the-art performance amongst handcrafted algorithms for action recognition. This choice is made since biolocomotion detection itself can be conceptualized as a type of an action.\\
	
	The iDTs are built on Dense Trajectories (DTs), which are obtained by densely sampling feature points on a grid space by $S$ pixels over several spatial scales to ensure that the feature points are sampled from all spatial positions and scales. While denser sampling (e.g., $S=2$ to sample every other pixel) offers better performance, it significantly increases computational complexity. Thus, DTs extracted at a sampling stride of $S=20$ are used in present work. 
Since points in homogeneous areas are difficult to track reliably, they are removed using the good-features-to-track criterion \cite{Shi94}, which removes points with very small eigenvalues of the auto-correlation matrix. 
Feature points are tracked at each spatial scale separately. Each feature point $\mathbf{x}(t) = (x(t),y(t))$ at frame $t$ is tracked to the next frame $t+1$ by median filtering on a dense optical flow field $\textbf{u}(t)=(u(t),v(t))$, where $u(t)$ and $v(t)$ are horizontal and vertical components of the optical flow, respectively. Specifically, given a point $\mathbf{x}(t)$, its tracked position in the next image frame is smoothed by applying a median filter on $\mathbf{u}(t)$: 
\begin{equation}
	\mathbf{x}(t+1) = \mathbf{x}(t) + (M \ast \mathbf{u}(t))|_{\mathbf{x}(t)} \text{,}
\end{equation}
where $M$ is a $3 \times 3$ median filtering kernel.
Points of subsequent frames are concatenated to form trajectories $\mathcal{T} = [\mathbf{x}(t) \ \mathbf{x}(t+1) \ \dots \ \mathbf{x}(t+L)]$. To overcome the drifting effect (i.e., points drifting from their true locations during the tracking process), the length of the trajectories are limited to $L$ frames. Empirically, dense trajectories that span $L=15-20$ frames have been found effective in the action recognition literature \cite{Wang13b}. Thus, trajectories of $L=15$ are used in the current work.\\


Dense trajectories can be improved by removing the global background motion created by camera motion. Here, camera motion is estimated by assuming that two consecutive frames are related by a homography, where the homography is estimated by finding correspondences between two frames. The correspondences can be found by: (i) extracting SURF features \cite{Bay06} and matching them based on the nearest neighbour rule, and (ii) sampling motion vectors from optical flow using the good-features-to-track criterion. 
The candidates from the two approaches are used to estimate the homography using RANSAC \cite{Fischler81} to rectify the image to remove camera motion. Compared to the original flow, the rectified version suppresses the background camera motion and enhances the foreground moving objects. 
Trajectories generated by camera motion are removed by thresholding the displacement vectors of the trajectories in the warped flow field. If the displacement is too small, the trajectory is considered to be too similar to camera motion, and thus removed. \\
	
In this report, a tracked point trajectory, 
\begin{equation}
	\mathcal{T}^k(t) = 
	\begin{bmatrix}
		\mathbf{x}_1^k(t) & \mathbf{x}_2^k(t+1) & \cdots & \mathbf{x}_{L^k}^k(t+L^k-1)
	\end{bmatrix}
\end{equation}
	denotes trajectory $k$ that begins at frame $t$ with a temporal length of $L^k$ and its $l^{th}$ point for $1 \leq l \leq L^k$ is specified by 
\begin{equation}
	\mathbf{x}^k_l(t)=(x_l^k(t),y_l^k(t)) \text{.}
\end{equation} 
Note $\mathcal{T}^k(t)$ and $\mathcal{T}^k$ as well as $\mathbf{x}_l^k(t)$, $\mathbf{x}^k(t)$, and $\mathbf{x}_l^k$ will be used interchangeably for simplicity throughout this report, where $\mathbf{x}^k(t)$ will be used to emphasize a point present at frame $t$ and $\mathbf{x}^k_l$ to emphasize the $l^{th}$ point of trajectory $\mathcal{T}^k$. Furthermore, $x$ and $y$ components will be referred to as horizontal and vertical components, respectively, of the trajectory.\\
	
A \textbf{displacement vector} of trajectory $\mathcal{T}^k$ at frame $t$ from $\Delta t$ previous frames is defined as 
\begin{equation}\label{eq:traj_disp}
	\Delta \mathbf{x}^k_{(t,\Delta t)} = \mathbf{x}^k(t) - \mathbf{x}^k(t-\Delta t) \text{.}
\end{equation}
For simplicity, $\Delta \mathbf{x}^k(t)$ will be used to denote $\Delta \mathbf{x}^k_{(t,1)}$. \\
	
The \textbf{arc length} of trajectory $\mathcal{T}^k$ is defined as
\begin{equation}\label{eq:arc_length}
	\| \mathcal{T}^k \| = 
	\sum_{l=1}^{L^k-1}{ \sqrt{ (x_{l+1}^k - x_l^k)^2 + (y_{l+1}^k - y_l^k)^2 } } \text{.}
\end{equation}
	
Select measurements, such as amplitude and asymmetry, are sensitive to the spatial direction of the trajectory as it unfolds across time. 
To ensure these calculations are robust to such situations, it is necessary to \textbf{detrend} trajectory $\mathcal{T}^k$ to $\tilde{\mathcal{T}}^k$. Trajectory $\mathcal{T}^k$ can be detrended by 
(i) determining the line of best fit using least squares,
(ii) finding the angle, $\theta$, between the line of best fit and the positive x-axis of a 2D Cartesian plane,
(iii) rotating trajectory $\mathcal{T}^k$ by $-\theta$, then
(iv) applying vertical translation such that its horizontal mean is 0 (i.e., $\displaystyle \text{mean}_{\forall l}(\tilde{y_l}^k)=0$). 
Note that proper measurement of a trajectory's oscillation amplitude depends on the correct ordering of rotation and demeaning; see Figure~\ref{fig:detrendTraj}.\\

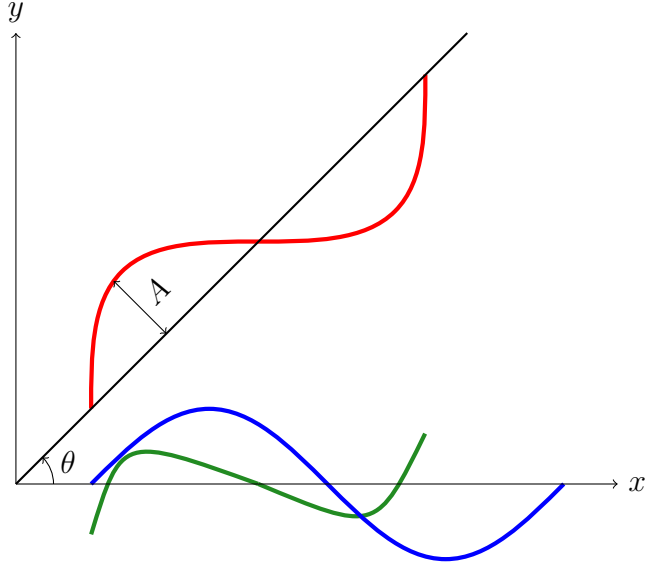
\begin{figure}[!h]
	\begin{center}
%
%

\begin{tikzpicture}
	
	\draw[ultra thick, red, rotate=45, shift={(0.71,0.71)}] (0,0) sin (pi/2,1) cos (pi,0) sin (3*pi/2,-1) cos (2*pi,0); 
	\draw[ultra thick, ForestGreen] (0,-0.67) .. controls (0.4036,0.6558) .. (2.22,0.0011);
	\draw[ultra thick, ForestGreen] (2.22,0.0011) .. controls (3.79,-0.6558) .. (4.44,0.67); 
	\draw[ultra thick, blue] (0,0) sin (pi/2,1) cos (pi,0) sin (3*pi/2,-1) cos (2*pi,0); 
		
	\draw[<->] (0.3,3-0.3) -- (1,2) node [above,right,sloped,midway,rotate=90] {$A$}; 

	\coordinate (origo) at (-1,0);
	\draw[black,->] (origo) -- ++ (8,0) node (x) [black,right] {$x$};
	\draw[black,->] (origo) -- ++(0,6) node (y) [black,above] {$y$};
	\draw[black, thick] (origo) -- ++ (6,6) node (lof) {};
	
	\pic [draw, ->, "$\theta$", angle eccentricity=1.5] {angle = x--origo--lof};

\end{tikzpicture}
	\end{center}
	\caption[Detrending a trajectory]{Detrending a trajectory. To obtain an accurate measurement of the trajectory amplitude, the trajectory must be rotated then demeaned. Given a trajectory representative of a sinusoid rotated $\theta=45^\circ$ counterclockwise about the origin with an amplitude of $A=1$ (red), the amplitude of its demeaned trajectory (green) is $\frac{1}{\sqrt{2}}$, while the amplitude of a rotated then demeaned trajectory (blue) is 1.
	\label{fig:detrendTraj}}
\end{figure}
\end{subsubsection}

\begin{subsubsection}{Trajectory post-processing}\label{sec:traj_postproc}
Collections of point trajectories serve as input to the proposed biolocomotion detector. While iDTs have previously supported state-of-the-art performance in action recognition, it is necessary to further post-process the iDT results prior to passing them to the biolocomotion detector, as follows. First, they must be pruned to remove unuseful trajectories for biolocomotion. Second, they need to be clustered so that biolocomotion detection operates on sets of trajectories that are likely to correspond to a single or similarly moving entity in the world. Third, iDTs do not have adequate robustness to camera motion for present purposes. Correspondingly, they need to be further stabilized. Fourth, it is useful to elongate them to allow for more temporal support in biolocomotion detection. In the remainder of this section, the entailed processing steps are outlined with details provided in Appendix \ref{app:traj_proc}.

\vspace{4pt}\textbf{Pruning trajectories.}
	Static trajectories and random trajectories are unlikely to provide meaningful information in identifying biolocomotion. Hence, trajectories that do not contain motion information or trajectories with sudden large displacements, which are likely to be erroneous, are removed before they are further processed. While original iDT calculations include measures to remove static and random trajectories \cite{Wang13}, they are deemed either insufficient or too aggressive for current purposes. In response, variants on conditions considered in iDT calculations have been defined and employed. See Appendix A.1 for more details.

\vspace{4pt}\textbf{Clustering trajectories.}
	Given a set of trajectories, the trajectories are clustered into disjoint sets such that each cluster corresponds to a single or similarly moving object in the world. A variant on spectral trajectory clustering is employed as the original formulation \cite{Brox10} produced poor results for the considered trajectories. 
Correspondingly, alternative measures of positional and shape affinities are defined for trajectory pairs in spectral clustering. The output of this processing stage are the centres as well as the horizontal and vertical extents of ellipses that cover each cluster. See Appendix A.2 for more details.

\vspace{4pt}\textbf{Stabilizing trajectories.}
	Many videos in-the-wild have camera motion. While iDTs are designed for some robustness to camera motion, their processing is inadequate for current purposes for two reasons. (1) In some cases, objects do not exhibit their actual locomotion in the captured video because it was recorded to stabilize the object of interest in the field of view. 
	In such cases, the tracked object would remain in the same position (exhibiting near zero displacement) within the image, while it is in locomotion in the real world (exhibiting non-zero displacement). By undoing the stabilization the camera operator has imposed, the tracked object in the image would be more representative of its locomotion, also exhibiting non-zero displacement;
	see Figure~\ref{fig:cam_stab}. The iDTs do not model such situations. (2) In other cases, iDTs simply contain too much residual motion arising from the camera to adequately capture signatures representative of biolocomotion. In response, the trajectories are stabilized to reveal the motion of the objects in the field of view with additional robustness to camera motion. See Appendix A.3 for more details.
	
\vspace{4pt}\textbf{Elongating trajectories.}
	To overcome the drifting effect when tracking points, it is recommended that iDTs only span $15-20$ frames for videos captured at 30 fps \cite{Wang13}. In contrast, biolocomotion detection benefits from longer trajectories, especially to provide sufficient time to diagnose asymmetric oscillatory behaviour. In response, iDTs are concatenated across frames based on spatial proximity and appearance similarity to obtain elongated trajectories. See Appendix A.4 for more details.

\newgeometry{top=1.25in,bottom=1.25in}
\begin{figure}[H]
	\begin{center}
		\begin{tabular}{C{\textwidth}}
			\includegraphics[width=\textwidth]{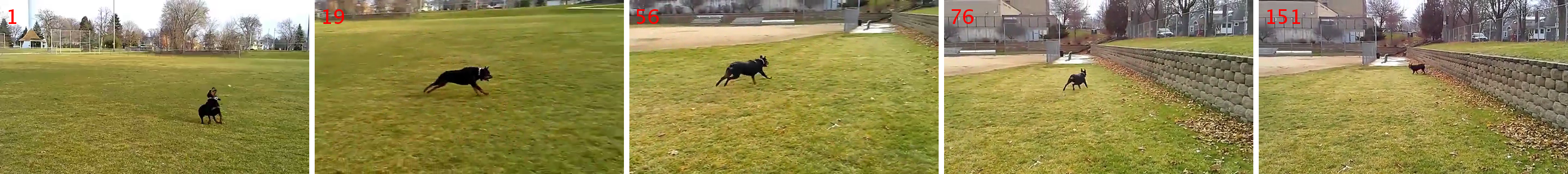} \\
			(a) Object-centric image sequence from operator imposed camera motion to track object of interest. \vspace{3mm} \\ 
			\includegraphics[width=0.5\textwidth]{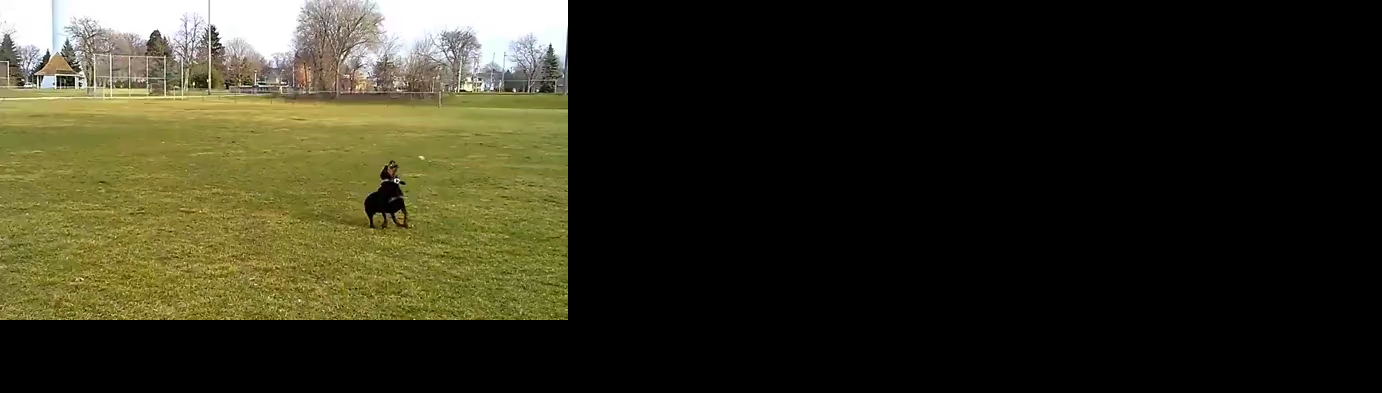}\\
			\includegraphics[width=0.5\textwidth]{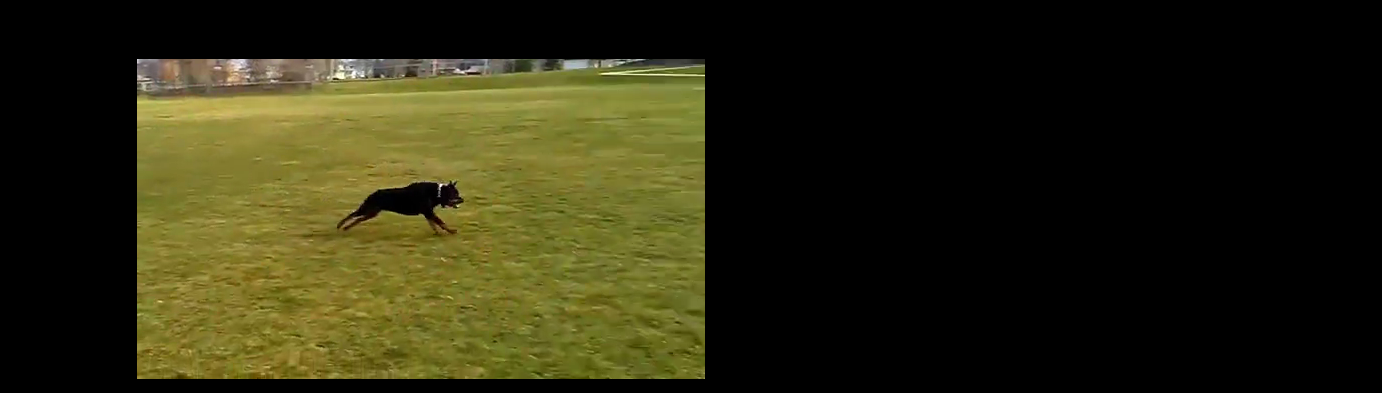}\\
			\includegraphics[width=0.5\textwidth]{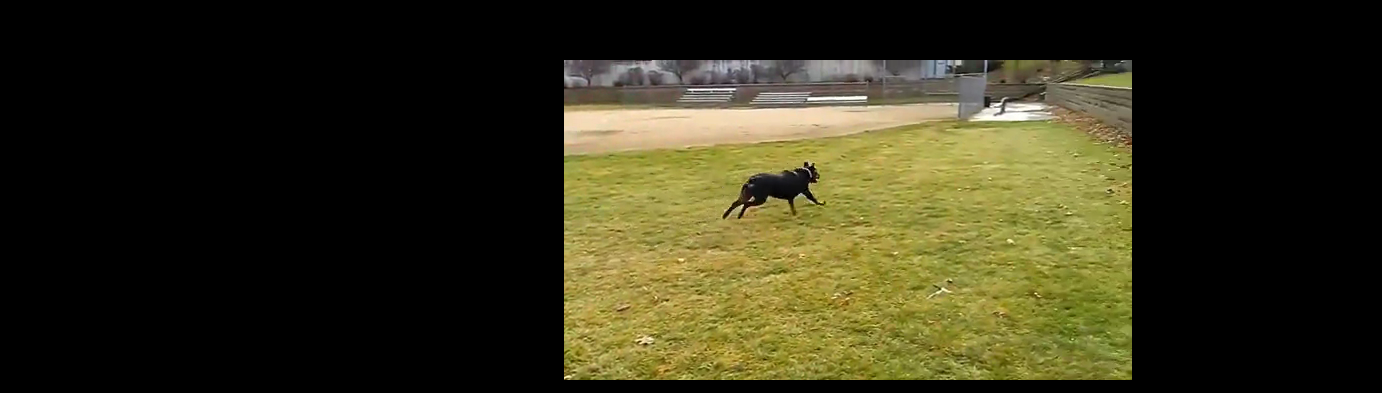}\\
			\includegraphics[width=0.5\textwidth]{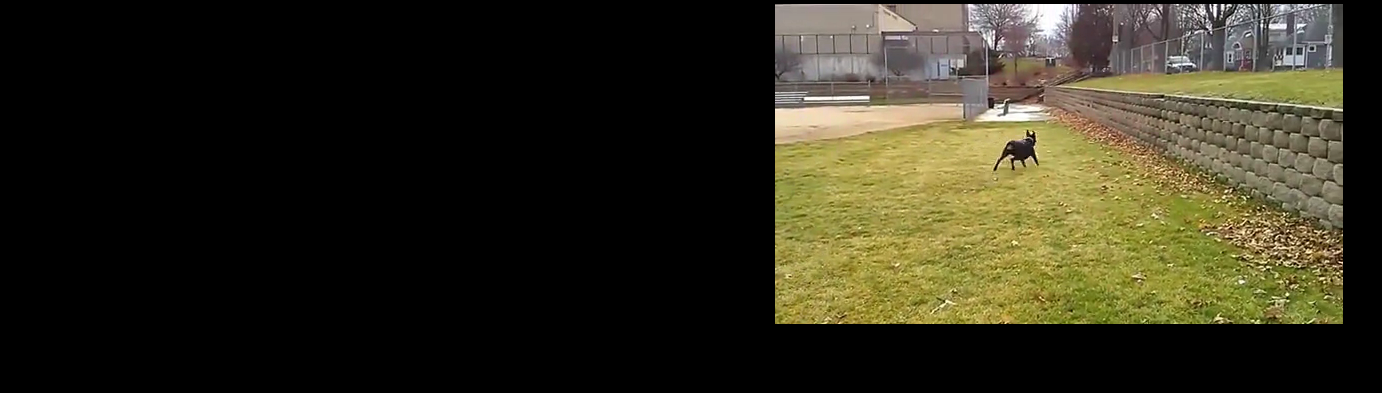}\\
			\includegraphics[width=0.5\textwidth]{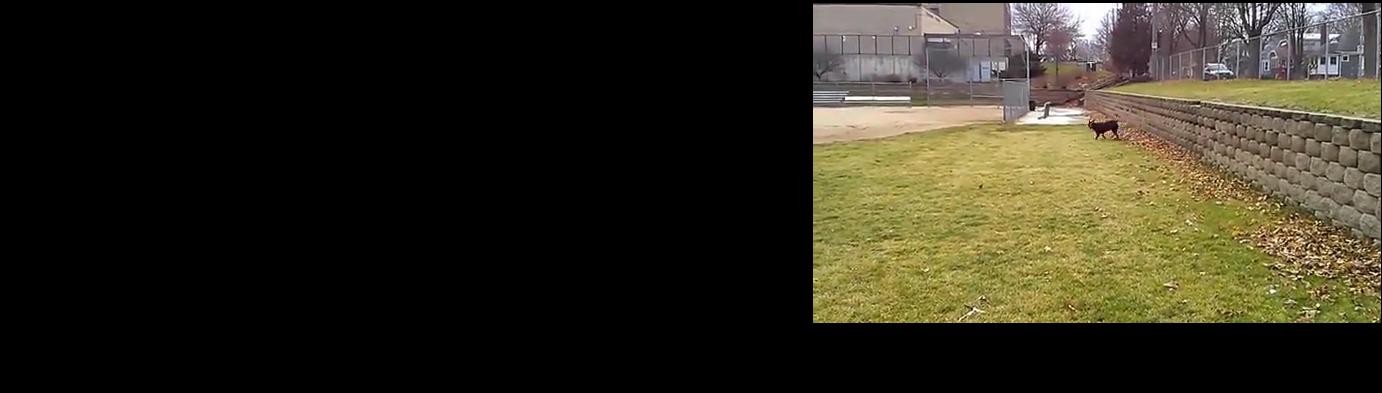}\\
			(b) Camera motion stabilization can reestablish object motion.
		\end{tabular}
	\end{center}
	\caption[Camera motion compensation via stabilization]{Camera motion compensation via stabilization. (a) Select frames demonstrating an object-centric stabilized video. The object of interest (dog) appears near the centre of the frame across the image sequence even though it is locomoting in the real-world.   (b) Corresponding frames with camera motion stabilization. Camera motion that the camera operator has imposed to centre the object of interest within the field of view (i.e., pan) is undone, such that features extracted from the camera motion stabilized video is more representative of biolocomotion in the real-world. \label{fig:cam_stab}}
\end{figure}
\restoregeometry

\end{subsubsection}
\end{subsection}

\begin{subsection}{Biolocomotion detector}\label{sec:tech_algorithm}
A collection of tracked point trajectories are used as the input to the proposed algorithm to analyze motion in terms of (i) locomotion, (ii) oscillation, (iii) asymmetry, and (iv) extrinsic motion dissimilarity\footnotemark; see Figure~\ref{fig:biolocomotion_breakdown}. 
\footnotetext{For the sake of compactness, the dissimilarity between extrinsic and intrinsic motion will be referred to as \textbf{extrinsic motion dissimilarity} in the following.}
In this section, approaches to quantifying these components will be defined, followed by a way to combine these measures into a single biolocomotion detector. Throughout the section, each component will be defined while alluding to the simple real-life example in Figure~\ref{fig:toy_example} that variously exhibits locomotion without oscillation (toy car), oscillation without locomotion (pendulum), symmetric oscillation with locomotion (rolling ball), asymmetric oscillation with coinciding extrinsic and intrinsic motions (bouncing ball), and asymmetric oscillation with dissimilar extrinsic and intrinsic motions (person). 

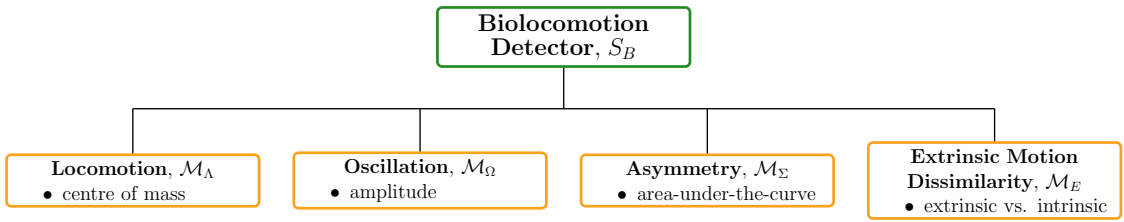
\begin{figure}[!h]
	\begin{center}
	\resizebox{\textwidth}{!}{
		\begin{tikzpicture} 
	
	\node[draw=ForestGreen, text width=5cm, align=center, rounded corners=1mm, ultra thick] (rect) at (9,3) (Eb) {\large{\textbf{Biolocomotion Detector}}, $S_B$};
	
	\node[draw=YellowOrange, text width=5cm, align=center, rounded corners=1mm, ultra thick] (rect) at (0,0) (Eloco) {\textbf{Locomotion}, $\mathcal{M}_\Lambda$ \begin{itemize}[topsep=0pt] \item centre of mass \end{itemize}};
	
	\node[draw=YellowOrange, text width=5cm, align=center, rounded corners=1mm, ultra thick] (rect) at (6,0) (Eosc) {\textbf{Oscillation}, $\mathcal{M}_\Omega$ \begin{itemize}[topsep=0pt] \item amplitude \end{itemize}};
	
	\node[draw=YellowOrange, text width=5cm, align=center, rounded corners=1mm, ultra thick] (rect) at (12,0) (Easym) {\textbf{Asymmetry}, $\mathcal{M}_\Sigma$ \begin{itemize}[topsep=0pt] \item area-under-the-curve \end{itemize}};
	
	\node[draw=YellowOrange, text width=5cm, align=center, rounded corners=1mm, ultra thick] (rect) at (18,0) (Elocal) {\textbf{Extrinsic Motion Dissimilarity}, $\mathcal{M}_E$ \begin{itemize}[topsep=0pt] \item extrinsic vs. intrinsic \end{itemize}};
	
	\draw[-,thick] (0,1.5) -- (18,1.5); 
	\draw[-,thick] (Eb) -- (9,1.5);
	\draw[-,thick] (0,1.5) -- (Eloco);
	\draw[-,thick] (6,1.5) -- (Eosc);
	\draw[-,thick] (12,1.5) -- (Easym);
	\draw[-,thick] (18,1.5) -- (Elocal);
		
\end{tikzpicture}
		}
	\end{center}
	\caption[Developed biolocomotion detector component composition]{Composition of components of the developed biolocomotion detector. The detector consists of components that measure locomotion, $\mathcal{M}_\Lambda$, oscillation, $\mathcal{M}_\Omega$, asymmetry, $\mathcal{M}_\Sigma$, and extrinsic motion dissimilarity, $\mathcal{M}_E$. \textit{Locomotion} is measured based on the aggregate displacement of the centre of mass, \textit{oscillation} and \textit{asymmetry} are calculated based on amplitude and area-under-the-curve of the trajectories, respectively, and \textit{extrinsic motion dissimilarity} is measured based on the dissimilarity between extrinsic and intrinsic displacement vectors. \label{fig:biolocomotion_breakdown}}
\end{figure}

\begin{landscape}
	\begin{figure}
		\begin{center}
			\resizebox{1.2\textwidth}{!}{
			\begin{tabular}{c c c c}
				\multicolumn{4}{c}{\includegraphics[width=1.6\paperwidth]{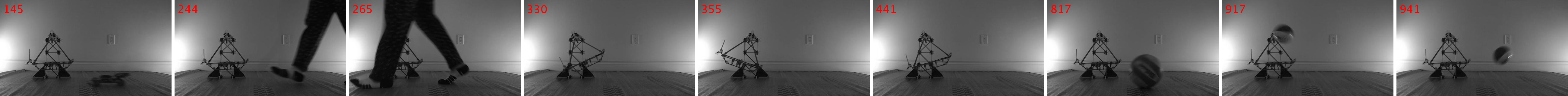}} \\
				\multicolumn{4}{c}{(a) Select frames \vspace{3mm}} \\
				\multicolumn{4}{c}{\includegraphics[width=1.6\paperwidth]{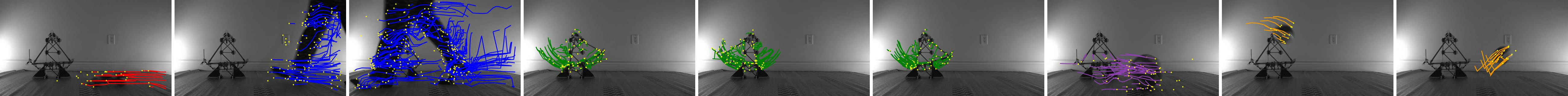}} \\
				\multicolumn{4}{c}{(b) Tracked points and their trajectories \vspace{3mm}} \\
				\includegraphics[width=0.5\textwidth]{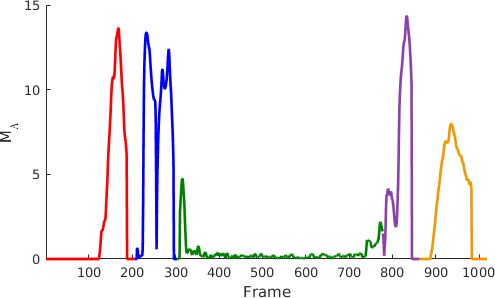}
				& \includegraphics[width=0.5\textwidth]{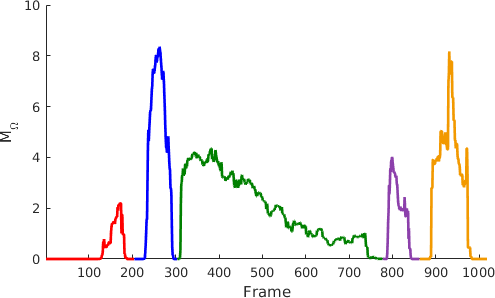} 
				& \includegraphics[width=0.5\textwidth]{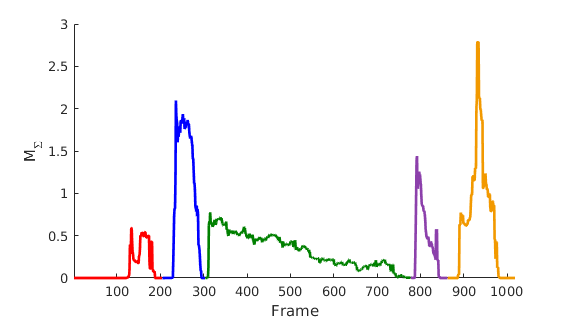} 
				& \includegraphics[width=0.5\textwidth]{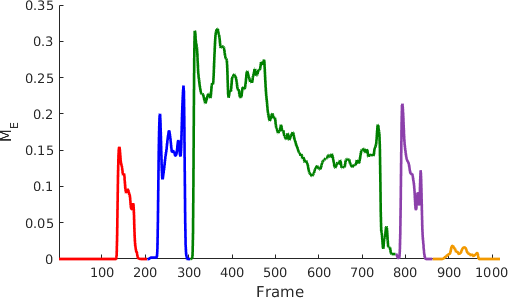} \\
				(c) Measure of Locomotion, $\mathcal{M}_\Lambda$ \vspace{3mm}
				& (d) Measure of Oscillation, $\mathcal{M}_\Omega$ \vspace{3mm}		
				& (e) Measure of Asymmetry, $\mathcal{M}_\Sigma$ \vspace{3mm}
				& (f) Measure of Extrinsic Motion Dissimilarity, $\mathcal{M}_E$ \vspace{3mm} \\
				\multicolumn{4}{c}{\includegraphics[width=0.5\textwidth]{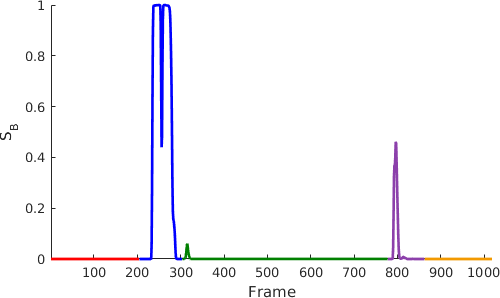}} \\
				\multicolumn{4}{c}{(g) Measure of Biolocomotion, $S_B$} \\
			\end{tabular}
			}
		\end{center}
	\caption[A simple real-world illustration of biolocomotion analysis]{A simple real-world illustration of biolocomotion analysis. 
	(a) Select frames demonstrating non-biological objects in locomotion (toy car at frame 145, rolling ball at frame 817, and bouncing ball at frames 917 and 941), a biological object in locomotion (person at frames 244 and 265), and a non-biological object in oscillation (pendulum at frames 330, 355 and 441). 
	(b) Tracked points (yellow) and their trajectories overlaid with red, blue, green, purple, and orange indicating the toy car, person, pendulum, rolling ball, and bouncing ball, respectively. 
	(c-g) Recovered measures of locomotion ($\mathcal{M}_\Lambda$), oscillation ($\mathcal{M}_\Omega$), asymmetry ($\mathcal{M}_\Sigma$), extrinsic motion dissimilarity ($\mathcal{M}_E$), and biolocomotion ($S_B$) with respect to time (in frames). \label{fig:toy_example}}
	\end{figure}
\end{landscape}

\begin{subsubsection}{Measure of Locomotion, $\mathcal{M}_\Lambda$}\label{sec:E_loc}
	The amount an object displaces from one location to another in a given temporal window is quantified using a set of trajectories by calculating the \textbf{measure of locomotion}, $\mathcal{M}_\Lambda$; see blue curves in Figure~\ref{fig:teaser}a,b or lack thereof in Figure~\ref{fig:teaser}c. The measure of locomotion can be evaluated by considering the norm (e.g., $L_1$, $L_2$, etc.) of the average displacements of the object over some temporal window. In the following, a measure of locomotion is described.\\

\begin{figure}[!h]
	\begin{center}
		\begin{tabular}{c c c}
			\includegraphics[width=0.3\textwidth]{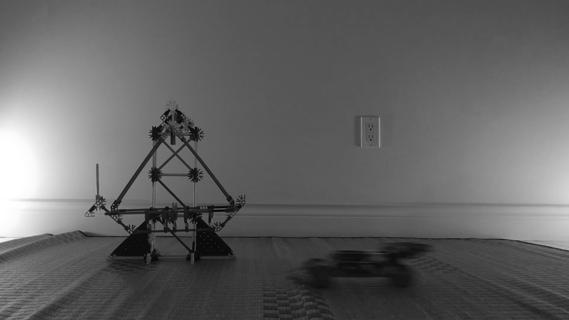}
			& \includegraphics[width=0.3\textwidth]{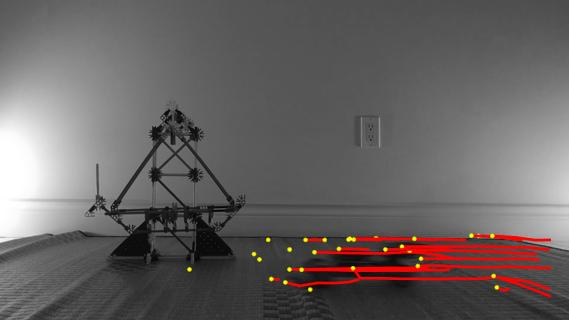}
			& \includegraphics[width=0.3\textwidth]{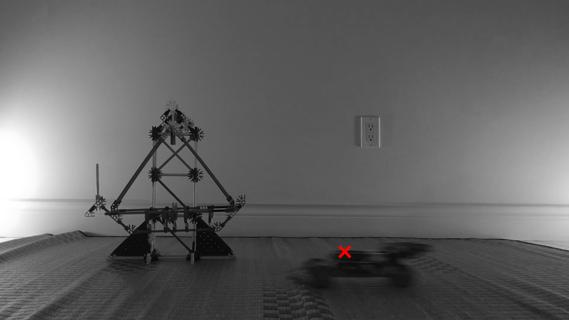} \\
			(a) Original Frame 
			& (b) Recovered Trajectories
			& (c) Centroid \\
		\end{tabular}
	\end{center}
	\caption[Centroid of points at frame $t$]{Centroid of points at frame $t$. (a) Input video at frame $t$. (b) Points present at frame $t$ (yellow dots) can be extracted from a set of trajectories that are present at frame $t$ (red lines). (c) The centroid (red cross) can be obtained by computing the trimmed mean of the points that are present at frame $t$ (yellow dots in (b)). \label{fig:centroidsATt}}
\end{figure}

The displacement of an object is measured by considering the displacements of the centroid of an object over some temporal window; see Figure \ref{fig:centroidsATt}. The \textbf{centroid of an object} at frame $t$, $\bar{\mathbf{x}}^c(t)$, is calculated by considering the trimmed mean of points that are present at frame $t$ for $1 \leq t \leq T$ according to
\begin{equation}\label{eq:centroid}
	\bar{\mathbf{x}}^c(t) 
	= \left( \bar{x}^c(t), \bar{y}^c(t) \right)
	= \left( \frac{1}{|X(t)|}\sum_{x^k(t)\in{X(t)}}{x^k(t)}, \frac{1}{|Y(t)|}\sum_{y^k(t)\in{Y(t)}}{y^k(t)} \right)
	\text{,}
\end{equation}
where $X(t)$ and $Y(t)$ are sets of points measured along the horizontal and vertical axes, respectively, present at frame $t$ within 1.5 of the standard deviation from its respective mean. That is, suppose $\mu_x(t)$ and $\sigma_x(t)$ denotes the mean and the standard deviation, respectively, of trajectory points present at frame $t$ measured along the horizontal axis. Then a set of points, $X(t)$, along the horizontal axis at frame $t$ is defined as
\begin{equation}\label{eq:setXatT}
	X(t)=\{x^k(t) \ | \ \mu_x(t) - 1.5\sigma_x(t) \leq x^k(t) \leq \mu_x(t) + 1.5\sigma_x(t)\} \text{.}
\end{equation}
To ensure that reliable data contributes to the calculation of $\bar{x}^c(t)$, only data within 1.5 of the standard deviation from the mean are considered. Discarding data that are outside 1.5 of the standard deviation from the mean ensures the final measurement, $\bar{x}^c(t)$, is based on approximately 86.67\% of the total data while the lowest 6.68\% and highest 6.68\%, that may correspond to noise, are discarded. \\

Similarly, a set of points, $Y(t)$, along the vertical axis at frame $t$ is defined as
\begin{equation}\label{eq:setYatT}
	Y(t)=\{y^k(t) \ | \ \mu_y(t) - 1.5\sigma_y(t) \leq y^k(t) \leq \mu_y(t) + 1.5\sigma_y(t)\}
\end{equation}
for mean and standard deviation, $\mu_y(t)$ and $\sigma_y(t)$, respectively, of trajectory points present at frame $t$ along the vertical axis. \\

Consequently, the \textbf{centroid displacement vector} at frame $t$ from a previous frame at $t-\Delta t$ is defined as
\begin{align}\label{eq:centroid_disp}
	\Delta \bar{\mathbf{x}}^c_{(t,\Delta t)}
	&= \bar{\mathbf{x}}^c(t) - \bar{\mathbf{x}}^c(t - \Delta t) 
	\text{.}
\end{align} 
Note that $\Delta \bar{\mathbf{x}}^c_{(t,\Delta t)}$ is similar to $\Delta \bar{\mathbf{x}}^k_{(t,\Delta t)}$ for trajectory $\mathcal{T}^k$ \eqref{eq:traj_disp}, except $\Delta \bar{\mathbf{x}}^c_{(t,\Delta t)}$ is specific for centroids. \\

Finally, the measure of locomotion is obtained by combining the overall horizontal and vertical displacements of the centroids over a temporal window. Mathematically, it is defined as
\begin{equation}\label{eq:Elambda}
	\mathcal{M}_{\Lambda}(t) \equiv 
	\sqrt{\left(\frac{1}{\min{ \{\omega,t\} }} \sum_{n=\max{\{t-\omega,1\}}}^{t}{\Delta \bar{x}^c(n)} \right)^2 
	+ \left(\frac{1}{\min{ \{\omega,t\} }} \sum_{n=\max{\{t-\omega,1\}}}^{t}{\Delta \bar{y}^c(n)} \right)^2}
	\text{,}
\end{equation}
for some constant $\omega>0$, which represents the length of the temporal window. Empirically, $\omega$ is set to the frame rate of the video. 
Since the goal is to quantify displacement rather than distance, framewise displacements are aggregated (without considering their absolute values) before overall magnitude is calculated. \\

It can be observed in Figure~\ref{fig:toy_example}c that as the toy car, person, and the balls make a spatial advance, their locomotion measurements (red, blue, purple, and orange, respectively) are large, while the pendulum (green) has a low locomotion measure since its overall displacement is zero. \\

\end{subsubsection}

\begin{subsubsection}{Measure of Oscillation, $\mathcal{M}_\Omega$}\label{sec:E_osc}
	The amount of oscillation exhibited by an object can be quantitatively measured by considering the behaviour of the entire trajectory (as opposed to the centroid of the tracked points). In particular, the oscillatory behaviour induced by the motion of a localized body part (e.g., a limb) can be captured by considering its amplitude, where large amplitudes generally provide a stronger evidence of oscillation; cf. locomoting biological species and locomoting non-biological objects in Figure~\ref{fig:teaser}. Thus, to obtain a \textbf{measure of oscillation}, $\mathcal{M}_\Omega$, an aggregate of trajectory amplitudes are considered. \\

To calculate the amplitude, $a^k$, of trajectory $\mathcal{T}^k$, the trajectory must be detrended to $\tilde{\mathcal{T}}^k$ such that the calculated amplitude is independent of the spatial direction the trajectory unfolds over time; see Figure~\ref{fig:detrendTraj}. Once the trajectory of interest has been detrended, its amplitude can be computed in various ways: (i) frequencies, (ii) integrals, or (iii) differentials \cite{Karra10}. Based on preliminary experimentation, a frequency-based approach is employed; see Appendix~\ref{app:amp} for further discussion of integral- and differential-based approaches. 
The amplitude of a detrended trajectory, $\tilde{\mathcal{T}}^k$, can be calculated using \textit{frequencies} by obtaining the maximum magnitude of the Fourier Transformation of $\tilde{\mathcal{T}}^k$. That is, 
\begin{equation}
	a_{\mathcal{F}}^k = \max{ \left\{ \mathcal{F}(\tilde{\mathcal{T}}^k) \right\} } \text{,}
\end{equation}
where $\mathcal{F}(\cdot)$ denotes the Fourier Transform.
One maximum value is sufficient since considered trajectories only span a few (i.e., 15) frames to compensate for the drifting effect. \\

With the amplitude estimated for each detrended trajectory, the overall measure of oscillation at frame $t$, $\mathcal{M}_\Omega(t)$, is calculated by considering the weighted mean of the amplitudes. That is, for a detrended trajectory $\mathcal{\tilde{T}}^k(t)$, the measurement of oscillation is defined as
	\begin{equation}\label{eq:Eomega}
		\mathcal{M}_\Omega(t) \equiv \frac{w_{\nu}(t)}{|A(t)|} \sum_{a^k(t) \in A(t)}{a^k(t)} \text{,}
	\end{equation}
	where $a^k$ is the amplitude of trajectory $\tilde{\mathcal{T}}^k$, $A(t)$ is a set of amplitudes considered at frame $t$, $|A(t)|$ is the number of amplitudes considered at frame $t$, and $w_{\nu}(t)$ is a weight assigned to account for the percentage of total trajectories present in a given frame, as follows.\\

To ensure that reliable data contributes to the calculation of $\mathcal{M}_\Omega(t)$, only amplitudes within 1.5 of the standard deviation from the mean are considered in $A(t)$, analogous to the computation of centroid of an object, $\mathbf{x}^c(t)$, as in \eqref{eq:centroid}. That is, 
\begin{equation}\label{eq:trimmedSet}
	A(t) = \{ a^k(t) \ | \ \mu_a(t)-1.5\sigma_a(t) \leq a^k(t) \leq \mu_a(t)+1.5\sigma_a(t) \} \text{,}
\end{equation}
where $\mu_a(t)$ and $\sigma_a(t)$ are the mean and standard deviation, respectively, of the amplitudes present at frame $t$. \\

It is desirable to avoid having frames with exceptionally small number of trajectories unduly bias the overall measure of oscillation, $\mathcal{M}_\Omega(t)$. Thus, $\mathcal{M}_\Omega$ is weighted by $w_\nu(t)$, which is defined based on the amount of data present at frame $t$, $|A(t)|$, relative to the typical amount of data that is available in a frame across the entire video. A sigmoid function can be used to model a fair distribution based on the amount of data according to
\begin{equation}\label{eq:w_nu}
	w_\nu(t) = \frac{1}{1+\exp{\left[ -(|A(t)| - |\mu_\nu - \sigma_\nu|) \right]}} \text{,}
\end{equation}
where $\mu_\nu$ and $\sigma_\nu$ are the mean and standard deviation, respectively, of the amount of data in a given set. Thus, a unit weight is assigned when there are more trajectories present at frame $t$ relative to the average across the video, since it suggests that the frame contains a reliable amount of data. On the contrary, frames with exceptionally small number of trajectories are assigned a low weight through the sigmoid's saturation to prevent a few exceptional data points from dominating the final outcome of the overall measure, $\mathcal{M}_\Omega$. \\
	
It can be observed in Figure~\ref{fig:toy_example}d that the oscillation measure, $\mathcal{M}_\Omega$, captures the amount of oscillation that is exhibited by each object in motion. Specifically, the toy car with near linear trajectories (red) has very low $\mathcal{M}_\Omega$ values, while the person (blue), pendulum (green), and the rolling and bouncing balls (purple and orange, respectively) have relatively large $\mathcal{M}_\Omega$ values. Furthermore, as the pendulum slows down to a gradual stop, its oscillation values also approach zero.
\end{subsubsection}

\begin{subsubsection}{Measure of Asymmetry, $\mathcal{M}_\Sigma$}\label{sec:E_asym}

The resistive forces that species must battle as they move through an environment can be captured through an asymmetric motion trace these species exhibit during their path of motion (cf. locomoting biological species and rolling ball in Figure~\ref{fig:teaser}); this amount is quantitatively measured by considering the behaviour of the entire trajectory (similar to the measure of oscillation). In particular, the resistive force that must be battled can be calculated by measuring the magnitude of asymmetry of the individual trajectories, the \textbf{measure of asymmetry}, $\mathcal{M}_\Sigma$, where presence of asymmetry is more indicative of biolocomotion. \\

A variety of approaches can be employed to measure asymmetry: (i) direct calculation of vertical acceleration, (ii) skewness \cite{Borowski91} of the trajectory, and (iii) comparison of the left and right areas under the trajectories from its highest peak. Preliminary investigations showed difficulties in reliably estimating asymmetry via direct calculation of vertical acceleration, by considering the second-derivative of the vertical components of the trajectories, as computational instability was encountered and amplified for each considered derivative. Similar challenges appeared in estimating asymmetry via skewness, as the third-order moment statistic was very sensitive to the presence of noise in the trajectories. Thus, the magnitude of asymmetry, $s^k$, of a detrended trajectory, $\mathcal{\tilde{T}}^k$, is quantified by finding its local maximum valued point and comparing the integrals from that value to its left and right minimal valued points. \\

Since the recovered trajectories traces the path of the object from frame $t$ to $t+L^k$, the recovered trajectory is not guaranteed to contain a \textit{full cycle} of the tracked motion from the lowest point to its next lowest point (e.g., stance-swing-stance phases of a walk). Rather, it is likely to track a cycle of the motion at random points (e.g., latter half of swing-stance-former half of swing phases of a walk). To increase the likelihood that the considered trajectory will always contain the desired information when computing the area-under-the-curve (e.g., stance-swing-stance), the input trajectories are replicated. Subsequently, any part of the replicated trajectory is representative of the stance-swing-stance cycle; therefore, it can be used to extract asymmetry information. 
Furthermore, the detrended trajectory is vertically shifted such that its vertical minimum is zero to ensure the calculated area is non-negative. 
Then the  integrals from the (local) maximum valued point is compared to its left and right minimal valued points. More precisely, let 
\begin{equation}
	l_L = \arg \min{\tilde{y}^k}
	\text{ and }
	l_R = \arg \min{\tilde{y}^k}
\end{equation}
be the left and right minimal points of $\tilde{\mathcal{T}}^k$, respectively, for 
\begin{equation}
	l_{\max}=\arg \max \tilde{y}^k
\end{equation}
such that $l_L < l_{\max} < l_R$; see Figure \ref{fig:traj_asym}. Then, the asymmetry magnitude of $\tilde{\mathcal{T}}^k$ is computed by comparing the discrete approximation of the integrals from the left and right of the highest peak as
\begin{equation}
	s^k = \left| \sum_{l=l_L}^{l_{\max}}{\left[\tilde{y}^k_l+\min{\tilde{y}^k}\right]} - \sum_{l=l_{\max}}^{l_R}{\left[\tilde{y}^k_l+\min{\tilde{y}^k}\right]} \right| \text{,}
\end{equation} 
where $\min{\tilde{y}^k}$ is the vertical minimum of $\mathcal{\tilde{T}}^k$ and serves to shift vertically the $\tilde{y}^k$, so that the calculated values are non-negative. A large asymmetric magnitude, $s^k$, is indicative of biolocomotion.\\

\begin{figure}[H]
	\begin{center}
		\begin{tikzpicture}
	\draw [<->] (-1, 0) -- (8,0); 
	\draw [<->] (0,-1) -- (0, 3); 
	
	\foreach \x in {0.1, 0.4, ..., 3.7}{
		\filldraw[fill=YellowOrange] (\x,0) -- (\x,2*sin{deg(pi/8*\x)}) -- (\x+0.2,2*sin{deg(pi/8*\x)}) -- (\x+0.2,0) -- cycle; 
	}
	\foreach \x in {4.0, 4.3, ..., 6}{
		\filldraw[fill=ForestGreen] (\x,0) -- (\x,2*sin{deg(pi/4*(\x-2))}) -- (\x+0.2,2*sin{deg(pi/4*(\x-2))}) -- (\x+0.2,0) -- cycle; 
	}
	
	\draw[red, ultra thick, -] (0,0) .. controls (4.3,2.8) .. (6.1,0) .. controls (6.5,0.05) .. (7,0.2);
	\draw[red, ultra thick,-] (-1,0.5) .. controls (-0.5,0.1) .. (0,0);
	\draw[red] (5,2) circle node[right] {$\tilde{\mathcal{T}}^k + \min{\tilde{y}^k}$};
	
	\filldraw[red] (3.9,2.1) circle (2pt); 
	\filldraw[red] (0,0) circle (2pt); 
	\filldraw[red] (6.1,0) circle (2pt); 
	\draw (0,0) circle node[below right] {$x_{l_L}$};
	\draw (3.9,0) circle node[below] {$x_{l_{max}}$};
	\draw (6.1,0) node[below] {$x_{l_R}$}; 
	\draw (3.9,-0.1) -- (3.9,0.1);

	\draw (-0.1,2.1) -- (0.1,2.1);
	\draw (0,2.1) circle node[left] {$\max(\tilde{y}^k) + \min(\tilde{y}^k)$};
\end{tikzpicture}
	\end{center}
	\caption[Calculation of the magnitude of asymmetry]{Calculation of magnitude of asymmetry. The magnitude of asymmetry of a detrended trajectory $\tilde{\mathcal{T}}^k$ can be computed by comparing the discrete approximation of integrals from the left of the highest peak (orange) and right of the highest peak (green) of a vertically shifted detrended trajectory such that its vertical minimum is 0. \label{fig:traj_asym}}
\end{figure}
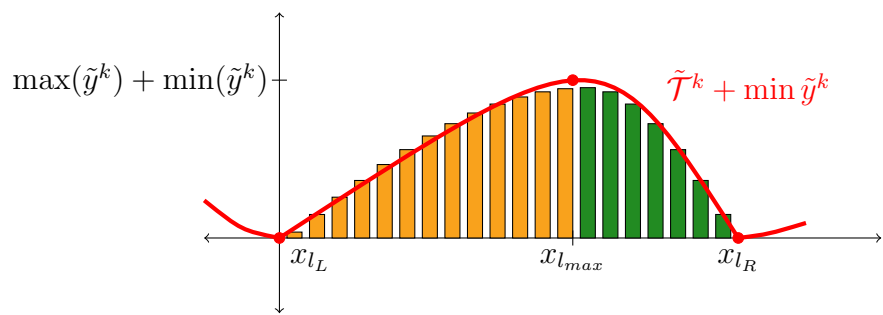

To ensure that the measure of asymmetry is invariant to the stride length of the object, $s^k$ is normalized by the arc length of $\tilde{\mathcal{T}}^k$, $\| \tilde{\mathcal{T}}^k \|$ as in \eqref{eq:arc_length}, as
\begin{equation}
	\tilde{s}^k = \frac{s^k}{\|\tilde{\mathcal{T}}^k\|} \text{.} 
\end{equation} 

While normalizing the amplitude by the arc length could also make the measure of oscillation, \eqref{eq:Eomega}, invariant to the stride length of the object's motion, such normalization retracts informative data. Loss of information when normalizing amplitude by arc length can be understood through an analogy to various parts of a circle. That is, suppose the radius of a circle is to the amplitude of a detrended trajectory as circumference is to arc length, and area is to area-under-the-curve. Furthermore, let the radius of a circle be $r$ such that its circumference is $2\pi r$ and area is $\pi r^2$. Then normalizing the radius (or amplitude) by the circumference (or arc length) yields 
\begin{equation}
	\frac{r}{2\pi r} = \frac{1}{2\pi} \text{,}
\end{equation}
where informative data, $r$, is lost.
Normalizing the area (or area-under-the-curve) by the circumference (or arc length), however, yields
\begin{equation}
	\frac{\pi r^2}{2 \pi r} = \frac{r}{2} \text{,}
\end{equation}
where informative data $r$ is multiplied with a constant.\\

Similar to the measure of oscillation, the measure of asymmetry at frame $t$, $\mathcal{M}_\Sigma(t)$, is obtained by aggregating the asymmetry magnitudes. Specifically, it is calculated by considering the weighted mean of the asymmetry magnitudes according to
\begin{equation}
	\mathcal{M}_\Sigma(t) \equiv \frac{\ w_\nu(t)}{|S(t)|} \sum_{\tilde{s}^k \in S(t)}{\tilde{s}^k} \text{,}
\end{equation}
where $\tilde{s}^k$ is the normalized magnitude of asymmetry of $\tilde{\mathcal{T}}^k$, $S(t)$ is a trimmed set of asymmetric values that are present at frame $t$ analogous to \eqref{eq:trimmedSet}, $|S(t)|$ is its cardinality, and $w_\nu$ is a weight that accounts for the percent of total asymmetry values considered in a given frame exactly analogous to the weights applied to the aggregation of oscillation measures as in \eqref{eq:w_nu}. \\

It can be observed in Figure~\ref{fig:toy_example}e that the person (blue) and bouncing ball (orange) exhibit large asymmetry measures, $\mathcal{M}_\Sigma$, as they have to fight against gravity during their lift phase, while the car (red), pendulum (green), and rolling ball (purple) do not. As a result, $\mathcal{M}_\Sigma$ is lower for the toy car, pendulum, and rolling ball.
\end{subsubsection}

\begin{subsubsection}{Measure of Extrinsic Motion Dissimilarity, $\mathcal{M}_E$}\label{sec:E_ext}
The biolocomotion detector can benefit from emphasizing the dissimilarity between the overall path of the object (extrinsic motion) and the trace exhibited by the individual body parts (intrinsic motion)\footnotemark; cf. locomoting biological objects and bouncing ball in Figure~\ref{fig:teaser}. \\
\footnotetext{In the psychophysical literature, \textit{intrinsic motion} is obtained by removing \textit{common (extrinsic) motion} from the observed motion, via subtraction \cite{Chang09}. However, such subtraction limits the ability to compare common and relative motion. Thus, the individual traces exhibited by the body parts are used without the subtraction of common (extrinsic) motion for such comparison in this report.}


The amount of deviation between extrinsic and intrinsic motion exhibited by an object, the \textbf{measure of extrinsic motion dissimilarity}, $\mathcal{M}_E$, is measured by comparing the dissimilarity between the overall path of the object (i.e., centroid) and the individual trajectory shape. In particular, the angle between the displacement vectors of a (non-detrended) trajectory $\mathcal{T}^k$ and the centroid are compared. Suppose $\Delta \mathbf{x}^k_{(t_{ext},\Delta t_{ext})}$ denotes a displacement vector of trajectory $\mathcal{T}^k$ between frames $t_{ext} - \Delta t_{ext}$ and $t_{ext}$ as in \eqref{eq:traj_disp}, while $\Delta \mathbf{x}^c_{(t_{ext},\Delta t_{ext})}$ denotes a displacement vector of the centroid between frames $t_{ext} - \Delta t_{ext}$ and $t_{ext}$ as in \eqref{eq:centroid_disp}, where $t_{ext}$ is a frame at which a vertical extremum is reached by $\mathcal{T}^k$ and $\Delta t_{ext}$ is the temporal separation between vertical extrema; see Figure~\ref{fig:traj_extrinsic}. Then the extrinsic motion dissimilarity at frame $t_{ext}$ is calculated using the normalized inner product between extrinsic and intrinsic displacement vectors from frame $t_{ext}-\Delta t_{ext}$ to $t_{ext}$ according to
\begin{align}\label{eq:ext_int_disp_vec}
\begin{split}
e^k_{(t_{ext},\Delta t_{ext})}
&= 1 - \frac{1}{2}\left( {\frac{\langle \Delta \mathbf{x}^k_{(t_{ext},\Delta t_{ext})},\Delta \mathbf{x}^c_{(t_{ext},\Delta t_{ext})}\rangle}
				   	    {\|\Delta \mathbf{x}^k_{(t_{ext},\Delta t_{ext})}\| \|\Delta \mathbf{x}^c_{(t_{ext},\Delta t_{ext})}\|}} +1\right)  \\
&= \frac{1}{2}\left( 1 - {\frac{\langle \Delta \mathbf{x}^k_{(t_{ext},\Delta t_{ext})},\Delta \mathbf{x}^c_{(t_{ext},\Delta t_{ext})}\rangle}
				   	    {\|\Delta \mathbf{x}^k_{(t_{ext},\Delta t_{ext})}\| \|\Delta \mathbf{x}^c_{(t_{ext},\Delta t_{ext})}\|}} \right) \text{.}
\end{split}
\end{align}
\indent
The normalized inner product in \eqref{eq:ext_int_disp_vec} is increased by $1$ then multiplied by a factor of $\frac{1}{2}$ to ensure the considered inner product ranges between $0$ and $1$ instead of $[-1,1]$. 
The range is altered via shift followed by a multiplicative constant instead of considering the absolute of the normalized inner product for the following reason. To reduce angular momentum of the body during locomotion, animals often accompany their (extrinsic) overall positional advance with (intrinsic) motion of body parts in the opposite direction (e.g., backward swing of arms during walking). Correspondingly, for present purposes, it is beneficial that antiparallel extrinsic vs. intrinsic motion be distinguished from parallel. Simply taking the absolute value of the normalized inner product fails to make this distinction, while formula \eqref{eq:ext_int_disp_vec} does. 
Further, the normalized inner product is subtracted from 1 to ensure a large corresponding value is indicative of biolocomotion, similar to other computed measures for biolocomotion (e.g., large amplitude values are indicative of biolocomotion). 
Finally, the measure of extrinsic motion dissimilarity for trajectory $\mathcal{T}^k$ is obtained by considering the weighted combination of all normalized inner products of extrinsic and intrinsic displacement vectors according to
\begin{equation}
	\displaystyle
	e^k = \frac{\sum_{t_{ext}}{\Delta t_{ext} \ e^k_{(t_{ext},\Delta t_{ext})}}}
			  {\sum_{t_{ext}}{\Delta t_{ext}}} \text{.}
\end{equation}
The temporal separation between two extreme points, $\Delta t_{ext}$, is used as a weight, such that the difference between the considered extrinsic and intrinsic displacements are maximized. For example, comparing the extrinsic and intrinsic displacement vectors with $\Delta t = 1$ is likely to be more alike than vectors with $\Delta t = 15$ since the centroid is calculated from the motion of the object. 
This manipulation is analogous to discrete approximation for estimating the derivative of a mathematical function. A more accurate estimation of the slope is obtained when the discretization ($\Delta t$) approaches zero. Similarly, extrinsic and intrinsic displacement vectors will be more alike as $\Delta t \rightarrow 0$, while the goal is to determine how different they are.\\

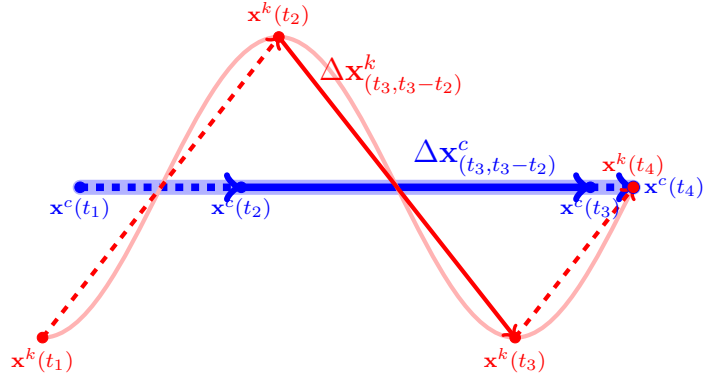
\begin{figure}[!h]
	\begin{center}
		\begin{tikzpicture}

	\draw[blue!30,line width=2mm, cap=round] (0.5,0) -- (5*pi/2,0); 
	\draw[blue, line width=1mm, ->, dashed] (0.5,0) -- (pi-0.5,0); 
	\draw[blue, line width=1mm, ->] (pi-0.5,0) -- (2*pi+1,0); 
	\draw[blue, line width=1mm, ->, dashed] (2*pi+1,0) -- (5*pi/2,0); 
	
	\draw[red!30,ultra thick] (0,-2) cos (pi/2,0) sin (pi,2) cos (3*pi/2,0) sin (2*pi,-2) cos (5*pi/2,0);
	\draw[red, ultra thick, ->, dashed] (0,-2) -- (pi,2); 
	\draw[red, ultra thick, ->] (pi,2) -- (2*pi,-2); 
	\draw[red, ultra thick, ->, dashed] (2*pi,-2) -- (5*pi/2,0); 
	
	\draw[blue] (15*pi/8,0) node[above] {$\Delta \mathbf{x}^c_{(t_3,t_3-t_2)}$}; 
	\draw[red] (9*pi/8,1.5) node[right] {$\Delta \mathbf{x}^k_{(t_3,t_3-t_2)}$}; 

	\filldraw[blue] (0.5,0) circle(2pt) node[below] {\scriptsize{$\mathbf{x}^c(t_1)$}};
	\filldraw[blue] (pi-0.5,0) circle(2pt) node[below] {\scriptsize{$\mathbf{x}^c(t_2)$}};
	\filldraw[blue] (2*pi+1,0) circle(2pt) node[below] {\scriptsize{$\mathbf{x}^c(t_3)$}};
	\filldraw[blue] (5*pi/2,0) circle(2.3pt) node[right] {\scriptsize{$\mathbf{x}^c(t_4)$}};
	
	\filldraw[red] (0,-2) circle (2pt) node[below] {\scriptsize{$\mathbf{x}^k(t_1)$}};
	\filldraw[red] (pi,2) circle (2pt) node[above] {\scriptsize{$\mathbf{x}^k(t_2)$}};
	\filldraw[red] (2*pi,-2) circle (2pt) node[below] {\scriptsize{$\mathbf{x}^k(t_3)$}};
	\filldraw[red] (5*pi/2,0) circle (2pt) node[above] {\scriptsize{$\mathbf{x}^k(t_4)$}};
	
\end{tikzpicture}
	\end{center}
	\caption[Calculation of extrinsic and intrinsic motion dissimilarity]{Calculation of extrinsic and intrinsic motion dissimilarity. Given trajectory $\mathcal{T}^k$ (light red curve) from frame $t_1$ to $t_4$, where vertical extreme points, $\mathbf{x}(t_1), \dots, \mathbf{x}(t_4)$ (red dots), are achieved at frames $t_1, \dots, t_4$, the coordinates of the centroid at its respective frames are given by $\mathbf{x}^c(t_1), \dots, \mathbf{x}^c(t_4)$ (blue dots). Correspondingly, the displacement vector of trajectory $\mathcal{T}^k$ between extreme points $\mathbf{x}^k(t_2)$ and $\mathbf{x}^k(t_3)$ is given by $\Delta \mathbf{x}^k_{(t_3,t_3-t_2)}$ (red solid arrow) and the displacement vector of the centroid at its corresponding frames is given by $\Delta \mathbf{x}^c_{(t_3,t_3-t_2)}$ (blue solid arrow). The extrinsic motion dissimilarity at frame $t_2$, $e^k_{(t_3,t_3-t_2)}$, is obtained by computing the normalized inner product of $\Delta \mathbf{x}^k_{(t_3,t_3-t_2)}$ and $\Delta \mathbf{x}^c_{(t_3,t_3-t_2)}$. Finally, the measure of extrinsic motion for trajectory $\mathcal{T}^k$, $e^k$, is obtained by computing the weighted sum of the normalized inner products between extrinsic and intrinsic displacement vectors at all extreme points $t_1, \dots, t_4$ (i.e., $e^k_{(t_2,t_2-t_1)}$, $e^k_{(t_3,t_3-t_2)}$, and $e^k_{(t_4,t_4-t_3)}$). \label{fig:traj_extrinsic}}
\end{figure}

Similar to the measurement of oscillation and asymmetry, the measure of extrinsic motion dissimilarity, $\mathcal{M}_E$, is obtained by aggregating the dissimilarities of extrinsic and intrinsic values, $e^k$, via a weighted mean. That is, the measurement of extrinsic motion dissimilarity at frame $t$, $\mathcal{M}_E(t)$, is defined as
\begin{equation}
	\mathcal{M}_E(t) \equiv \frac{w_\nu}{|E(t)|} \sum_{e^k \in E(t)}{e^k} \text{,}
\end{equation}
where $e^k$ is the measurement of dissimilarity between extrinsic and intrinsic motion of trajectory $\mathcal{T}^k$, $E(t)$ is a trimmed set of dissimilarity calculations of trajectories present at frame $t$ analogous to \eqref{eq:trimmedSet}, $|E(t)|$ is its cardinality, and $w_\nu$ is a weight that accounts for the percent of total trajectories present in a given frame analogous to \eqref{eq:w_nu}. \\

It can be observed in Figure~\ref{fig:toy_example}f that the measurement of extrinsic motion dissimilarity is relatively large when the person, pendulum, and rolling ball are in motion (blue, green, and purple, respectively), while the car and bouncing ball (red and orange, respectively) have low measures since their extrinsic and intrinsic motions coincide.

\end{subsubsection}

\begin{subsubsection}{Biolocomotion Detector, $S_B$}\label{sec:E_bioloc}
	Once the raw measurements for each of the biolocomotion components have been computed, they must be normalized into a common range such that each component has equal contribution to the overall measure of biolocomotion. Then they must be combined into a single measure. \\

Each raw measurement, $\mathcal{M}_\chi$ for $\chi \in \{\Lambda, \Omega, \Sigma, E\}$, is normalized to range between $[0,1]$. This conversion allows each measure to be treated as a confidence score. The raw measurement is converted into a confidence score using a sigmoidal function according to
	\begin{equation}\label{eq:conf_score}
		S_\chi(t) = S(\mathcal{M}_\chi(t); \tau_\chi, k_\chi) \equiv \frac{1}{1+\exp{\left[-k_\chi(\mathcal{M}_\chi(t) - \tau_\chi)\right]}}
	\end{equation}
	for some values $\tau_\chi$ and $k_\chi$. The sigmoid function restricts the range to the desired interval $[0,1]$, with a smooth asymptotic behaviour as the extreme values are approached. \\

Treating the scores of locomotion, oscillation, asymmetry, and extrinsic motion akin to independent probabilities, the overall biolocomotion confidence score is obtained by taking the product of each component's confidence scores 
\begin{equation}\label{eq:bioloco_score}
	S_B(t) = \prod_{\chi \in X(t)}{S(\mathcal{M}_\chi(t); \tau_\chi, k_\chi)} \text{,}
\end{equation}
where $X(t)$ is a set of measurements to be considered in the final biolocomotion calculation. \\

Since asymmetric motion trace is most applicable when objects move orthogonal to a resistive force (e.g., gravity), the measurement of asymmetry, $\mathcal{M}_\Sigma$, is constrained to situations when objects move in the orthogonal direction. For example, as the foot pushes off against gravity for terrestrial animals, it results in a rapid rising trajectory, which is followed by a swing phase that moves entirely under the influence of gravity to yield a more elongated trace; e.g., see \cite{Chang09, Johansson73}. This phenomenon is less apparent when motion is parallel to gravity and similar pattern holds for other animals; e.g., see \cite{Berg10, Biewener03, Fish16}. Assuming the resistive forces are aligned with the image vertical, motion orthogonal to that direction, $\mathbf{r}=(1,0)$, should show an asymmetric trace. To capture the motion direction, the normalized inner product of the centroid displacement vector, $\Delta \bar{\mathbf{x}}^c_{(t,\Delta t)}$ as in \eqref{eq:centroid_disp}, and a reference vector, $\mathbf{r}$, is used to determine if the general direction of the object motion is perpendicular to gravity. More specifically, the average absolute value of the normalized inner products within a temporal window is calculated as
\begin{equation}
	w_\perp(t) = \frac{1}{k} \sum_{\Delta t = 1}^{k} \left\{\left| \frac{\langle \Delta\mathbf{x}^c_{(t,\Delta t)},\mathbf{r} \rangle}
								   {\|\Delta\mathbf{x}^c_{(t,\Delta t)}\| \|\mathbf{r}\|} \right| \right\}
	\text{,}
\end{equation}
where the absolute value of the normalized inner product maintains invariance to the direction of motion (left vs. right or toward vs. against gravity). Since we want to evaluate the \textit{general} path of the object, displacement vectors comparing centroid at frame $t$ to numerous centroids in previous frames are compared (i.e., $\Delta t = [1,k]$, where $k$ is empirically set to the frame rate of the video); see Figure \ref{fig:w_perp_mot}.
Thus,
\begin{equation}
	X(t) =
	\begin{cases}
		\{\Lambda, \Omega, \Gamma\} \text{,} & \text{ if } w_\perp(t) < \tau_\perp \\
		\{\Lambda, \Omega, \Sigma, \Gamma\} \text{,} & \text{ if } w_\perp(t) \geq \tau_\perp
	\end{cases}
	\text{,}
\end{equation}
where $\tau_\perp$ is empirically set to 0.9.\\

\begin{figure}[!h]
	\begin{center}
		\begin{tikzpicture}
	
	\draw[blue!30, line width=2mm, cap=round] (0,0) .. controls (1,0) and (2,1) .. (3,1);
	\draw[blue!30, line width=2mm, cap=round] (3,1) .. controls (4,1) and (5,-0.5) .. (6,-0.5);
	\draw[blue!30, line width=2mm, cap=round] (6,-0.5) .. controls (7,-0.5) and (8,0.5) .. (9,0.5);
	
	\filldraw[blue] (0,0) circle (2pt) node[below] {\scriptsize{$\mathbf{x}^c(t_1)$}};
	\filldraw[blue] (3,1) circle (2pt) node[above] (P1) {\scriptsize{$\mathbf{x}^c(t_2)$}};
	\filldraw[blue] (5,-0.1) circle (2pt) node[below] {\scriptsize{$\mathbf{x}^c(t_3)$}};
	\filldraw[blue] (9,0.5) circle (2pt) node[above] {\scriptsize{$\mathbf{x}^c(t_4)$}};

	\draw[blue,->,ultra thick] (3,1) -- (9,0.5) node[above,sloped,midway] {$\Delta \mathbf{x}^c_{(t_4,t_4-t_2)}$};
	\draw[blue,->,ultra thick] (5,-0.1) -- (9,0.5) node[below,sloped,midway] {$\Delta \mathbf{x}^c_{(t_4,t_4-t_3)}$};
	\draw[blue,->,ultra thick] (0,0) -- (9,0.5) node[above,sloped,midway] {$\Delta \mathbf{x}^c_{(t_4,t_4-t_1)}$};
	
	\draw[->,ultra thick] (9,0.5) -- (10,0.5) node[right] {$\mathbf{r}$};
\end{tikzpicture}
	\end{center}
	\caption[Computing displacement vectors for various temporal windows]{Computing displacement vectors for various temporal windows. To determine the \textit{direction} of the object, displacement vectors (blue arrows) across $k+1=4$ time steps, $t_1, \dots, t_4$, are considered along the path of the object (light blue curve). Each displacement vector, $\Delta \mathbf{x}^c_{(t_4,t_4-t_1)}, \dots, \Delta \mathbf{x}^c_{(t_4,t_4-t_3)}$, is compared with a reference vector, $\mathbf{r}=(1,0)$ (black arrow), via a normalized inner product. The absolute value of these differences are averaged to determine the overall direction of the object at frame $t_4$, $w_\perp(t_4)$. \label{fig:w_perp_mot}}
\end{figure}
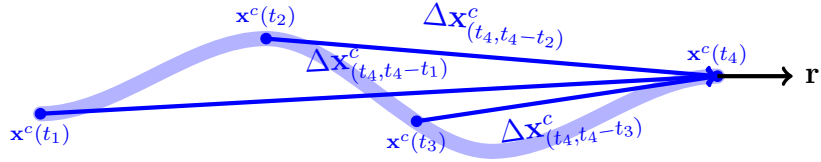

By construction, $S_B(t)$ is in the range of $[0,1]$ with large values more indicative of biolocomotion. As the product of the confidence scores for each component are taken, objects that lack locomotion, oscillation, asymmetry, or difference in extrinsic and intrinsic motion (e.g., pendulum, car, rolling ball, and bouncing ball) possess low $S_B$ values, while objects that exhibit locomotion, asymmetric oscillation, and difference in extrinsic and intrinsic motion (e.g., walking person) retain large $S_B$ values; see Figure~\ref{fig:toy_example}g. Hence, $S_B$ \eqref{eq:bioloco_score} is highly indicative of biolocomotion.
\end{subsubsection}
\end{subsection}

\begin{subsection}{Adaptive sliding temporal window}\label{sec:tech_aSTW}
Since the input to the biolocomotion detector is a cluster of trajectories, it is not necessary to process an entire video at once. Instead, an adaptively defined sliding temporal window is used to allow for incremental video processing. More specifically, the video is temporally segmented into disjoint sets of frames and each segment is processed successively. The segment lengths are defined adaptively to break the video into temporal extents, where the overall frame-to-frame image motion maintains the same coarse direction. Such subdivision provides a natural way to break the video into segments dominated by a single direction of camera motion (or in the absence of camera motion) with a single overall direction of object motion, e.g., the overall locomotion \eqref{eq:Elambda}. The overall direction of motion for frame $t$ is determined by using displacement vectors that are present at that frame. Specifically, for displacement vector $\Delta \mathbf{x}^k(t)$ of trajectory $\mathcal{T}^k$ at frame $t$ as in \eqref{eq:traj_disp}, the direction of the displacement for $\mathcal{T}^k$ is defined as
\begin{equation}
	\theta^k(t) = \arctan{ \left( \frac{\Delta y^k(t)}{\Delta x^k(t)} \right) } \text{, } 0^\circ \leq \theta^k(t) \leq 360^\circ \text{.}
\end{equation}
Then, the overall displacement direction is calculated by taking the average of $\theta^k(t)$ according to
\begin{equation}
	\Theta(t) = \frac{1}{N(t)}\sum_k{\theta^k(t)}\text{,} 
\end{equation}
where $N(t)$ is the number of trajectories.
The \textbf{adaptive Sliding Temporal Windows} (aSTW) are then defined to partition the video at times where $\Theta(t)$ and $\Theta(t+1)$ differ significantly (i.e., if $| \Theta(t) - \Theta(t+1) | > \tau_\theta$, then segment). Empirically, $\tau_\theta$ is set to $\frac{3}{2}f$ for frame rate $f$.

\end{subsection}

\begin{subsection}{Algorithmic summary}\label{sec:tech_summary}
	Algorithm 1 provides a summary of the overall approach to biolocomotion detection. Note that the approach both localizes and labels regions of biolocomotion of an input video.

\resizebox{0.9\textwidth}{!}{
\begin{algorithm}[H] \label{alg:summary}
	\SetKwInOut{input}{Input}\SetKwInOut{output}{Output}
	
	\input{Video, $V$}
	\output{(i) Biolocomotion confidence score, $S_B$ \eqref{eq:bioloco_score}, for each recovered trajectory cluster, \\
			(ii) the spatiotemporal location of the centre of each cluster, and \\
			(iii) the horizontal and vertical extents of the cluster at each frame}
	\BlankLine
	
	Extract and prune iDTs (\S \ref{sec:traj_postproc} and \ref{sec:traj_prune})\;
	Estimate average direction of motion for a crude division of temporal windows, $aSTW$ (\S \ref{sec:tech_aSTW})\;
	\BlankLine
	\For{$aSTW \in V$}{
		Cluster (\ref{sec:traj_clstr}), stabilize (\ref{sec:traj_stab}), and elongate trajectories (\ref{sec:traj_cat})\;
		\For{cluster}{
			Measure locomotion, $\mathcal{M}_\Lambda$ (\S \ref{sec:E_loc})\;
			Measure oscillation, $\mathcal{M}_\Omega$ (\S \ref{sec:E_osc})\;
			Measure asymmetry, $\mathcal{M}_\Sigma$ (\S \ref{sec:E_asym})\;
			Measure extrinsic motion dissimilarity, $\mathcal{M}_E$ (\S \ref{sec:E_ext})\;
			Measure biolocomotion, $S_B$ (\S \ref{sec:E_bioloc}) \;
		}
	}
	\BlankLine
\end{algorithm}
}
\\
\textbf{Algorithm 1:} Summary of the proposed biolocomotion detection algorithm.
\end{subsection}
\end{section}

\begin{section}[Evaluation]{Empirical evaluation}\label{ch:evaluation}
	\begin{subsection}{Overview}
Quantitative analysis of an algorithm can assist in the development and understanding of its strengths and weaknesses. A benchmark dataset can provide a way for comparative analysis with other algorithms or identification of important/unimportant components within an algorithm. To provide a benchmark for biolocomotion detection in videos, an extant dataset from a related field is exploited and a new dataset is also constructed to further test the robustness of the algorithm of interest. \\

This section unfolds in six subsections. This subsection has served to motivate the need for empirical evaluation. Subsection~\ref{sec:datasets} describes the datasets considered for the task of biolocomotion. Subsection~\ref{sec:eval_metric} describes the performance metrics used to quantitatively compare different components within an algorithm and across algorithms. Subsection~\ref{sec:param} describes how the set of parameters used for the presented algorithm is determined.
Subsection~\ref{sec:alternate} presents an alternative approach to biolocomotion detection that is based on generic handcrafted features combined with learning. Since it appears that no previous algorithms for biolocomotion detection have been developed, this alternative approach provides a basis of comparison for the main approach presented in Section~\ref{ch:techApproach}. Subsection~\ref{sec:results} provides qualitative and quantitative results along with discussions comparing individual components of the proposed and alternative approaches as well as more general comparisons between the two approaches. Finally, Subsection~\ref{sec:eval_conclusion} provides an overall summary of the evaluations.
\end{subsection}

\begin{subsection}{Datasets}\label{sec:datasets}
	While there are many benchmark datasets in the field of computer vision that can be used for quantitative evaluation of various detection algorithms in the context of humans (e.g., CMU Crowded Videos \cite{Ke07}, UCF Sports \cite{Rodriguez08, Soomro14}, UCF101 \cite{Soomro12}, ActivityNet \cite{Heilbron15}, J-HMDB \cite{Jhuang13}, etc.), there are none available for biolocomotion detection. 
Consequently, a novel dataset, Biological Object in Locomotion Detection (BOLD) dataset, is developed and biolocomotion annotations for an extant camouflaged animals dataset (CAD) \cite{Bideau16} are provided.
This section contains a description of the considered datasets as well as biolocomotion annotation.

\begin{subsubsection}{Biological Objects in Locomotion Detection (BOLD) Dataset}
The Biological Objects in Locomotion Detection (\textbf{BOLD}) dataset builds on an extant dataset, A2D \cite{Xu15}. A2D is taken as a starting point as it provides diversity of objects in motion (i.e., more than humans) and also includes labelled regions where they are present. 
However, it lacks generality to validate biolocomotion detection. Specifically, A2D does not consider undulation nor swim as one of its actions and its species diversity remains too limited for present purposes. Thus, a richer set that contains more variety of objects and locomotion types to provide stronger justification of a biolocomotion detection algorithm is needed. Consequently, the BOLD dataset is constructed to provide greater diversity in terms of object type and modes of locomotion. BOLD builds on a subset of videos from A2D and is substantially augmented with various videos from YouTube. It expands the diversity of objects by including: \textit{reptiles}, \textit{cetaceans}, \textit{seals}, \textit{fish}, \textit{stingray}, \textit{eel}, \textit{sea snakes}, \textit{snakes}, \textit{insects}, \textit{spiders}, \textit{scorpions}, \textit{lobsters}, \textit{trains}, \textit{motorbikes}, \textit{submarines}, \textit{airplanes}, \textit{helicopters}, \textit{rockets}, \textit{metronomes}, and \textit{pendulums} and modes of locomotion by adding: \textit{swim}, \textit{undulate}, and \textit{oscillate}; a comparison of A2D and BOLD is visualized in Figure \ref{fig:bold_wordle} and a detailed breakdown is provided in Table \ref{tab:a2d_bold}. \\

BOLD contains 1,348 videos split into 1,078 training and 270 test videos. Videos range between $22-504$ frames and on average span 143.98 frames. Bounding box annotations are provided for at least a frame per video, up to 18 frames in longer videos; see Figure \ref{fig:bold}. 

\begin{figure}[!h]
	\begin{center}
	\resizebox{0.9\linewidth}{!}{
		\includegraphics[width=\textwidth]{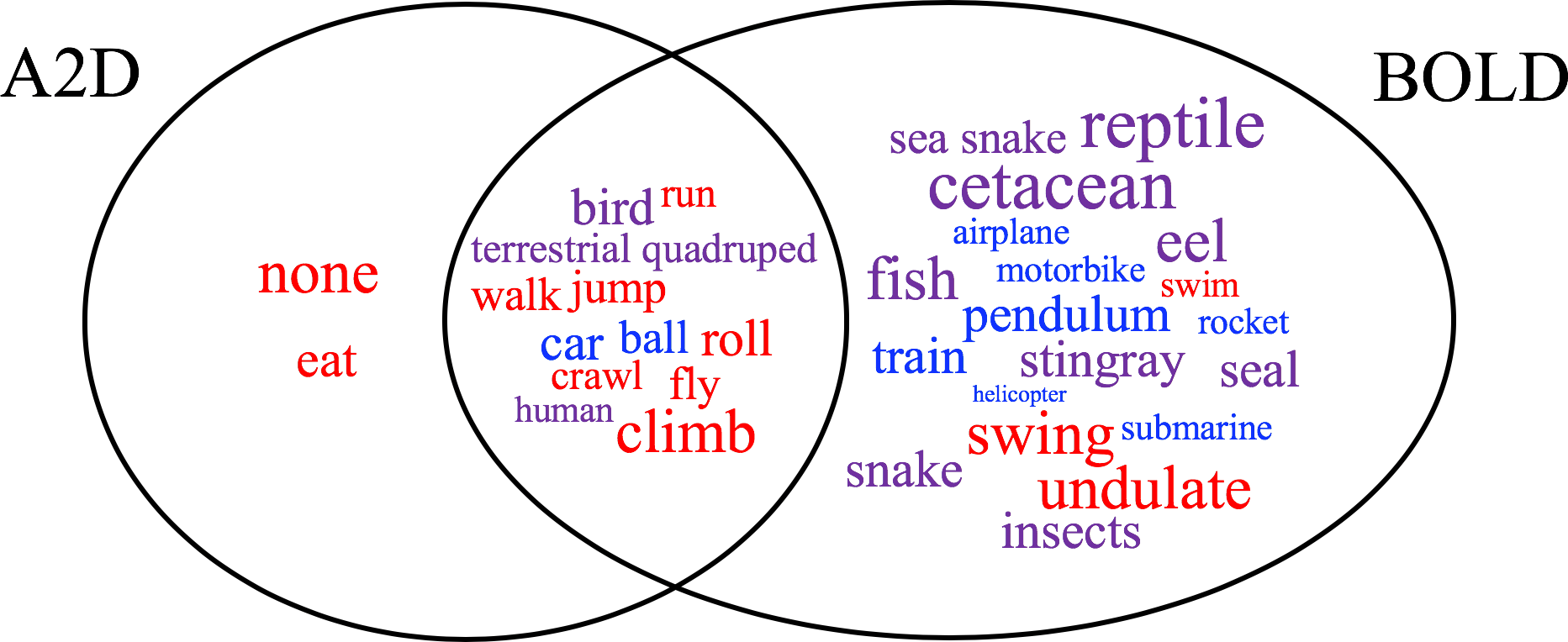}
	}
	\end{center}
	\caption[Venn diagram of A2D and BOLD]{Venn diagram of motions and species that are contained in A2D and BOLD. The wordle in each component of the Venn diagram illustrates a set of actions (red), biological objects (purple), and non-biological objects (blue) that are contained in A2D $\setminus$ BOLD, A2D $\cap$ BOLD, and BOLD $\setminus$ A2D.  \label{fig:bold_wordle}}
\end{figure}

	\begin{table}[!h]
		\begin{center}
		\resizebox{\textwidth}{!}{
			\begin{tabular}{| c | r | c | c | c | c | c | c | c | c | c | c | c | c | c |}
				\cline{1-13}
				\multicolumn{2}{|c|}{\backslashbox[110mm]{\textbf{Object Type}}{\textbf{Motion Type}}}& climb 	& crawl 	& fly 	& jump 	& roll	& run & swim & undulate & walk & locomotion & swing \\
				\cline{1-13}
				\parbox[t]{2mm}{\multirow{12}{*}{\rotatebox[origin=c]{90}{Biological Species}}}& human (adult, baby) & \aad{33} & \aad{27}	&  -	& \aad{36}	& \aad{33} & \aad{33} & \maad{20} & - & \aad{31} 	& - & - \\
				& terrestrial quadruped mammal (cat, dog) & \aad{40} & \aad{25} &  - & \aad{37} & \aad{18} & \aad{67} & -  & - & \aad{35} 	& - & - \\
				& bird	(incl. penguin) & \aad{35} & - & \aad{100} & \aad{33}	& \aad{10} & -	 & - & - & \aad{26} + \maad{8}	&  - & -\\
				& reptile (alligator, chameleon, crocodile, lizard, turtle) &  -	& \maad{25} &  -	& -	& - & - & \maad{15} & - & - & - & -\\
				& cetacean (dolphin, shark, whale) &  -		& -		&  -	& -	& - & - & \maad{23}	& - & - & - & -\\
				& seal &  -		& -		&  -	& -	& - & - & \maad{18} & - & - & - &  -\\
				& fish & -		& -		&  -	& -	& - & - & \maad{20} & - & - & - &  -\\
				& stingray & -	& -		&  -	& -	& - & - & \maad{20} & - & - & - &  -\\
				& eel &  -		& -		&  -	& -	& - & - & - & \maad{20} & - & -  &  -\\
				& sea snake &  -		&  -		& -	& -	& - & - & - & \maad{20} & - & - &  -\\
				& snake &   -		&  -		& -	& -	& - & - & - & \maad{63}	& - & - &  -\\
				& insects, spiders, scorpions, lobster & -		&  \maad{25} & -	& -	& - & - & -	& - 	& - & - &  -\\
				\cline{1-13}
				\parbox[t]{2mm}{\multirow{9}{*}{\rotatebox[origin=c]{90}{Non-biological Objects}}} & ball	&	- 	& 	-	& \aad{25} & \aad{50} & \aad{50} & -	 & - & - & 	- 	& - & -\\
				& car	& -		& -		&  -	& \aad{50} & \aad{50} & - & - & - & -		& \aad{25} & -\\
				& train & -		& -		&  -	& -	& - & - & - & - & -		& \maad{33} & - \\
				& motorbike & -		& -		&  -	& -	& - & - & - & - & -		& \maad{33} & - \\
				& submarine &  -		&  -		& -	& -	& - & - & \maad{25}	& - 	& - & - & - \\
				& airplane &  -		&  -		& \maad{27}	& -	& - & - & -	& - 	& - & \maad{9} & - \\
				& helicopter &  -		&  -		& \maad{25}	& -	& - & - & -	& - 	& - & - & - \\
				& rocket &  -		&  -		& \maad{25} & -	& - & - & -	& - 	& - & - & - \\
				& oscillating stuff (metronome, pendulum, boat) &  -		&  -		& - & -	& - & - & -	& - 	& - & - & \maad{25} \\
				\hline
				& Total (Biological) &  \aad{108}+\maad{0} & \aad{52} + \maad{50} & \aad{100} + \maad{0} & \aad{106} + \maad{0} & \aad{61} + \maad{0} & \aad{100} + \maad{0} & \aad{0} + \maad{116} & \aad{0} + \maad{103} & \aad{92} + \maad{8} &  - & - & \aad{619} + \maad{277}\\
				& Total (Non-biological) &  - & - & \aad{25} + \maad{77} & \aad{100} + \maad{0} & \aad{100} + \maad{0} & - & \aad{0} + \maad{25} & - & - & \aad{25} + \maad{75} & \maad{25} & \aad{250} + \maad{202}\\
				\hline
			\end{tabular}
			}
		\end{center}
		\caption[Breakdown of motion and objects in BOLD]{Summary of the number of videos in each motion and object class for the BOLD dataset. The numbers indicate the number of videos in each object-motion class, hyphen (-) indicates lack of videos in the respective categories. Numbers in orange indicate videos extracted from A2D, and green indicates new videos that were obtained from YouTube. \label{tab:a2d_bold}}
	\end{table}
	
\begin{figure}
	\begin{center}
	\resizebox{\linewidth}{!}{
		\begin{tabular}{C{\textwidth}}
			\includegraphics[width=\textwidth]{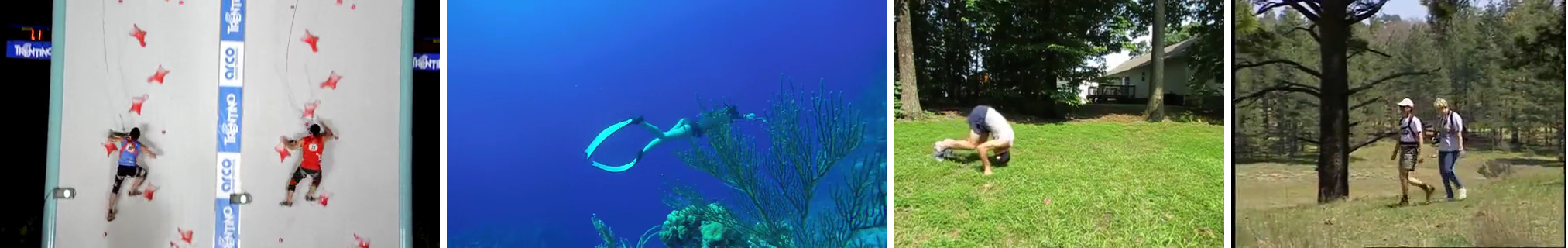} \\
			\includegraphics[width=\textwidth]{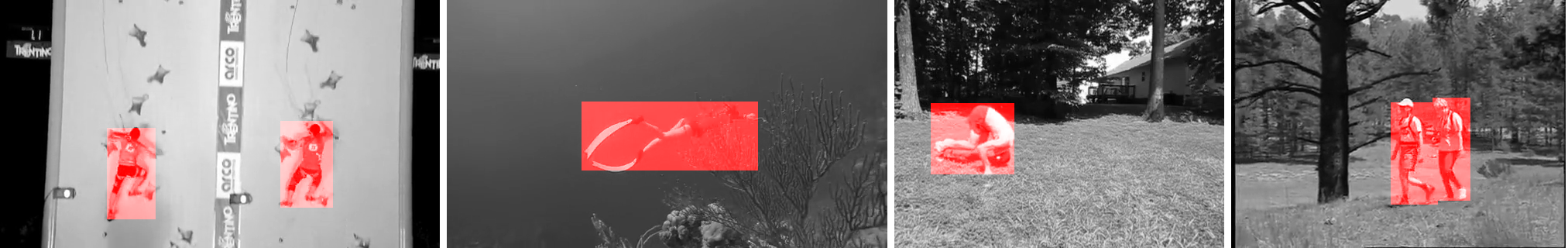} \\
			(a) Humans locomoting in various forms (e.g., climb, swim, roll, and walk).\vspace{3mm} \\
		
			\includegraphics[width=\textwidth]{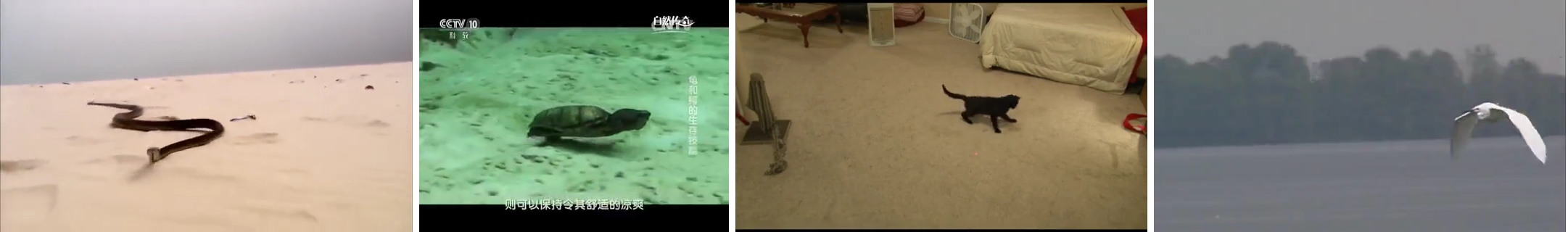} \\
			\includegraphics[width=\textwidth]{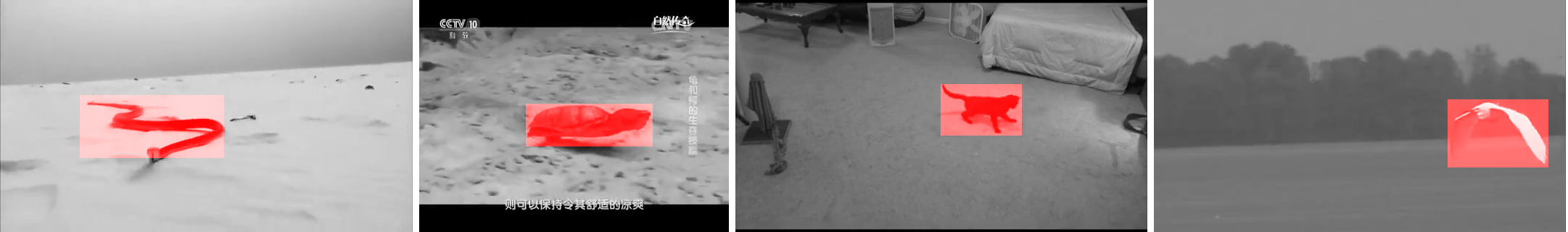} \\
			(b) Variations in biological species (e.g., snake, turtle, terrestrial quadruped, and bird). \vspace{3mm} \\
	
			\includegraphics[width=\textwidth]{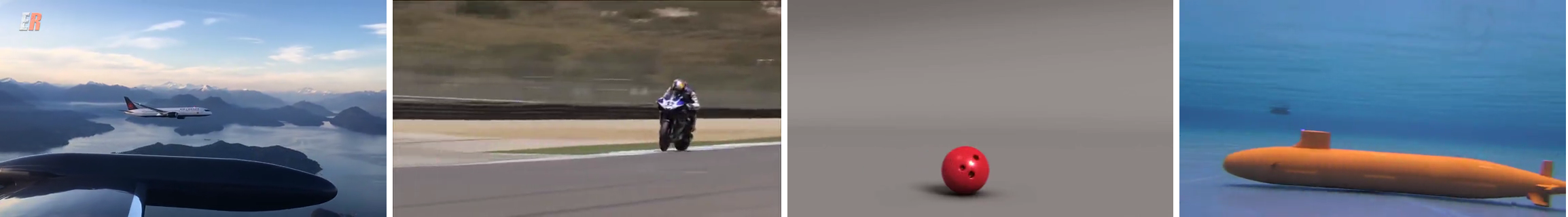} \\
			\includegraphics[width=\textwidth]{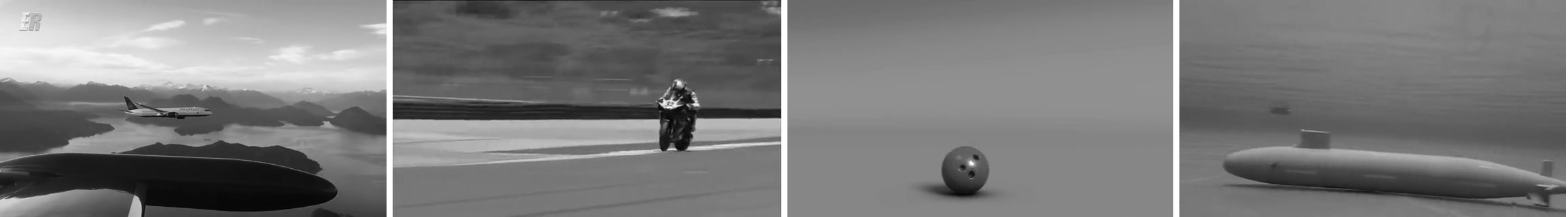} \\
			(c) Various non-biological objects in locomotion (e.g., airplane, motorcycle, ball, and submarine). \\
		\end{tabular}
	}
	\end{center}
	\caption[Illustration of videos in BOLD and its annotations]{Illustration of videos (top) in BOLD and its biolocomotion annotations indicated with overlaid red boxes (bottom). Select frames of videos from BOLD demonstrating variation in modes of locomotion exhibited by (a) humans, (b) other biological species, and (c) non-biological objects. \label{fig:bold}}
\end{figure}
\end{subsubsection}

\begin{subsubsection}{Camouflage Animals Dataset (CAD)}
	The Camouflage Animals Dataset (\textbf{CAD}) \cite{Bideau16} was created for motion segmentation with the challenge of detecting camouflaging animals in motion. It contains nine videos of animals in motion: \textit{chameleon},  \textit{frog}, \textit{glowworm beetle}, 4 \textit{scorpions}, \textit{snail}, and \textit{stick insect}. All videos in CAD are test videos. As CAD lacks training data, transfer learning is applied, where training data from BOLD is used to test videos in CAD. Videos on average span $92.89$ frames and range between $30-218$ frames. Dense-pixel level annotations are provided for at least 12 frames; see Figure~\ref{fig:cad}.
	
\begin{figure}[!h]
	\begin{center}
	\resizebox{\linewidth}{!}{
	\begin{tabular}{c}
		\includegraphics[width=\textwidth]{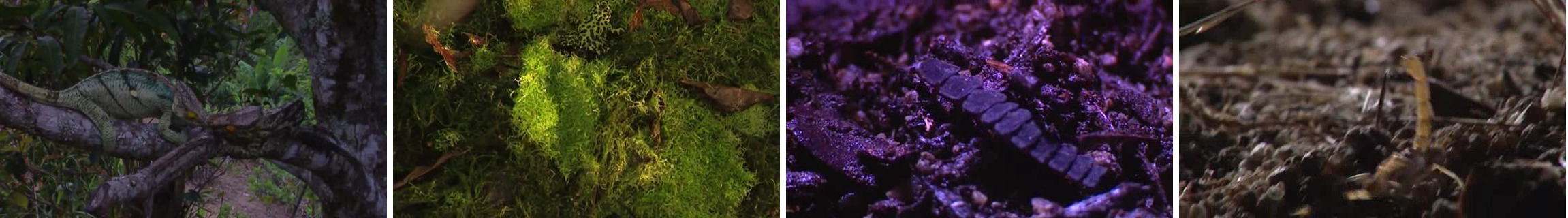}\\
		\includegraphics[width=\textwidth]{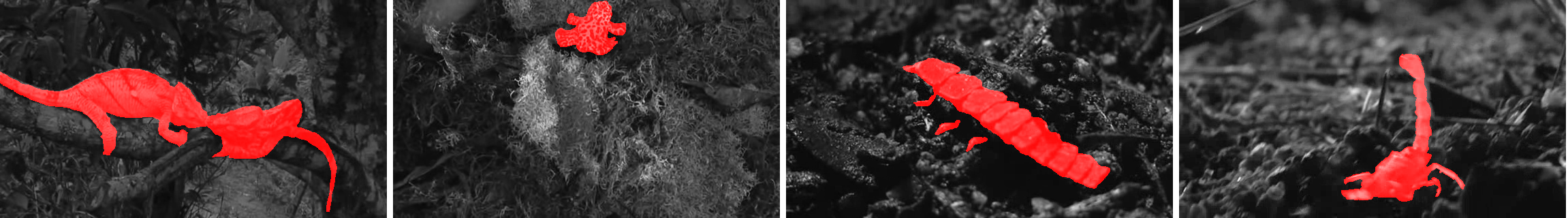} 
	\end{tabular}
	}
	\end{center}
	\caption[Illustration of videos from CAD and its annotations]{Illustration of select videos (top) in CAD \cite{Bideau16} and its fine-grained biolocomotion annotations overlaid in red on its respective raw image frames (bottom). \label{fig:cad}}
\end{figure}
\end{subsubsection}

\begin{subsubsection}{Biolocomotion annotations}\label{sec:bioloco_anno}
To determine how well a proposed algorithm performs in the task of biolocomotion detection, each annotated region is supplied with a biolocomotion label. Thus, each object is spatially identified, via bounding boxes (i.e., BOLD) or fine-grained segmentation (i.e., CAD), for select frames and categorized into its appropriate class (i.e., biolocomotion or non-biolocomotion). The annotated regions serve as the set of positives (or negatives) for ensuing experiments.
While the main focus is placed on biolocomotion, finer grained annotations are provided for more general use.
In particular, all labelled objects are identified as being in one of six categories:
\begin{enumerate}[label=(\roman*)]
	\item biological objects in locomotion (i.e., biolocomotion), 
	\item biological objects that are oscillating but not locomoting, 
	\item biological objects that are neither locomoting nor oscillating,
	\item non-biological objects that are only locomoting, 
	\item non-biological objects that are only oscillating, and
	\item non-biological objects that are neither locomoting nor oscillating.
\end{enumerate}
Visualization of these classes can be seen in Figure \ref{fig:bioloco_annot}.\\

Here, locomotion corresponds to non-zero displacement exhibited by the object in the real-world. For example, a cat (biological object) walking (action related to locomotion) on a treadmill results in zero displacement in the real-world, thus is classified as non-biolocomotion. \\

In addition to the six object-action labels, each video also has an associated label for camera motion according to five categories:
\begin{enumerate}[label=(\roman*)]
	\item translation,
	\item rotation,
	\item zoom,
	\item more than one of the above, and
	\item no camera motion.
\end{enumerate}
These labels could be used in future work to study algorithm performance as a function of camera motion. 

\end{subsubsection}

\begin{subsubsection}{Summary}
Table~\ref{tab:summary_dataset} provides a summary of the key features of the considered datasets for the task of biolocomotion detection. Also, for the sake of size consistency, all videos are resized to a fixed height of $320$ pixels and the widths are adjusted accordingly to maintain the original aspect ratio.

\newgeometry{top=1.25in,bottom=1.25in}
\begin{figure}
	\begin{tabular}{C{\linewidth}}
		\includegraphics[width=\textwidth]{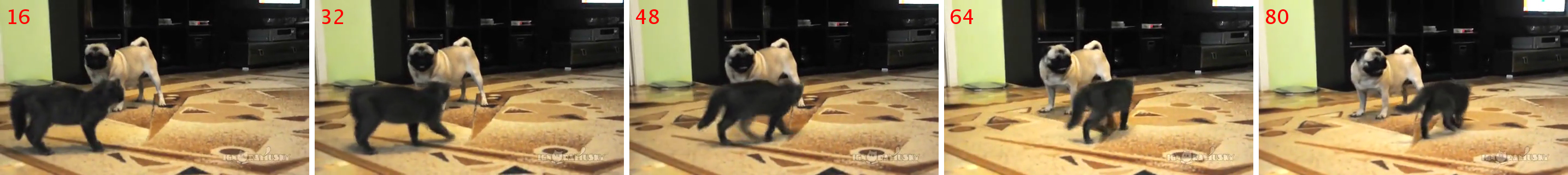} \\
		\includegraphics[width=\textwidth]{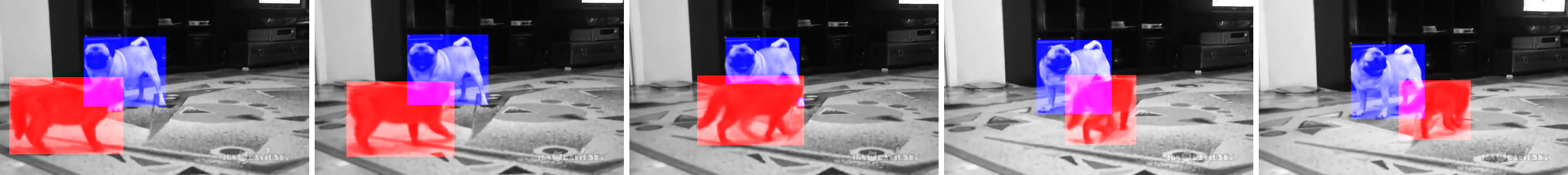} \\
		(a) biolocomotion (red) and biological objects not in locomotion nor oscillation (blue). \vspace{2mm} \\
		
		\includegraphics[width=\textwidth]{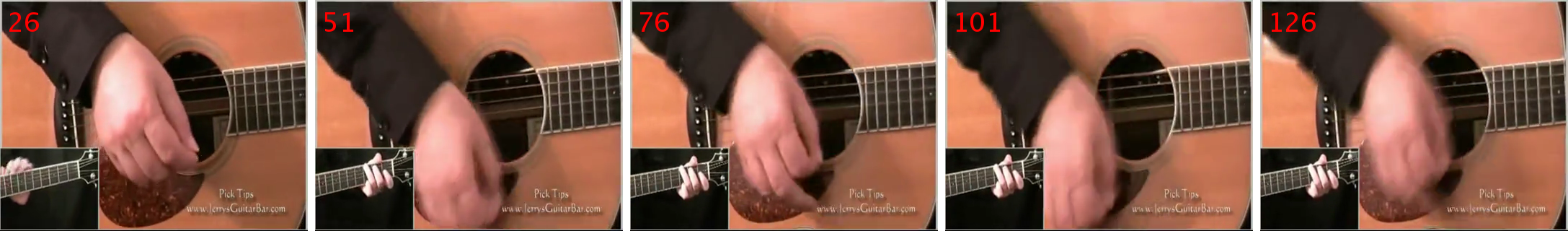} \\
		\includegraphics[width=\textwidth]{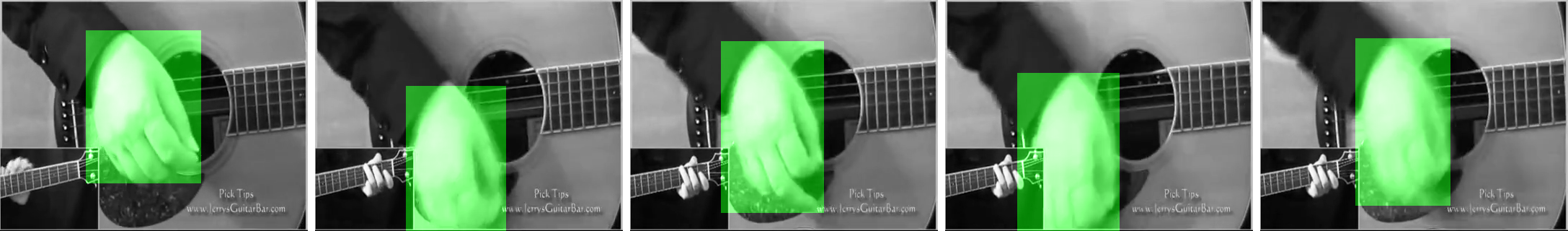} \\
		(b) biological object in oscillation. \vspace{2mm} \\
		
		\includegraphics[width=\textwidth]{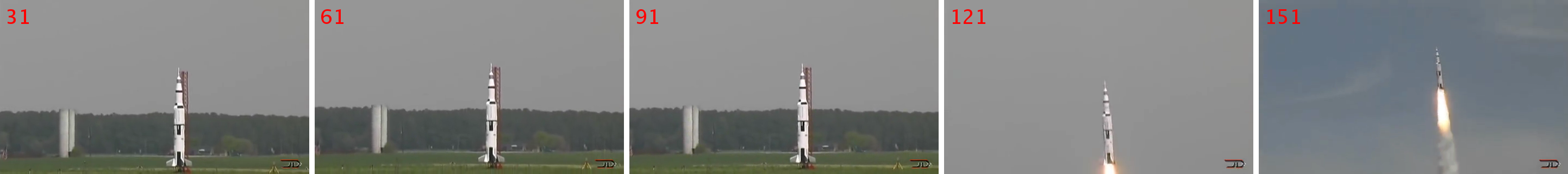}\\
		\includegraphics[width=\textwidth]{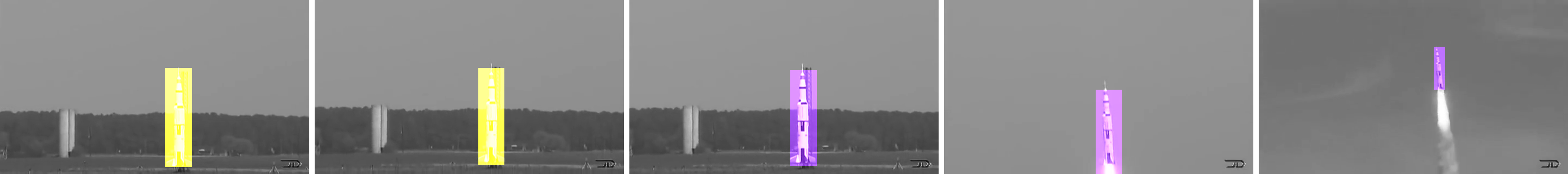}\\
		(c) non-biological object in locomotion (purple) and non-biological object neither in locomotion nor oscillation (yellow). \vspace{2mm} \\
		
		\includegraphics[width=\textwidth]{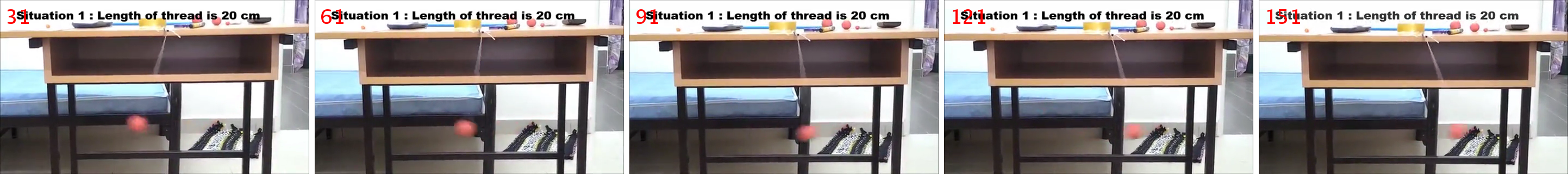}\\
		\includegraphics[width=\textwidth]{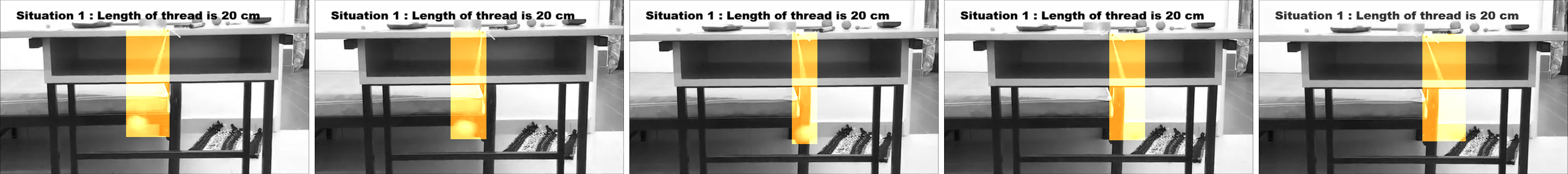}\\
		(d) non-biological object in oscillation (orange).\\
	\end{tabular}
	\caption[Biolocomotion annotation illustrations]{Biolocomotion annotation illustrations. Select videos demonstrating the provided annotations with overlaid boxes: biolocomotion (red), biological objects in oscillation (green), biological objects not in locomotion nor oscillation (blue), non-biological objects in locomotion (purple), non-biological objects in oscillation (orange) and non-biological objects in neither locomotion nor oscillation (yellow). \label{fig:bioloco_annot}}
\end{figure}
\restoregeometry

\begin{landscape}
	\begin{table}
		\begin{center}
		\resizebox{\linewidth}{!}{
		\begin{tabular}{| L{0.2\linewidth} | m{0.4\linewidth} | m{0.4\linewidth} |}
			\hline
			\textbf{Dataset} & \textbf{BOLD} & \textbf{CAD} \cite{Bideau16} \\
			\hline
			\textbf{Task} & biolocomotion detection & causal motion segmentation \\
			\hline
			\textbf{Source} & \multicolumn{2}{c |}{YouTube} \\
			\hline
			\textbf{Objects} & human, terrestrial quadruped, bird, reptile, cetacean, seal, fish, stingray, eel, sea snake, insects, spiders, scorpion, lobster, ball, car, train, motorbike, submarine, airplane, helicopter, rocket, oscillating stuff 
 						& chameleon, frog, glowworm beetle, scorpion, snail, stick insect \\
			\hline
			\textbf{Number of videos} & $1,348$ videos \begin{itemize}[topsep=0pt]\setlength\itemsep{-0.5em} \item $1,078$ training \item $270$ test \end{itemize}
					     & $9$ videos \begin{itemize}[topsep=0pt]\setlength\itemsep{-0.5em} \item 0 training \item 9 test \end{itemize} \\
			\hline
			\textbf{Duration} & \begin{itemize}\setlength\itemsep{-0.5em} \item $22-504$ frames \item avg. $143.98$ frames \end{itemize}
			& \begin{itemize}\setlength\itemsep{-0.5em} \item $30-218$ frames \item avg. $92.89$ frames \end{itemize}\\
			\hline
			\textbf{Number of videos with biolocomotion} & 882 & 9 \\
			\hline
			\textbf{Challenges} & \multicolumn{2}{c |}{Variations in viewpoint, camera motion present, background and foreground clutter present} \\
			\hline
			\textbf{Groundtruth labelling} 
				& \begin{itemize}\setlength\itemsep{-0.5em} \item bounding boxes \item $1-18$ frames \item avg. 4.03 frames \end{itemize} 
				& \begin{itemize}\setlength\itemsep{-0.5em} \item fine-grained \item $12-43$ frames \item avg. 21.22 frames \end{itemize}\\
			\hline
		\end{tabular}
		}
		\end{center}
		\caption{Summary of the datasets. \label{tab:summary_dataset}}
	\end{table}
\end{landscape}

\end{subsubsection}
\end{subsection}

\begin{subsection}{Evaluation metrics}\label{sec:eval_metric}
	To quantitatively evaluate the algorithms of interest, two standard detection metrics are applied to the test sets. To evaluate the value of each component of an algorithm, detection results are plotted as precision-recall (PR) curves. To compare between opposing algorithms, average precision (AP) scores are reported. In the following, detection results (e.g., cluster outputs with biolocomotion score $S_B$ from Algorithm~\ref{alg:summary}) are referred to as \textit{biolocomotion proposals}, which are regions likely to contain biolocomotion, associated with confidence scores.\\

Precision-recall (\textbf{PR}) curves are obtained by calculating precision and recall values at various confidence thresholds. \textit{Precision} is the percentage of correctly assigned pixels, which is defined as
\begin{equation}\label{eq:precision}
	precision = \frac{TP}{TP + FP} \text{,}
\end{equation}
where TP denotes the number of correctly predicted pixels with respect to the groundtruth (i.e., true positive) and FP denotes the number of incorrectly predicted pixels with respect to the groundtruth (i.e., false positive).
\textit{Recall} is the percentage of detected pixels with respect to the labelled groundtruth pixels, which is defined as
\begin{equation}
	recall = \frac{TP}{TP+FN} \text{,}
\end{equation}
where FN denotes the number of incorrectly missed pixels (i.e., false negative).
A set of pixels within the proposal are considered a positive, which outputs a mask, $M_P$, if the biolocomotion score $S_B$ is greater than or equal to some threshold $\tau_B \in [0,1]$.
PR curves are generated by computing the precision and recall values at various detection thresholds, $\tau_B$, for each annotated frame with precision and recall plotted along the ordinate and abscissa, respectively. \\

Average precision (\textbf{AP}) is a standard evaluation methodology used in action detection (e.g., \cite{Heilbron15, Jiang14}) that evaluates how well an action proposal (i.e., region likely to contain an action) is ranked for the specified action class. Similarly, AP can be used for biolocomotion detection to determine how well each biolocomotion proposal is ranked. AP is defined as 
\begin{equation}\label{eq:ap}
	AP = \frac{\sum_{k=1}^{n}{(P(k)\times rel(k))}}{\sum_{k=1}^{n}{rel(k)}} \text{,}
\end{equation}
where $n$ is the total number of proposals, $P(k)$ is the precision at cutoff $k$ of the list of proposals, and $rel(k)$ is an indicator function which equals 1 if the $k^{th}$ ranked proposal is a true positive and 0 otherwise. The denominator in \eqref{eq:ap} represents the total number of true positives in the list. To determine whether the proposal should be considered a true or false positive, the Intersection over Union (\textbf{IoU}) between the predicted mask and the groundtruth is considered, which is defined as
\begin{equation}
	IoU(M_P,M_{GT}) = \frac{M_P \cap M_{GT}}{M_P \cup M_{GT}} \text{,}
\end{equation}
where $M_P$ denotes the predicted mask and $M_{GT}$ denotes the groundtruth mask. 
A predicted mask is considered correct if $IoU(M_P,M_{GT})$ is greater than or equal to some constant, $\iota \in [0,1]$. AP scores are calculated as $\iota$ varies. Note that a typical action detection problem considers mean AP (mAP) scores as there are multiple classes to be considered, but the problem of interest considers AP as there is only one class (i.e., biolocomotion). \\

The AP metric evaluates the \textit{quality} of biolocomotion detection results by ranking the proposals using biolocomotion scores. 
However, its consideration of positives only disregards missed regions, limiting a thorough investigation of an algorithm.
The PR curve, on the other hand, considers both positives and negatives of a detection algorithm. The recall value in a PR curve, however, is often given less importance in detection work \cite{Borji15, Cheng15, Liu07, Perazzi12}, since a perfect recall value (i.e., $recall = 1$) is attained if the predicted mask is larger than the groundtruth (i.e., $M_{GT} \subseteq M_P$), which can be easily achieved by simply selecting the whole image. This often comes at a cost of a reduced precision value, while a good detection algorithm should be able to (spatiotemporally) locate the biolocomoting object as accurately as possible (i.e., with high precision) \cite{Liu07, Perazzi12}. 
Thus, PR curves become insufficient for comparison, especially if the recall values of the results reside in very little overlapping ranges.
Consequently, AP values and PR curves are used in tandem for the evaluation of biolocomotion algorithms, where PR curves serve as a reliable measure for comparing various components of an algorithm and AP values serve as a good metric for comparing the quality of biolocomotion proposals output by different algorithms. 

\end{subsection}

\begin{subsection}{Algorithm parameter values}\label{sec:param}
	The parameters, $\tau_\chi$ and $k_\chi$, necessary for normalizing the raw measurements, $\mathcal{M}_\chi$, into confidence scores, $S_\chi$ in \eqref{eq:conf_score}, are determined via 1D grid search. The 1D grid search is performed by obtaining PR curves for each component on the BOLD training data for numerous thresholds. The \textit{locomotion}, $\Lambda$, component is compared with annotations labelled as (biological or non-biological) locomotion; the \textit{oscillation}, $\Omega$, component is compared with annotations labelled as biolocomotion or (biological or non-biological) oscillation; the \textit{asymmetry}, $\Sigma$, and \textit{extrinsic motion dissimilarity}, $E$, components are compared with annotations labelled as biolocomotion. 
	The values that optimize the area under the PR curve are chosen.
	Since CAD lacks training data, the same set of parameters as in BOLD are used in CAD.
	Table~\ref{tab:param} provides a summary of the empirically set parameters for evaluation. \\
	
\begin{table}[!h]
	\begin{center}
		\begin{tabular}{| c | c | c |}
			\hline
			\textbf{Measure} ($\mathbf{\chi}$) & $\tau_\chi$ & $k_\chi$ \\
			\hline
			Locomotion ($\Lambda$) & $\sfrac{51}{10}=5.1$ & $(\sfrac{10}{51})\ln 99$ \\
			Oscillation ($\Omega$) & $\sfrac{27}{10}=2.7$ & $(\sfrac{10}{27})\ln 99$ \\
			Asymmetry ($\Sigma$) & $\sfrac{3}{2}=1.5$ & $(\sfrac{2}{3})\ln 99$ \\
			Extrinsic Motion Dissimilarity ($E$) & $\sfrac{21}{200}=0.105$ & $(\sfrac{200}{21})\ln 99$ \\
			\hline
		\end{tabular}		
	\end{center}
	\caption[Parameters to convert raw measurements into confidence scores]{Parameters used to obtain confidence score, $S_\chi$, for measure $\mathcal{M}_\chi$ for $\chi\in\{\Lambda, \Omega, \Sigma, E\}$. \label{tab:param}}
\end{table}

To ensure the predicted masks and groundtruth masks are compared fairly, a tight bounding box fit around the ellipse is used as predicted masks for BOLD since its annotations are bounding boxes. 

\end{subsection}

\begin{subsection}{Alternative Algorithm}\label{sec:alternate}
	Since biolocomotion detection in videos is a new field of research in computer vision, there are no previous algorithms for comparison. Nevertheless, it remains of interest to have some basis for comparison in evaluating the approach to biolocomotion detection presented in Section~\ref{ch:techApproach}. In response, an alternative algorithm has been assembled from components of extant algorithms from the related-field of video-based action analysis. In particular, an extant algorithm representative of state-of-the-art in handcrafted approaches to action proposals \cite{vanGemert15} and action recognition \cite{Wang13} is considered and adapted to biolocomotion detection. Along with their reliance on generic handcrafted features, these approaches also make use of learning to derive mid-level feature encodings and final classification; see Figure~\ref{fig:overview_alter}. Beyond their documented strong performance in action analysis \cite{vanGemert15, Wang13}, these approaches have the added comparative benefit in that their input are also dense trajectories, similar to the proposed algorithm. \\

\begin{figure}[t]
	\begin{center}
	\resizebox{\textwidth}{!}{
		\begin{tikzpicture} 
	
	\node[draw=NavyBlue, text width=3cm, align=center, rounded corners=1mm, ultra thick] (rect) at (0,0) (trainVid) {\textbf{Raw Video}};
	
	\node[draw=ForestGreen, text width=3cm, align=center, rounded corners=1mm, ultra thick] (rect) at (4,0) (prop) {\textbf{Generate Proposals}};
	\draw [->, thick] (trainVid) -- (prop);
	
	\node[draw=ForestGreen, text width=4.5cm, align=center, rounded corners=1mm, ultra thick] (rect) at (9,0) (descriptor) {\textbf{Extract Descriptors}\\ e.g., trajectory shape, HOG, HOF, MBH};
	\draw [->, thick] (prop) -- (descriptor);
	
	\node[draw=ForestGreen, text width=4.5cm, align=center, rounded corners=1mm, ultra thick] (rect) at (15,0) (encoder) {\textbf{Encode Descriptors}\\
	e.g., iFV};
	\draw[->, thick] (descriptor) -- (encoder);
		
	\node[draw=NavyBlue, text width=4.5cm, align=center, rounded corners=1mm, ultra thick] (rect) at (25,0) (score) {\textbf{Biolocomotion Scores}\\ e.g., Platt scaling};
	\draw[->, thick] (encoder) .. controls (20,1.7) .. (score);
	
	\node[draw=ForestGreen, text width=3.5cm, align=center, rounded corners=1mm, ultra thick] (rect) at (20,2) (classifier) {\textbf{Train Classifier}\\
	e.g., SVM};
	\draw[-, thick] (15,0.6) -- (15,2);
	\draw[->, thick] (15,2) -- (classifier);
\end{tikzpicture}
		}
	\end{center}
	\caption[General framework of the alternative algorithm]{General framework of the alternative algorithm \cite{vanGemert15}. For each video, a set of proposals \cite{vanGemert15} are generated and a generic set of descriptors that typically appear with iDTs (e.g., trajectory shape \cite{Wang13}, HOG \cite{Dalal05}, HOF \cite{Laptev08}, and MBH \cite{Dalal06}) are extracted. The descriptors are encoded (e.g., via iFV \cite{Perronnin10}) into a learned representation specific to biolocomotion detection. The encoded descriptors that result from a set of training videos are used to train a classifier (e.g., SVM \cite{Cortes95}), which is used to obtain a biolocomotion score (e.g., via Platt scaling \cite{Platt99}) for encoded descriptors from a test video. The input (raw video) and output (biolocomotion score) are marked in blue and the intermediate steps are marked in green. \label{fig:overview_alter}}
\end{figure}
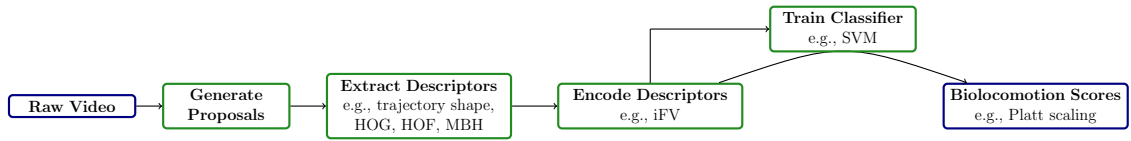

	An algorithm based on a ConvNet with deep-learning is not considered as it is not feasible for biolocomotion detection given the relatively small amounts of training data available (cf. action recognition training on UCF101 \cite{Soomro12} and kinetics \cite{Carreira17}, where the datasets are 1-2 orders of magnitude greater). Therefore, the proposed approach is compared to a handcrafted approach with learned mid-level feature encoding and classification. \\

	The action proposal \cite{vanGemert15} and recognition \cite{Wang13} algorithms are combined as follows.
	A biolocomotion score is associated to each generated \textit{action proposal based on trajectories} (\textbf{APT}) \cite{vanGemert15} by 
	(i) recovering a set of descriptors that typically appear with (improved) dense trajectories (i.e., trajectory shape \cite{Wang11}, HOG \cite{Dalal05}, HOF \cite{Laptev08}, and MBH \cite{Dalal06}), 
	(ii) encoding them using improved Fisher Vectors (iFVs) \cite{Perronnin10}, and 
	(iii) obtaining a biolocomotion score from a classifier (i.e., analogous to the iDT approach to action detection \cite{Wang13}). \\
	
	Trajectory shape, HOG, HOF, and MBH are standard descriptors used in video analysis tasks, especially action recognition (e.g., \cite{Feichtenhofer16, Wang11}).
	\textit{Trajectory shape} features are obtained by concatenating the normalized displacement vectors of tracked trajectories \cite{Wang11}.
	\textit{Histogram of Oriented Gradients} (HOG) \cite{Dalal05} are descriptors that store spatially oriented gradients of image intensity along a discrete set of directions to capture appearance information.
	\textit{Histogram of Optical Flow} (HOF) \cite{Laptev08} captures local motion of the pattern by quantizing the orientation of the optical flow vectors.
	\textit{Motion Boundary Histogram} (MBH) \cite{Dalal06} captures spatial change of motion and is obtained by computing spatial derivatives along a discrete set of directions for both the horizontal and vertical components of the optical flow field. The employed trajectory shape, HOG, HOF, and MBH parameter values (e.g., discretization sampling) are the same as used in their previous applications to action recognition (e.g., \cite{Feichtenhofer16, Wang11}).\\
	
	The iFVs are computed by 
	(i) reducing the dimension of the features by a factor of two via Principal Component Analysis (PCA), 
	(ii) estimating a Gaussian Mixture Model (GMM) with $K=128$ components, representing each proposal by a $2DK$ FV, where $D$ is the dimension of the descriptor after PCA has been applied, and 
	(iii) converting to  iFVs by applying  power and L2 normalization to the FVs, cf. \cite{Wang13}. 
	The feature types are combined by concatenating their iFVs. 
	Finally, given such a test feature vector, a linear SVM \cite{Cortes95} (previously built on the training data with loss trade-off parameter set to $C=1$) assigns a biolocomotion probability score based on Platt scaling \cite{Platt99}. 

\end{subsection}

\begin{subsection}{Results}\label{sec:results}
	This section provides results and discussions on the proposed and alternative biolocomotion algorithm as evaluated on the BOLD dataset and CAD. 
	First, quantitative results, using AP scores, comparing the proposed to the alternative approach are provided. 
	Second, results that evaluate the performance of each component of the proposed and alternative approaches using PR curves are provided.
	Third, qualitative results from the proposed biolocomotion detection algorithm are provided.
	
	\begin{subsubsection}{Proposed Algorithm vs. Alternative Algorithm}

Table~\ref{tab:ap} shows AP scores as Intersection over Union (IoU), $\iota$, varies. The proposed algorithm maintains its superior performance over the alternative by a relatively large margin in the majority of the cases (i.e., difference of at least $0.1209$ for BOLD at $\iota\in[0.1,0.5]$ and CAD at $\iota\in\{0.1,0.2\}$), a small superiority in two cases (i.e., difference of 0.0028 and 0.0401 for CAD at $\iota=0.4$ and $\iota=0.5$, respectively) and performs slightly worse in a single case (i.e., difference of 0.0079 for CAD at $\iota=0.3$). More typically, the performance margin is in favour of the proposed approach (i.e., in nine of the ten cases) demonstrating the superiority of the proposed approach over the alternative algorithm.

\begin{table}[!h]
	\begin{center}
		\begin{tabular}{| c | c | c | c | c | c | c |} 
			\hline
			Dataset & \backslashbox{Algorithm}{$\iota$} & 0.1 & 0.2 & 0.3 & 0.4 & 0.5 \\
			\hline
			\multirow{2}{*}{BOLD}
			& Proposed & \textbf{0.6391} & \textbf{0.5362} & \textbf{0.4404} & \textbf{0.3362} & \textbf{0.2239} \\
			& Alternative \cite{vanGemert15} & 0.4566 & 0.3242 & 0.2294 & 0.1535 & 0.1030 \\
			\hline
			\multirow{2}{*}{CAD \cite{Bideau16}}
			& Proposed & \textbf{0.8014} & \textbf{0.5166} & 0.2458 & \textbf{0.1551} & \textbf{0.1275} \\
			& Alternative \cite{vanGemert15} & 0.4183 & 0.3262 & \textbf{0.2537} & 0.1523 & 0.0874 \\
			\hline
		\end{tabular}
	\end{center}
	\caption[AP scores for BOLD and CAD]{Average Precision (AP) scores as a function of Intersection over Union (IoU), $\iota$, for the BOLD dataset and CAD. \label{tab:ap}}
\end{table}

\end{subsubsection}

\begin{subsubsection}{Component analysis}

	\textbf{Proposed Algorithm.}\label{sec:results_my_ablation}
		The effect of each component in the biolocomotion algorithm (i.e., $\Lambda$, $\Omega$, $\Sigma$, and $E$) can be assessed through an ablation study; see Figure~\ref{fig:pr_ablation}. Here, the ablation study is designed to examine the effect of adding each component of the presented algorithm in sequence ($\Lambda$, $\Lambda+\Omega$, $\Lambda+\Omega+\Sigma$, and $\Lambda+\Omega+\Sigma+E$). A complete study that contains all $2^4-1$ combinations can be found in Appendix~\ref{app:results}. The ablation study is conducted only on the BOLD dataset, as it contains a better balanced set of biological and non-biological object videos as well as objects in locomotion and non-locomotion (as opposed to CAD). \\
		
		It can be observed in Figure~\ref{fig:pr_ablation}a that the recall values of the proposed method tend to be skewed to the left (i.e., ranging between $[0,0.4]$). This fact is largely due to the condition that labels a cluster  occupying the majority of the field of view as a cluster corresponding to camera motion; see Section \ref{sec:traj_postproc} and Appendix \ref{sec:traj_stab}. As a result, the number of missed pixels would increase, especially if the object occupies the majority of the frame, which in turn reduces the recall rate. \\
		
		It can be seen in Figure~\ref{fig:pr_ablation}b that adding each component increases the precision value with a slight decrease in recall for a specified $\tau_B$. Specifically, adding oscillation to locomotion (cf. navy and blue) improves the average of the precision differences by $+0.0383$; adding asymmetry to locomotion and oscillation (cf. blue and black) by $+0.0243$; and adding extrinsic motion dissimilarity to locomotion, oscillation, and asymmetry (cf. black and red) by $+0.0186$. The benefit of each component can be observed through the vertical shift upwards and is confirmed via positive average precision differences. The relatively low rate of improvement by adding the extrinsic motion dissimilarity component can be attributed to the low thresholds that were used (i.e., $\tau_E=0.105$ and $k_E=43.76$). This choice ensures that a relatively small number of FN is introduced. Nevertheless, this study confirms the benefits of each component in the proposed biomechanics and psychophysics driven algorithm.
		
\vspace{4pt}	\textbf{Alternative algorithm.}
		Figure~\ref{fig:pr_apt} provides an evaluation of the alternative algorithm to gauge the role of appearance, motion, and its combination by considering iFV-encodings of HOG (green), trajectory shape + HOF + MBH (blue), and trajectory shape + HOG + HOF + MBH (red), respectively. 
		It can be observed that considering appearance information alone (i.e., HOG) yields the most inferior result in detecting biolocomotion across both datasets. This result is potentially due to two possibilities: (1) There are so many in-class variations in appearance (e.g., humans, birds, fish, cars, and planes) that the considered training data does not suffice to capture all variations. (2) The detected biological object is not guaranteed to be locomoting. 
		Moreover, augmenting motion information with appearance information adds no apparent benefit. 
		These results confirm that eschewing appearance information and obtaining results strictly based on motion information is the desirable path to take when detecting biolocomotion in videos.

\newgeometry{top=1.25in,bottom=1.25in}
\begin{figure}[!h]
	\begin{center}
	\begin{tabular}{c}
		\includegraphics[width=0.63\textwidth]{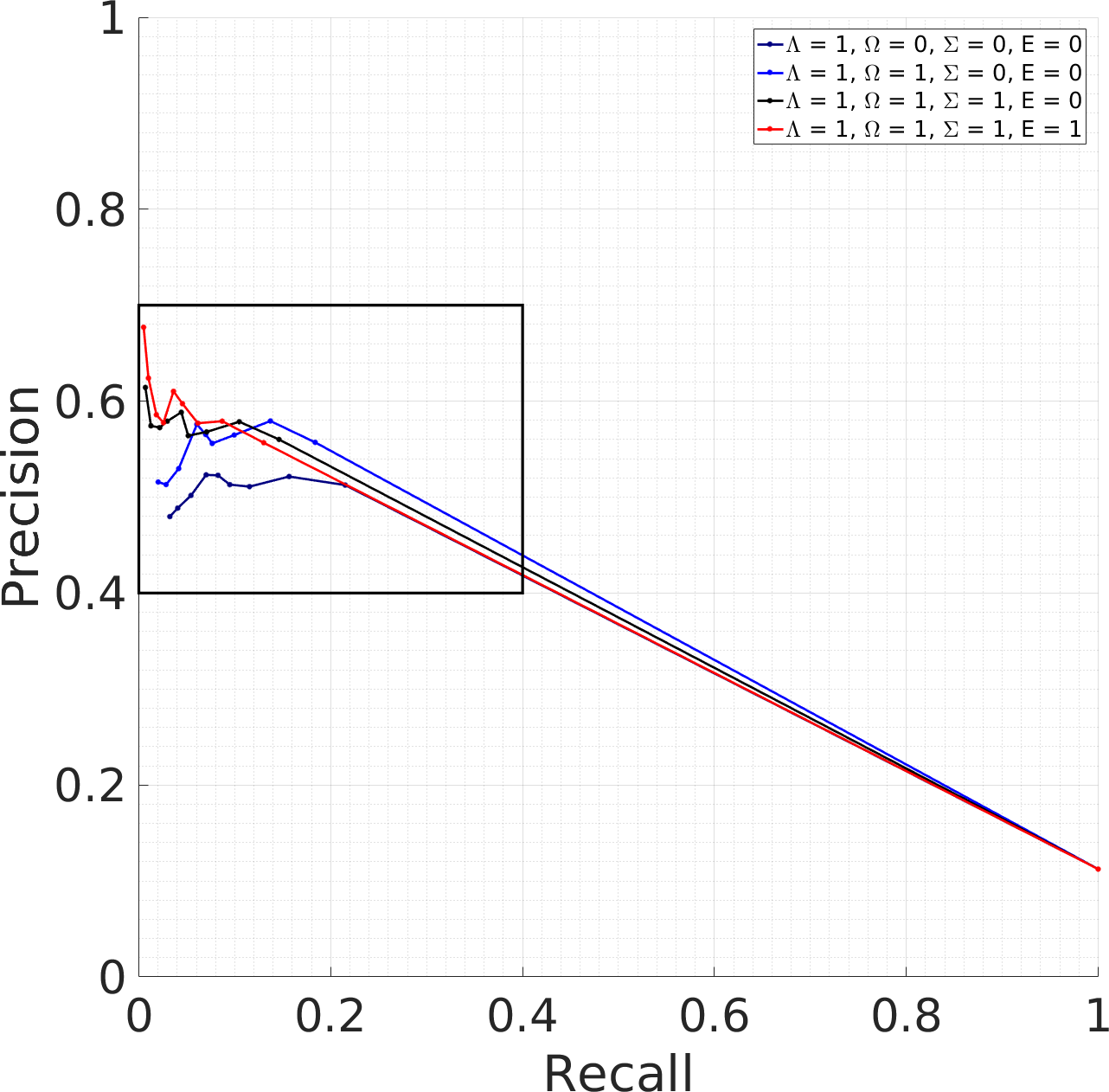}\\
		(a) PR curve in full scale. \vspace{3mm}\\
		 \includegraphics[width=0.63\textwidth]{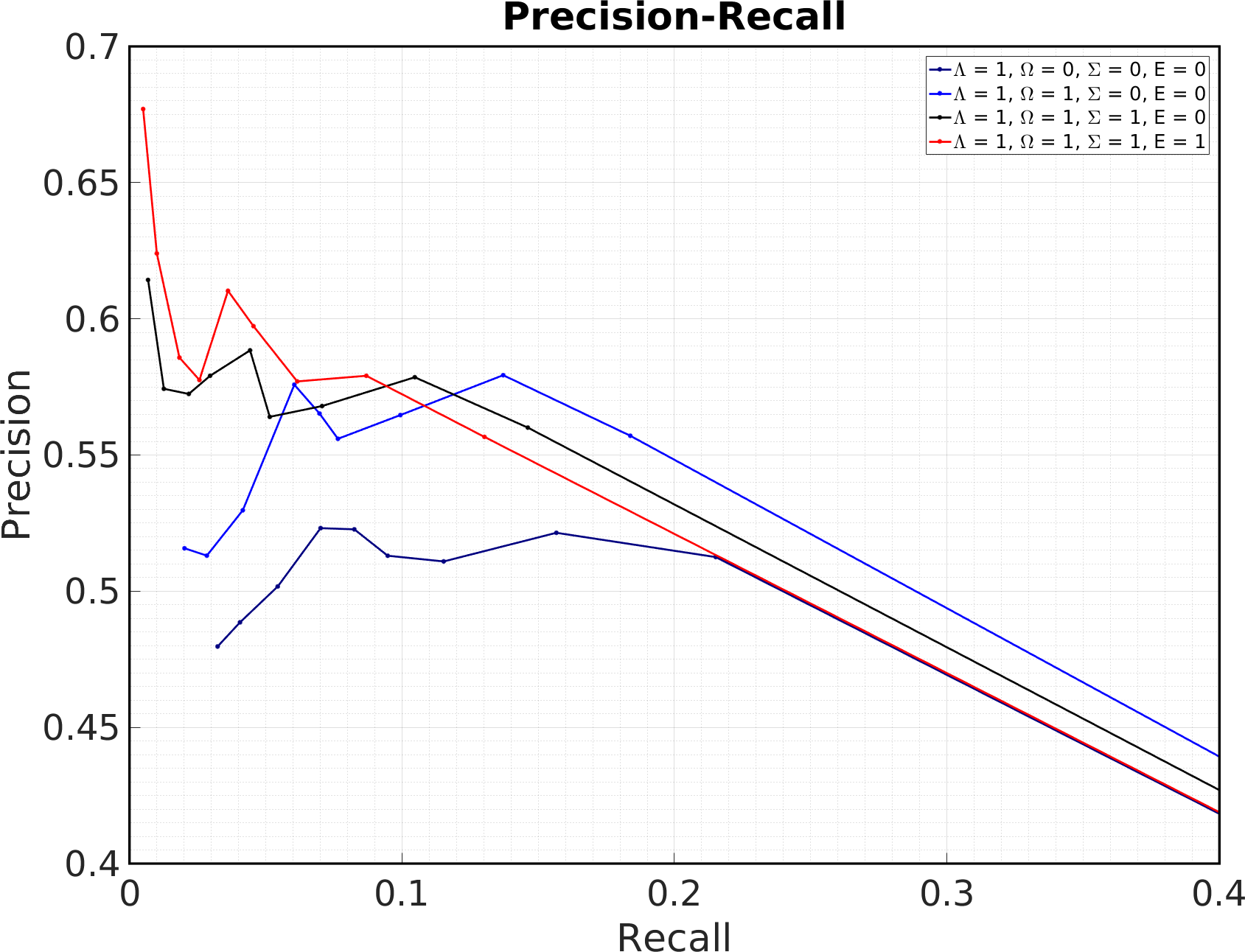}\\
		(b) PR curve for recall $\in [0,0.4]$ and precision $\in [0.4,0.7]$. \vspace{3mm}\\
	\end{tabular}
	\end{center}
	\caption[PR curve for the proposed algorithm]{Precision-Recall (PR) curves of the proposed algorithm. `1' and `0' indicate the enabled and disabled components, respectively. (a) a full PR curve is provided with both precision and recall in the ranges $[0,1]$ and (b) the same PR curve, as (a), restricted to recall values within $[0,0.4]$ and precision within $[0.4,0.7]$ as the majority of the points are focused in that select region of the PR curve. \label{fig:pr_ablation}}
\end{figure}
\restoregeometry
	
		
		\newgeometry{top=1.25in,bottom=1.25in}
		\begin{figure}[!h]
		\begin{center}
		\begin{tabular}{c}
			\includegraphics[width=0.63\textwidth]{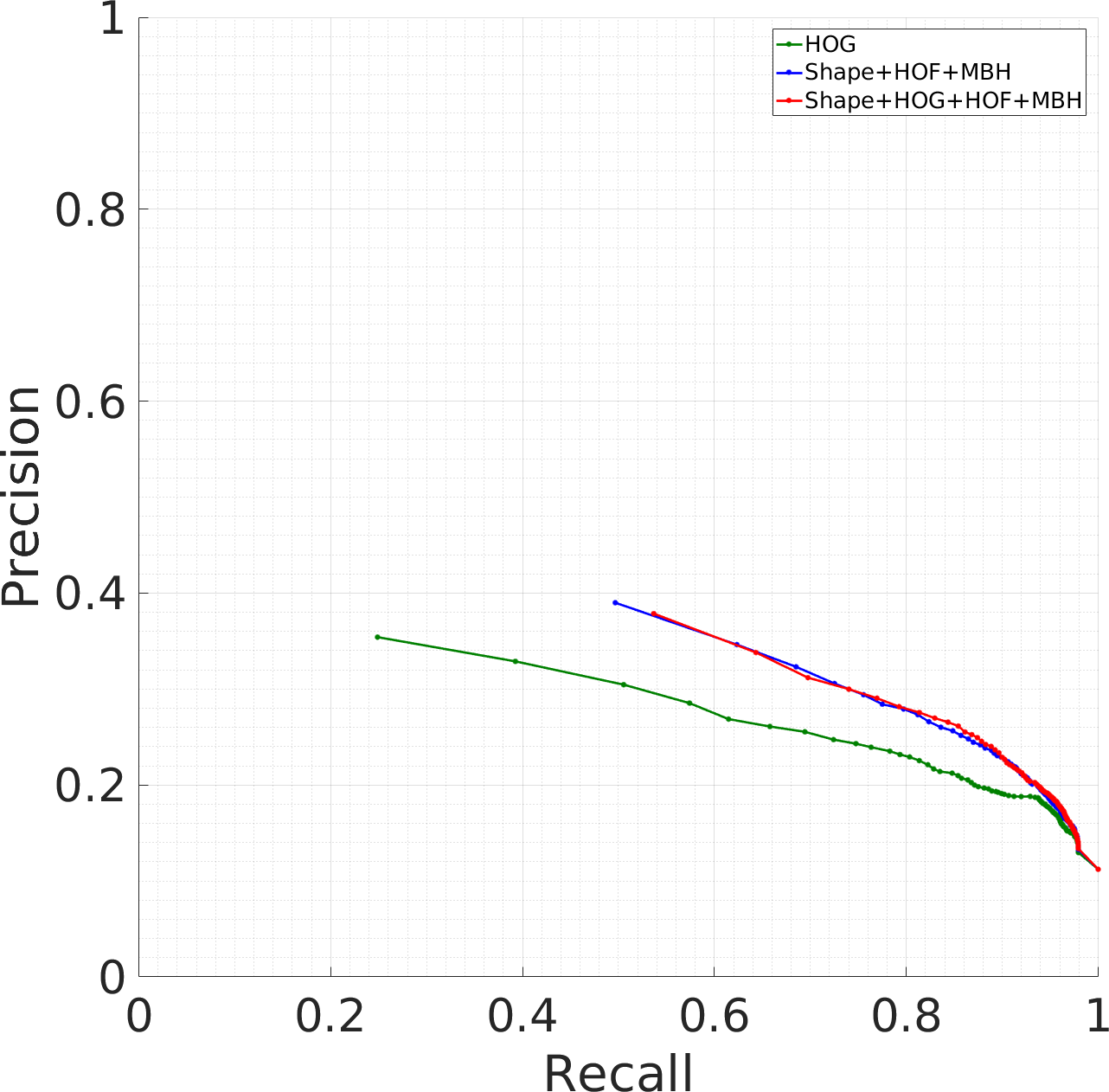} \\ 
		\end{tabular}
			\end{center}
			\caption[PR curve for the alternative algorithm]{Precision-Recall (PR) curves for alternative algorithm. \label{fig:pr_apt}}
		\end{figure}
		\restoregeometry
\end{subsubsection}

\begin{subsubsection}{Qualitative Results}
	Figure~\ref{fig:imageResults} provides qualitative detection results on a set of seven example frames from the BOLD dataset for the proposed algorithm. Various examples of locomoting biological objects are shown in Figure~\ref{fig:imageResults}a. While the species type varies (e.g., dog, bird, fish, and human), the way in which these species make a positional advance in the world (i.e., by running, flying, swimming, or climbing) all include oscillatory behaviour with extrinsic motion dissimilarity and an added asymmetry in its oscillation if moving orthogonal to the resistive force (i.e., running dog, flying bird, and swimming fish). As a result, they project oscillatory trajectories as they locomote inducing signs of biolocomotion. 
	On the contrary, a non-biological object (e.g., ball) locomoting is shown in Figure~\ref{fig:imageResults}b. While this object is making a spatial advance (i.e., by rolling) and even includes oscillation (in its cycloidal trajectory trace), since it does not exhibit \textit{asymmetric} oscillatory behaviour, it does not yield large $S_B$ values to indicate biolocomotion. 
	Conversely, an example of a biological object (bird) not locomoting is also shown in Figure~\ref{fig:imageResults}c. Even though the bird exhibits some oscillatory behaviour (i.e., jumping up and down), it does not locomote and hence is correctly not detected as biolocomotion.
	While the measure of locomotion, $\mathcal{M}_\Lambda$, is able to capture locomotion of (biological or non-biological) objects in the majority of the cases, it fails to capture objects primarily moving along the optical axis (i.e., towards (or away) from the camera); as depicted in Figure~\ref{fig:imageResults}d. As a result, a low signal for locomotion is obtained, resulting in a false negative biolocomotion detection. Future research can address this limitation by considering measurements that are indicative of change in the depth of an object within a temporal window.
	\end{subsubsection}
	
	\begin{subsection}{Summary}\label{sec:eval_conclusion}
This section has provided qualitative and quantitative results to validate the proposed algorithm's ability to detect biolocomotion and its superiority over an alternative approach that combines handcrafted features with learning. 
The results confirmed that biolocomotion can be modelled by measuring spatial advance of an object (locomotion) in conjunction with asymmetric oscillatory patterns and extrinsic motion dissimilarity. Furthermore, the results indicate that eschewing appearance information and relying solely on motion information in detecting biolocomotion yields superior performance. These studies were enabled by the introduction of the first dataset built for biolocomotion detection, BOLD.

\newgeometry{top=1.25in,bottom=1.25in}
\begin{figure}[!h]
	\begin{center}
	\resizebox{\linewidth}{!}{
		\begin{tabular}{c}
			\includegraphics[width=0.98\textwidth]{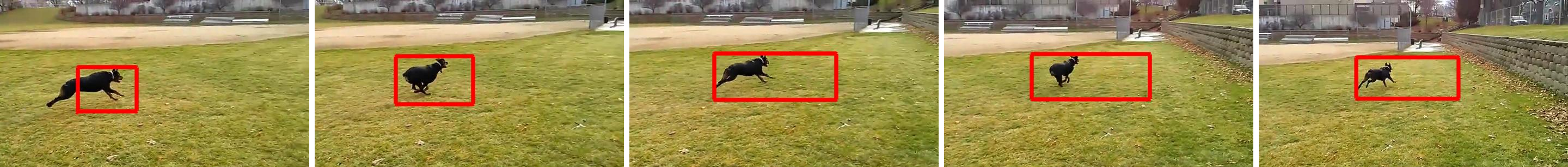}\\ 
			\includegraphics[width=0.98\textwidth]{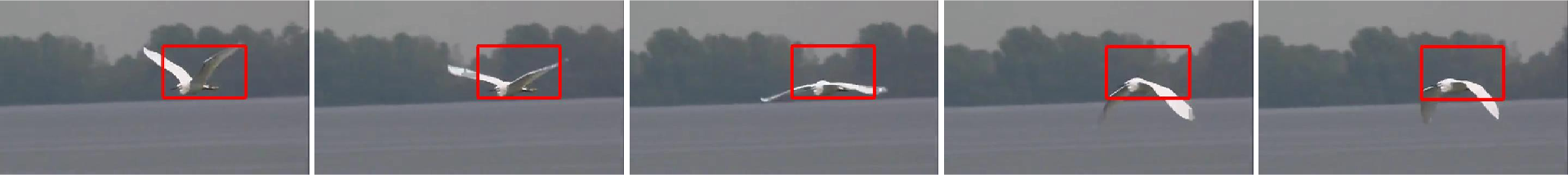} \\ 
			\includegraphics[width=0.98\textwidth]{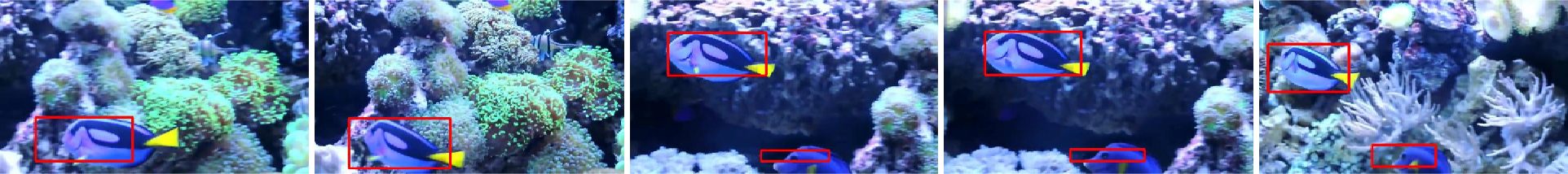} \\ 
			\includegraphics[width=0.98\textwidth]{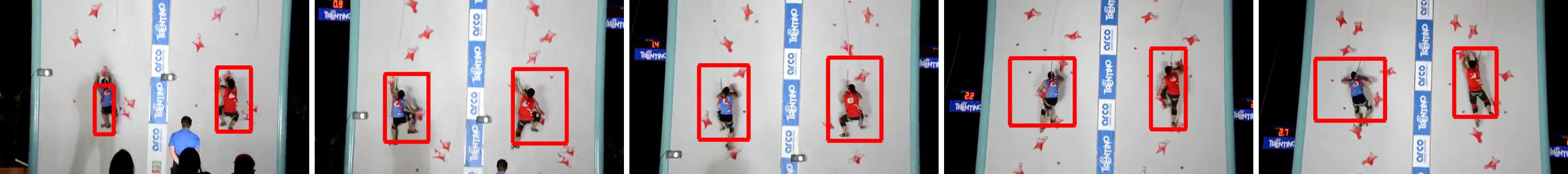}\\ 
			(a) Biological objects in locomotion \vspace{3mm} \\
			\includegraphics[width=0.98\textwidth]{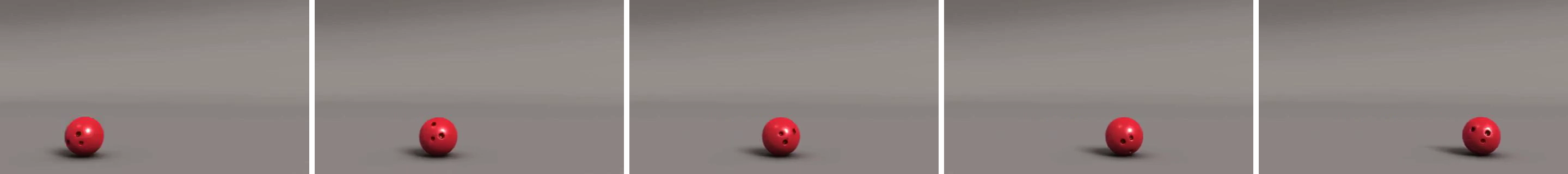} \\ 
			(b) Non-biological object in locomotion \vspace{3mm} \\
			
			\includegraphics[width=0.98\textwidth]{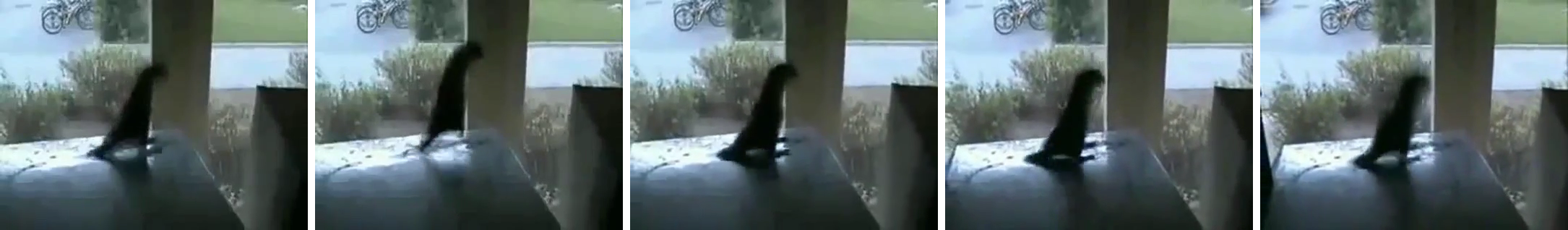} \\ 
			(c) Biological object not in locomotion \vspace{3mm} \\
			
			\includegraphics[width=0.98\textwidth]{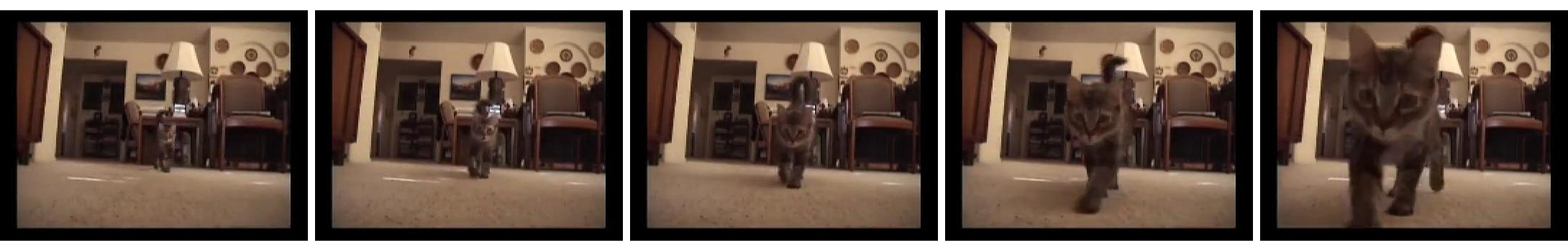} \\ 
			(d) Biological object in locomotion not detected as locomoting \\
		\end{tabular}
	}
	\end{center}	
	\vspace{-5mm}
	\caption[Qualitative results from the proposed biolocomotion algorithm]{Example results from the proposed biolocomotion detection algorithm. (a) Illustration of true detection results on different types of biological objects (dog, bird, fish, and human) locomoting in various ways (run, fly, swim, and climb). (b) Illustration of correctly undetected non-biological object (ball) in locomotion. (c) Illustration of correctly undetected biological object (bird) not in locomotion. (d) Illustration of a falsely undetected biological object (cat) in locomotion. 
	Detected regions are indicated by red bounding boxes that cover the trajectory clusters that led to the detection. 
} 
\label{fig:imageResults}
\end{figure}
\restoregeometry
\end{subsection}
\end{subsection}

\end{section}

\begin{section}{Conclusions}\label{ch:conclusion}
	\begin{subsection}{Overall summary}
This report has shed light on a new research topic in the field of computer vision: biolocomotion detection in videos. 
While previously not considered in computer vision, biolocomotion detection is an important topic not only because it presents an interesting and a well-defined challenge, but also for its wide spread applicability as a precursor for subsequent analysis (e.g., tracking and action recognition). Morever, research here has the potential to provide a computational model for the apparent ability of biological systems to detect biolocomotion from visual data. \\

Inspired by biomechanical properties of animals in locomotion and the perception of biological motion in psychophysics, a unified computational algorithm to detect biolocomotion in videos has been developed. In particular, the developed approach exploits the distinctive signature of asymmetric oscillatory pattern as biological entities make a positional advance, which also tends to yield dissimilar extrinsic and intrinsic motions. Since the developed algorithm is based on distinctive \textit{motion} patterns of locomoting biological objects, it enjoys the benefit of not needing to model the wide range of variations that exist between and within biological and non-biological objects. \\

A novel dataset, BOLD, has been assembled and biolocomotion annotations on BOLD as well as on an extant camouflage animals dataset have been supplied to provide evaluation benchmarks for the developed and future biolocomotion detection algorithms. 
To demonstrate the efficacy of the developed approach, an alternative approach based on generic handcrafted features with learning also has been developed and compared. Quantitative results indicate that the proposed algorithm considerably outperforms the alternative approach. These results support the hypothesis that biolocomotion can be detected reliably based on the biomechanically and psychophysically motivated signature of positional advance with asymmetric oscillation and extrinsic motion dissimilarity. 
\end{subsection}

\begin{subsection}{Future work}
In light of the work presented in this report, several directions for future work can be considered. \\

While the proposed approach has demonstrated its ability to detect a wide range of locomoting biological species in videos, it has also revealed its limitations in detecting objects that move parallel to the optical axis (i.e., towards (or away) from the camera, as in Figure \ref{fig:imageResults}d). An algorithm that can estimate the change in depth of such locomoting species could provide a solution to such scenarios. 
A potential approach could be based on measuring motion along the optical axis based on the relative rate of the visually apparent expansion (or contraction) of an object across time, as evidenced in biological systems (e.g., \cite{Regan78}) and employed in collision avoidance systems for autonomous vehicles (e.g., \cite{Kilicarslan18}). Combining the result with the currently proposed measurement of locomotion could better encompass the full three-dimensional displacement of biological species in motion.\\

One of the motivations for the computational study of biolocomotion detection is that it can serve as a precursor for many related tasks. 
Once a living creature has been detected, it can be recognized (e.g., human, cat, bird), its action can be classified (e.g., running, flying, jumping), or even more subtle distinctions can be made (e.g., gender, age, emotional state, personality trait) \cite{Troje08}. Correspondingly, an interesting direction for future research is to couple the biolocomotion detector with such subsequent processes. \\

Another source of motivation for the computational study of biolocomotion detection is that it might yield a computational theory of how biological systems solve this problem. Toward that end, one shortcoming of the current algorithm is that it is not subject to the inversion effect, whereby inverted biolocomotion displays are not easily recognized by animals \cite{Troje06,Vallortigara06}. An interesting direction for future research is to consider how this constraint could be incorporated into the detector and what advantages it might provide. Further, the specific mechanisms that are incorporated in the current detector could provide the basis for psychophysical experiments to further probe how biological systems operate.\\

Finally, various combinations of the components introduced in this report might be combined in different ways to detect more general types of motion, not restricted to locomotion. 
For example, studies have shown that animacy can be perceived from the motion of simple rigid geometric shapes, such as triangles, circles, and squares \cite{Heider44}; see Figure~\ref{fig:animacy_perception}. 
Therefore, future research can consider how the biolocomotion detection components defined in this report can be leveraged to detect more general animacy.
\\

\begin{figure}[!h]
	\begin{center}
		\resizebox{\textwidth}{!}{
			\begin{tikzpicture}
	
	\draw[ultra thick] (0,0) -- (5,0) -- (5,3.5) -- (0,3.5) -- (0,0); 
	\draw[thick] (3.5,2.5) -- (3.5,1) -- (1.5,1) -- (1.5,2.5) -- (2.5,2.5) -- (3.45,2.7); 
	\draw[fill=red, draw=none] (4,1) -- (4.7,1.9) -- (3.9,2.1) -- (4,1); 
	\draw[fill=blue, draw=none] (2,0.4) -- (2.5,0.2) -- (2.5,0.6) -- (2,0.4); 
	\draw[fill=ForestGreen, draw=none] (3,0.5) circle (0.25cm); 
	
	\draw[ultra thick] (5.5,0) -- (10.5,0) -- (10.5,3.5) -- (5.5,3.5) -- (5.5,0); 
	\draw[thick] (9,2.5) -- (9,1) -- (7,1) -- (7,2.5) -- (8,2.5) -- (8.95,2.7); 
	\draw[fill=red, draw=none] (8,0.5) -- (9.02,0.15) -- (8.94,0.94) -- (8,0.5); 
	\draw[fill=blue, draw=none] (6,1.5) -- (5.8,1.0) -- (6.2,1.0) -- (6,1.5); 
	\draw[fill=ForestGreen, draw=none] (6.5,0.5) circle (0.25cm); 
%
	\draw[ultra thick] (11,0) -- (16,0) -- (16,3.5) -- (11,3.5) -- (11,0); 
	\draw[thick] (14.5,2.5) -- (14.5,1) -- (12.5,1) -- (12.5,2.5) -- (13.5,2.5) -- (14.45,2.7); 
	\draw[fill=red, draw=none] (12,0.7) -- (12.9,0.15) -- (13.0,0.94) -- (12,0.7); 
	\draw[fill=blue, draw=none] (13,3) -- (12.5,2.8) -- (12.5,3.2) -- (13,3); 
	\draw[fill=ForestGreen, draw=none] (11.5,2.5) circle (0.25cm); 
\end{tikzpicture}
		}
	\end{center}
	\caption[Illustration of perception of animacy]{Illustration of perception of animacy. Select frames of rigid geometric shapes in motion displaying perception of animacy. The way geometric figures move provides sufficient information to give observers the perception of animacy. For example, an observer can interpret the display as the red triangle chasing after the blue triangle and the green circle moving around an obstacle (black opened box). Figure redrawn from \cite{Scholl00}. \label{fig:animacy_perception}} 
\end{figure}
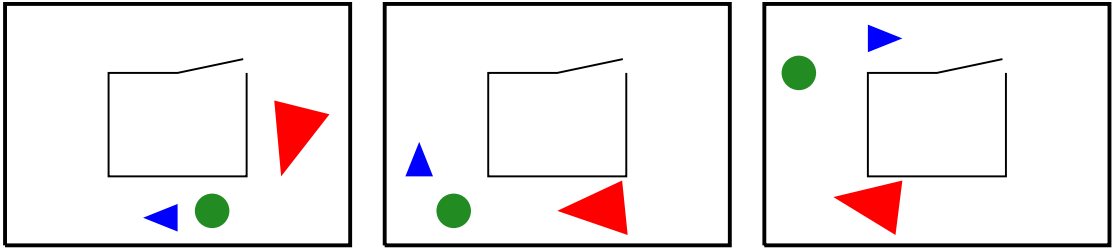

\end{subsection}

\end{section}

\section*{Acknowledgements}
The authors thank Robert Allison, Michael Brown, Marcus Brubaker, Dorita Chang and Nikolaus Troje for valuable comments on the work presented in this report.

\begin{appendices}
	\begin{section}{Trajectory processing}\label{app:traj_proc}
		Motivated by its state-of-the-art performance amongst handcrafted algorithms for action recognition, improved Dense Trajectories (iDTs) \cite{Wang13} are used as input to the proposed biolocomotion detector. 
For the best performance, trajectories had to be further processed prior to passing them to the biolocomotion detector, as follows. First, trajectories must be pruned to remove unuseful trajectories. Second, trajectories need to be clustered such that biolocomotion detection operates on sets of trajectories that are likely to correspond to a single or similarly moving entity in the world. Third, while iDTs are claimed to be robust to camera motion, trajectories need to be stabilized to (i) remove object-centric stabilization induced by the camera operator or (ii) remove residual motion arising from camera motion that contain corrupted data for measurement. Finally, select trajectories in a video can be elongated, without increasing its chance for drifting, to allow for more temporal support in biolocomotion detection. In this appendix, the details of each trajectory processing step is described.

\begin{subsection}{Pruning trajectories}\label{sec:traj_prune}
	To ensure that a meaningful set of trajectories are obtained, static trajectories that do not contain motion information and random trajectories are removed in the post-processing stage. While the need and conditions to remove static and random trajectories are addressed in the original iDT work \cite{Wang13}, their solution was insufficient for present purposes. In particular, the removal of static trajectories via standard deviation and the removal of random trajectories via arc length was insufficient. In addition, the removal of trajectories if the trajectory exceeded some pre-defined threshold was too aggressive in that they removed trajectories that corresponded to objects moving quickly (e.g., cars). Thus, more variations and less aggressive measures to prune static and random trajectories that would yield meaningful information in identifying biolocomotion are considered in this section.

	\begin{subsubsection}{Removing static trajectories}\label{sec:prune_static}
		Static trajectories can be removed using (i) standard deviation, (ii) arc length, and (iii) removing trajectories with $m$ relatively static points. Given a trajectory $\mathcal{T}^k$, its mean and standard deviation are defined as
		\begin{equation}\label{eq:mean_std}
			(\mu_x^k,\mu_y^k) = 
			\left( \frac{1}{L^k}\sum_{l=1}^{L^k}{x_l^k} , \frac{1}{L^k}\sum_{l=1}^{L^k}{y_l^k} \right)
			\end{equation}
			 and
		 \begin{equation}\label{eq:std}
			(\sigma_x^k,\sigma_y^k) =
			\left( \sqrt{\frac{1}{L^k} \sum_{l=1}^{L^k} |x_l^k - \mu_x^k|^2},
				\sqrt{\frac{1}{L^k} \sum_{l=1}^{L^k} |y_l^k - \mu_y^k|^2 } \right),
		\end{equation}
		respectively. 
		A trajectory is considered static if the standard deviation in the horizontal and vertical directions are below some threshold, $\tau_x$ and $\tau_y$ (i.e., if $(\sigma_x^k < \tau_x) \ \wedge \ (\sigma_y^k < \tau_y)$, then $\mathcal{T}^k$ is static). The thresholds can be set to some pre-defined values like $\tau_x = \tau_y  = \sqrt{3}$, which were empirically found to yield acceptable results. \\

		Trajectory $\mathcal{T}^k$ is considered static if its arc length, $\| \mathcal{T}^k \|$ as in \eqref{eq:arc_length}, is shorter than some threshold (i.e., if $\| \mathcal{T}^k\| < \tau$, then $\mathcal{T}^k$ is static). Again, $\tau$ can be set with some pre-defined value, like $\tau=10$, which has shown to yield good results.\\ 
		
		Lastly, a trajectory is considered static if it contains $m$ relatively static points. A point is considered static if its position has not changed in the next frame (i.e., $(|x_{l+1}^k-x_l^k| \approx 0) \wedge (|y_{l+1}^k-y_l^k|\approx 0)$). In practice, a small threshold, $\tau_s$, is assigned to determine if a point is \textit{relatively} static (i.e., $(|x_{l+1}^k-x_l^k| < \tau_s) \wedge (|y_{l+1}^k-y_l^k|<\tau_s)$). The number of relatively static points that a trajectory contains can be determined by summing the number of points that satisfy the relatively static condition. That is, if 
		\begin{equation}
			\sum_{l=1}^{L^k-1}{ \left[ (|x_{l+1}^k - x_n^k| < \tau) \ \wedge \ (|y_{l+1}^k - y_l^k| < \tau) \right] } > m \text{,}
		\end{equation}
		then $\mathcal{T}^k$ is considered static. Empirically, $\tau=1$ and $m=\frac{2}{3}L^k$ has shown to prune static trajectories well. 
	\end{subsubsection}
	
	\begin{subsubsection}{Removing Random Trajectories}
		Trajectories with sudden large displacements are likely to be erroneous. Thus, it would be helpful to remove them before further processing the data. Random trajectories can be removed using (i) standard deviation, (ii) arc length, and (iii) the `68-95-99.7' rule \cite{Wackerly08}. \\
		
		Using the standard deviation as defined in equation \eqref{eq:std}, a trajectory is likely to be erroneous if $(\sigma_x > \tau_x) \ \vee \ (\sigma_y > \tau_y)$, since it implies that $\mathcal{T}^k$ has a sudden large displacement in either the horizontal or vertical direction. Empirically, $\tau_x = \tau_y = 75$ has shown to prune random trajectories, while retaining non-random ones. \\
		
		Random trajectories are pruned using arc length as done in the original evaluation \cite{Wang13}. More specifically, if the displacement vector between two consecutive frames is larger than 70\% of the overall displacement, then it indicates a sudden large displacement, which is likely to be a random trajectory. Formally, if $\sqrt{(x_{l+1}^k - x_l^k)^2 + (y_{l+1}^k - y_l^k)^2} > 0.7 \| \mathcal{T}^k \| \ \forall \ k$, then $\mathcal{T}^k$ is likely to be a random trajectory. \\
		
		Finally, the standard deviation of the trajectories is used more adaptively via the `68-95-99.7' rule. That is, if 
		$(\mu_x - \tau_\sigma \sigma_x < x_l^k < \mu_x + \tau_\sigma \sigma_x) \ \wedge \ (\mu_y - \tau_\sigma \sigma_y < y_l^k < \mu_y + \tau_\sigma \sigma_y)$ for $1 \leq l \leq L^k$, it means that all the points of the trajectory lie within 68\%, 95\%, or 99.7\% of the data if $\tau_\sigma=1$, $2$, or $3$, respectively. Otherwise, a point is considered ``random'' within the trajectory. Empirically, $\tau_\sigma=3$ retained useful trajectories while removing random ones.
	\end{subsubsection}
\end{subsection}

\begin{subsection}{Clustering trajectories}\label{sec:traj_clstr}
	To ensure that the developed algorithm is applicable in the real world, it should be robust to various factors that could occur in videos from the wild, such as objects occuring simultaneously in a single frame (see Figure \ref{fig:multi_obj_fr}), presence of camera motion, etc. One of the ways in which these factors can be dealt with is to cluster them and subsequently handle them separately. Thus, in this section, a method to cluster trajectories is introduced followed by a way to identify and remove camera motion in videos. \\
	
	\begin{figure}[!h]
		\begin{center}
		\begin{tabular}{c c c}
			\includegraphics[width=0.3\textwidth]{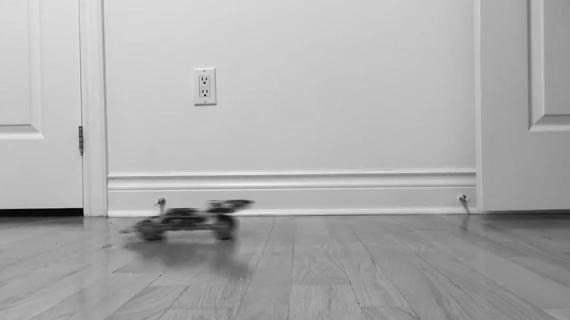}
			& \includegraphics[width=0.3\textwidth]{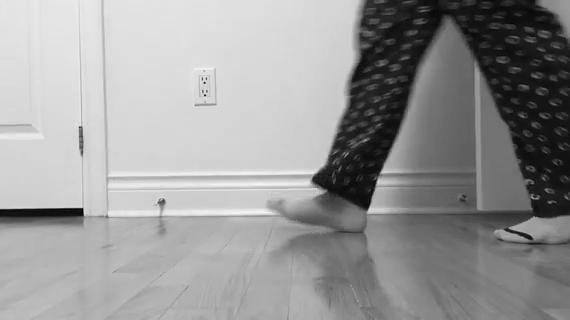}
			& \includegraphics[width=0.3\textwidth]{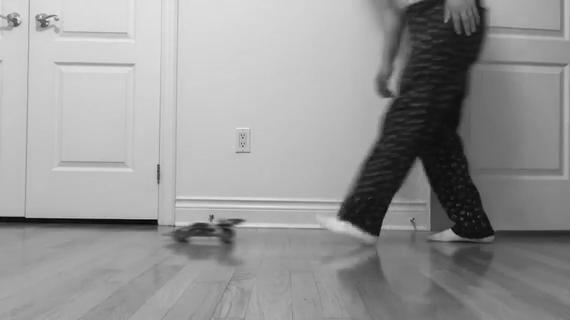} \\
			\multicolumn{2}{c}{(a) Single Object Per Frame} 
			& (b) Multiple Objects in a Frame \\
		\end{tabular}
		\end{center}
		\caption[Comparison of single object per frame vs. multiple objects per frame]{Comparison of single object per frame vs. multiple objects per frame.\label{fig:multi_obj_fr}}
	\end{figure}

Given a set of trajectories $\mathbf{\mathcal{T}} = \{\mathcal{T}^1, \dots, \mathcal{T}^N\}$, the goal is to cluster them into disjoint sets that correspond to similarly moving objects in the world. A variant on spectral trajectory clustering is to be employed since the original formulation \cite{Brox10} produced poor results with iDTs for the biolocomotion detection task. This shortcoming is likely due to the fact that the original intent of the algorithm was to cluster trajectories that were mostly linear, while the current goal is to cluster oscillatory trajectories. 
Correspondingly, novel measures of positional and shape affinity, $d_p$ and $d_s$, respectively, for trajectory pairs are defined to perform spectral clustering. The output of this processing stage are the centres as well as the horizontal and vertical extents of ellipses that cover each cluster. \\

	Let $\mathcal{T}^i$ and $\mathcal{T}^j$ represent two trajectories that span the same temporal length $L$ (i.e., $L=L^i=L^j$) in $\mathbf{\mathcal{T}}$. Then the \textit{proximity} of two trajectories can be measured by considering a weighted Euclidean distance between the average spatiotemporal coordinates of the trajectories,
	\begin{equation}\label{eq:d_sp}
		d_p(\mathcal{T}^i,\mathcal{T}^j) = 
			\alpha\left[\frac{ ( \bar{x}^i - \bar{x}^j)^2 }{w} + \frac{ (\bar{y}^i - \bar{y}^j )^2 }{h}\right]
			+ (1-\alpha) \left[ \frac{ (\bar{t}^i - \bar{t}^j)^2 }{f} \right]\text{,}
	\end{equation}
	where $\bar{x}^k$, $\bar{y}^k$, and $\bar{t}^k$ denote spatial and temporal averages, respectively, of trajectory $\mathcal{T}^k$ for $k \in \{i,j\}$. That is, $\bar{\mathbf{x}}^k = (\bar{x}^k,\bar{y}^k,\bar{t}^k) = (\frac{1}{L}\sum_{l=1}^{L}{x_l^k}, \frac{1}{L}\sum_{l=1}^{L}{y_l^k}, \frac{1}{L}\sum_{l=1}^{L}{t_l^k})$ with $\mathbf{x}_l^k = (x_l^k,y_l^k,t_l^k)$ the spacetime coordinates of the $l^{th}$ point in trajectory $k$ at frame $t_l$. To account for variations in resolution and frame rate of videos, the spatial distances are divided by the width, $w$, or height, $h$, and the temporal distance is divided by the frame rate, $f$. Greater emphasis on spatial distance over temporal distance (or vice versa) can be applied by the use of a constant term, $\alpha \in \mathbb{R}$, where $0 \leq \alpha \leq 1$; $\alpha=0.5$ is used in practice. The formulation in \eqref{eq:d_sp} ensures for a pair of spatiotemporally close trajectories $d_{p}(\mathcal{T}^i,\mathcal{T}^j) \rightarrow 0$, while those that are far tend to $\infty$. \\
	
	To compare the \textit{shape} of a pair of trajectories, $\mathcal{T}^i$ and $\mathcal{T}^j$, subsections of trajectories (or trajectory displacements) can be compared by considering either (i) the squared $L_2$-norm or (ii) the angle of the normalized inner product. While both are valid methods of comparison, considering the angle ensures that the overall \textit{shape} of the trajectory is compared while being minimally affected by the length of the trajectory displacements. Abstracting away the magnitude of the displacements between frames would ensure robustness to a potentially varying amount of displacement exhibited by different limbs of a biological object. That is, for displacement vector $\Delta \mathbf{x}_l^k$ of trajectory $\mathcal{T}^k$ at the $l^{th}$ point for $1 \leq l \leq L^k$ and $k \in \{i,j\}$, its angle of the normalized inner product of two vectors is defined as 
	\begin{equation}
		\Theta(\Delta \mathbf{x}^i_l, \Delta \mathbf{x}^j_l) 
		= \arccos \left( \frac{\langle \Delta \mathbf{x}^i_l, \Delta \mathbf{x}^j_l \rangle}
						  {\| \Delta \mathbf{x}^i_l \| \| \Delta \mathbf{x}^j_l \|} \right) 
		\text{.}
	\end{equation}
	The angle of the normalized inner product to consider is bound between $[0,180]$, where similar vectors tend to 0 while very different vectors tend to 180, instead of a bound between $[-1,1]$, where two alike vectors tend to 1 and two dissimilar vectors tend to $-1$. \\

	While the individual angular based difference between vectors could be combined across trajectories by simply taking their mean or maximum, taking a weighted sum according to the percentage of the overall trajectory length of the displacements is advantageous as longer portions of the trajectory will count more heavily. Correspondingly, a weight can be defined as 
	\begin{equation}
		w(\Delta \mathbf{x}_l^i, \Delta \mathbf{x}_l^j)
		= \frac{\| \Delta \mathbf{x}_l^i \|}{\sum_l \| \Delta \mathbf{x}_l^i \|} \frac{\| \Delta \mathbf{x}_l^j \|}{\sum_l \| \Delta \mathbf{x}_l^j \|}
		\text{.}
	\end{equation}

	Finally, while the overall shape difference between a pair of trajectories can be compared by taking the weighted sum of the individual similarities between vectors as 
	\begin{equation}
		d_{s}(\mathcal{T}^i,\mathcal{T}^j) 
		= \sum_{n=1}^{L-1}{ 
			w(\Delta \mathbf{x}_l^i,\Delta \mathbf{x}_l^i) \
			\Theta(\Delta \mathbf{x}_l^i, \Delta \mathbf{x}_l^j) } \text{,}
	\end{equation}
the phase shifts that could occur between opposing limbs (e.g., left and right arms of a human walker) must be accounted for. To obtain a phase invariant shape measure, the minimum of all phase shifts of $\mathcal{T}^j$ with respect to $\mathcal{T}^i$ are considered
\begin{equation}
	d_{s'}(\mathcal{T}^i, \mathcal{T}^j) = \min_{\iota=0}^{L-2}\left\{ 
									\sum_{l=1}^{L-1}{w\left(\Delta \mathbf{x}_l^i, \Delta \mathbf{x}_{l-\iota}^j\right)
												\Theta\left( \Delta \mathbf{x}_l^i, \Delta \mathbf{x}_{l-\iota}^j \right)} \right\}
	\text{,}
\end{equation}
which is used as the shape similarity between two trajectories. \\

The two traits that distinguish trajectories from another, $d_p$ and $d_{s'}$, can be transformed into affinities for spectral clustering via
\begin{equation}\label{eq:spectralClstr_trajs}
	W(\mathcal{T}^i,\mathcal{T}^j) = \exp{\left(-\kappa_s \cdot d_p(\mathcal{T}^i,\mathcal{T}^j) - d_{s'}(\mathcal{T}^i,\mathcal{T}^j)\right)} \text{,}
\end{equation}
where $\kappa_s \in \mathbb{R}$ is some constant, which is emprically set to $\kappa_s=0.01$.\\

Once the affinity matrix has been defined, spectral clustering can be employed. To compensate for potential noise in the defined affinity matrix, eigendecomposition is performed on a normalized Laplacian
\begin{equation}
	V^\top \Lambda V = D^{-\frac{1}{2}}(D-W)D^{-\frac{1}{2}}\text{,}
\end{equation} 
where $D$ is the degree matrix of $W$, and $V$ and $\Lambda$ correspond to matrices containing the eigenvector and eigenvalues, respectively \cite{Brox10}. 
Then $k$-means clustering is performed on the eigenvectors that correspond to $k$ smallest nonzero eigenvalues. \\

To allow spectral clustering to be as automated as possible, that is to limit the process of manually selecting the number of clusters for the $k$-means clustering step, a cluster validity index is employed. A cluster validity index is a measure that compares the compactness of a cluster (or the intra-class relationship between features \textit{within} a cluster) and the separatedness of clusters (or the inter-class relationship of features \textit{between} clusters). There are various cluster validity indices (e.g., Dunn \cite{Dunn74} and Davies-Bouldin \cite{Davies79}) that can be used to identify the ideal number of clusters \cite{Amorim15}. Here, \textit{silhouette} value \cite{Rousseeuw87} is employed.  \\

For simplicity, suppose $\mathcal{T}^i$ is a trajectory that belongs to cluster $\mathcal{C}^i$ and none other (i.e., $\mathcal{T}^i \in \mathcal{C}^i$ and $\mathcal{T}^i \notin \mathcal{C}^{k_i}$, where $\mathcal{C}^{k_i}$ denotes any cluster that is not $\mathcal{C}^i$ and $\cap_i{\mathcal{C}^{k_i}} = \emptyset$); see Figure \ref{fig:silhouettes}. Then the \textit{within} similarity of $\mathcal{T}^i$ to all the other trajectories in $\mathcal{C}^i$ can be calculated by taking the average similarity of $\mathcal{T}^i$ to all other trajectories in $\mathcal{C}^i$ according to
			\begin{equation}
				a(\mathcal{T}^i) = \frac{1}{|\mathcal{C}^i|} \sum_{\mathcal{T}^j}{ W(\mathcal{T}^i,\mathcal{T}^j)} \ \forall \ \mathcal{T}^j \in \mathcal{C}^i \text{,}
			\end{equation}
			where $W(\mathcal{T}^i,\mathcal{T}^j)$ is the similarity score between $\mathcal{T}^i$ and $\mathcal{T}^j$ as in \eqref{eq:spectralClstr_trajs}.
			The \textit{between} similarity of $\mathcal{T}^i$ to any other cluster $\mathcal{C}^{k_i}$ is calculated in a similar manner,
			\begin{equation}
				d(\mathcal{T}^i,\mathcal{C}^{k_i}) = \frac{1}{|\mathcal{C}^{k_i}|} \sum_{\mathcal{T}^j}{ W(\mathcal{T}^i,\mathcal{T}^j)} \ \forall \ \mathcal{T}^j \in \mathcal{C}^{k_i} \text{.}
			\end{equation}
			Once the \textit{between} similarity of $\mathcal{T}^i$ to all other clusters, $\mathcal{C}^{k_i} \ \forall i$, have been computed, the second best cluster for $\mathcal{T}^i$ is chosen by taking the maximum of the between clusters according to 
			\begin{equation}
				b(\mathcal{T}^i) = \max_{\mathcal{C}^{k_i} \ \forall i}{ \{ d(\mathcal{T}^i,C^{k_i}) \} }.
			\end{equation}\\
			\begin{figure}
	\begin{center}
	\begin{tikzpicture}
		
		\draw[dashed,color={rgb,255:red,255; green,128; blue,0}] (0,2) circle (1 cm) node[below=1cm] {$\mathcal{C}^i$};
		\node[circle,draw={rgb,255:red,255; green,128; blue,0},inner sep=0.5pt,scale=0.5] (x) at (0.5,2.2) {$\mathcal{T}^i$};
		\node[fill={rgb,255:red,255; green,128; blue,0},circle,scale=0.5] (A1) at (-0.25,2.4) {};
		\node[fill={rgb,255:red,255; green,128; blue,0},circle,scale=0.5] (A2) at (-0.25, 1.7) {};
		\draw[color={rgb,255:red,255; green,128; blue,0}] (x) -- (A1);
		\draw[color={rgb,255:red,255; green,128; blue,0}] (x) -- (A2);
		
		\draw[dashed,color={rgb:red,0;green,102;blue,204}] (6,4) circle(1.5 cm) node[below=1.5cm] {$\mathcal{C}^{k_1}$};
		\node[fill={rgb:red,0;green,102;blue,204},circle,scale=0.5] (B1) at (5.5, 4.5) {};
		\node[fill={rgb:red,0;green,102;blue,204},circle,scale=0.5] (B2) at (6.5, 4.5) {};
		\node[fill={rgb:red,0;green,102;blue,204},circle,scale=0.5] (B3) at (6,3.5) {};
		\draw[color={rgb:red,0;green,102;blue,204}] (x) -- (B1);
		\draw[color={rgb:red,0;green,102;blue,204}] (x) -- (B2);
		\draw[color={rgb:red,0;green,102;blue,204}] (x) -- (B3);
		
		\draw[dashed,color={rgb:red,0;green,102;blue,204}] (5,0) circle (0.5 cm) node[below=0.5cm] {$\mathcal{C}^{k_2}$};
		\node[fill={rgb:red,0;green,102;blue,204},circle,scale=0.5] (C1) at (5.2, 0.2) {};
		\node[fill={rgb:red,0;green,102;blue,204},circle,scale=0.5] (C2) at  (4.8, 0) {};
		\draw[color={rgb:red,0;green,102;blue,204}] (x) -- (C1);
		\draw[color={rgb:red,0;green,102;blue,204}] (x) -- (C2);
		
	\end{tikzpicture}
	\end{center}
\caption[Silhouettes]{Silhouettes \cite{Rousseeuw87}. The quality of the clusters is quantified using silhouette values. 
The \textit{within} similarity of $\mathcal{T}^i \in \mathcal{C}^i$, denoted $a(\mathcal{T}^i)$, is taken as the average similarity of $\mathcal{T}^i$ to all other trajectories in $\mathcal{C}^i$ (\textit{orange lines}). 
The \textit{between} similarity of $\mathcal{T}^i$ to other clusters $\mathcal{C}^{k_i}$ (\textit{blue lines}), denoted $d(\mathcal{T}^i,\mathcal{C}^{k_i})$, is calculated by taking the average similarity of $\mathcal{T}^i$ to all objects in cluster $\mathcal{C}^{k_i}$. 
The second most suitable cluster for $\mathcal{T}^i$ is chosen by selecting the cluster with the largest between similarity (e.g., $b(\mathcal{T}^i) = \max{\{ d(\mathcal{T}^i,\mathcal{C}^{k_1}),d(\mathcal{T}^i,\mathcal{C}^{k_2})\}}$). 
The \textit{within}, $a(\mathcal{T}^i)$, and \textit{between}, $b(\mathcal{T}^i)$, values are summarized into a single value using a silhouette value $s(\mathcal{T}^i)$. \label{fig:silhouettes}}
\end{figure}
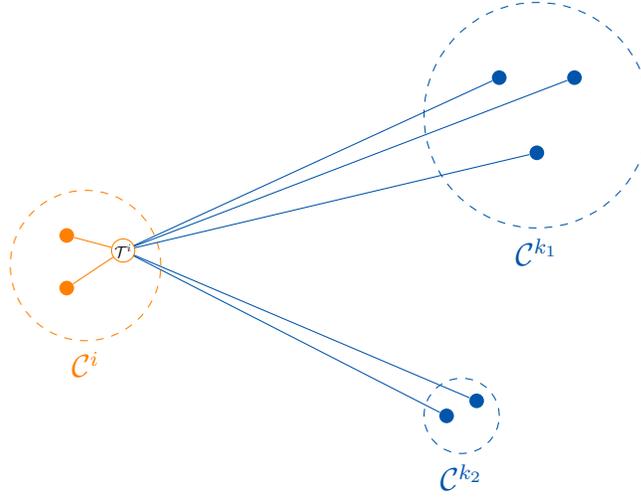
			
			The \textit{silhouette} of $\mathcal{T}^i$ is obtained by combining $a(\mathcal{T}^i)$ and $b(\mathcal{T}^i)$ according to
			\begin{equation}
				s(\mathcal{T}^i) = 
				\begin{cases}
					1 - \frac{a(\mathcal{T}^i)}{b(\mathcal{T}^i)}, & \text{if } a(\mathcal{T}^i) < b(\mathcal{T}^i) \\
					0, & \text{if } a(\mathcal{T}^i) = b(\mathcal{T}^i) \\
					\frac{b(\mathcal{T}^i)}{a(\mathcal{T}^i)} -1, & \text{if } a(\mathcal{T}^i) > b(\mathcal{T}^i) \\
				\end{cases}
				\text{,}
			\end{equation}
			which can be summarized into a single formula as
			\begin{equation}
				s(\mathcal{T}^i) = \frac{ a(\mathcal{T}^i) - b(\mathcal{T}^i) }{ \max{\{ a(\mathcal{T}^i),b(\mathcal{T}^i)} \}} \text{.}
			\end{equation}
			Since it is unclear how $a(\mathcal{T}^i)$ should be defined if there is only one trajectory in cluster $\mathcal{C}^i$, by convention, $s(\mathcal{T}^i)$ is set to 0. \\
			
			It is worth noting that $s(\mathcal{T}^i)$ is defined in a way such that $-1 \leq s(\mathcal{T}^i) \leq 1$. 
			$s(\mathcal{T}^i) \rightarrow 1$ if $a(\mathcal{T}^i)$ is large and $b(\mathcal{T}^i)$ is small, which implies that the \textit{within} similarity, $a(\mathcal{T}^i)$, is greater than the \textit{between} similarity, $b(\mathcal{T}^i)$, implying that $\mathcal{T}^i$ has been well-clustered. 
			$s(\mathcal{T}^i) \rightarrow 0$ if $a(\mathcal{T}^i) \approx b(\mathcal{T}^i)$, which implies it is unclear whether $\mathcal{T}^i$ should have been clustered into $\mathcal{C}^i$ or $\mathcal{C}^k$.
			$s(\mathcal{T}^i) \rightarrow -1$ if $a(\mathcal{T}^i)$ is small and $b(\mathcal{T}^i)$ is large, which implies that the \textit{within} similarity, $a(\mathcal{T}^i)$, is less than the \textit{between} similarity, $b(\mathcal{T}^i)$, implying that it would have been better to classify $\mathcal{T}^i$ into cluster $\mathcal{C}^{k_i}$ rather than $\mathcal{C}^i$. Hence, $\mathcal{T}^i$ has been misclassified. \\
			
			The quality of the final clustering result is evaluated by computing the \textit{overall average silhouette}, $S^k$, for $k$ clusters as follows
			\begin{equation}
				S^k = \frac{1}{N} \sum_{i=1}^{N}{ s(\mathcal{T}^i) \ \forall \mathcal{T}^i }\text{,}
			\end{equation}
			where $N$ is the number of trajectories in cluster $\mathcal{C}^k$. The cluster that yields the largest overall average silhouette, $S^k$, is indicative of the most optimal number of partitions. Silhouettes offer a quantitative analysis of the final outcome of a clustering algorithm without the knowledge of which clustering algorithm was actually used to acquire such result \cite{Rousseeuw87}. Since the most computationally expensive step of spectral clustering lies in the construction of the affinity matrix, $W$, numerous clusters can be evaluated, and the optimal number of clusters can be identified, rather efficiently, by selecting the cluster with the largest average silhouette width, $S$. \\


Finally, to recover the extent of each cluster, GMMs are used to model the shape of the clusters at frame $t$. Specifically, the points that are present at frame $t$ are represented by GMMs with varying number of components (e.g., two to five components) to yield a candidate of shapes for the cluster. Among the candidate of GMMs, the GMM with the smallest Akaike Information Criterion (AIC) value \cite{Akaike73} is selected to represent the cluster of points at frame $t$. To ensure temporal smoothness of the estimated GMM-based masks, each frame is weighted according to its neighbouring frames. That is, suppose $B_t^k$ represents a binary mask of cluster $k$ at frame $t$. Then its smoothed map is a weighted sum of its neighbouring frames 
\begin{equation}
	\tilde{B}_t^k = \sum_{t=1}^{T}{w(t,t') \ B_t^k}
\end{equation}
for $1 \leq t' \leq T$, where $w(t,t')=\exp\left( -\lambda (t-t')^2 \right)$ for constant $\lambda$, which controls the amount of contribution the neighbouring frames have on frame $t$. Empirically, $\lambda$ is set to 0.1. The weighted sum of the binary masks results in a real-valued map. Therefore, the final binary mask, $\mathcal{B}_t^k$, is defined as
\begin{equation}
	\mathcal{B}_t^k(\mathbf{x}) = 
	\begin{cases}
		1, & \text{if } \tilde{B}_t^k(\mathbf{x}) > \tau_d \\
		0, & \text{otherwise}
	\end{cases}
	\text{,}
\end{equation}
at some spatial coordinate $\mathbf{x}$, where $\tau_d$ is empirically set to $0.7$. \\

\end{subsection}

\begin{subsection}{Stabilizing trajectories}\label{sec:traj_stab}
While iDTs are designed to have a certain level of robustness to camera motion, the wide range of pans and zooms present in videos in-the-wild makes the entailed processing inadequate. Other recent approaches to image-based camera motion estimation and cancellation \cite{Park13} have also been inadequate. Thus, an approach to camera stabilization that corrects the overall global shift between frames (approximation to pan and tilt), linear expansion/contraction about a centre (approximation to zoom) as well as estimation of frames without camera motion is needed. In this section, a method to camera stabilization using trajectories is described. 
	
\begin{subsubsection}{Estimating Global Shift}
Trajectories that arise as a result of overall global translation (e.g., horizontal, vertical, and/or its combination) between frames should have similar shapes and therefore be grouped into a single cluster according to processing described in \S \ref{sec:traj_clstr}. Under the assumption that the field of view is not dominated by the object of interest (e.g., human, car), trajectories corresponding to camera motion would be well-dispersed throughout the field of view. Thus, the cluster whose trajectories occupy a widespread area of the frame can be taken as the cluster indicative of camera motion manifest as global shift. That is, cluster $\mathcal{C}^k$ containing a set of trajectories correspond to global shift if
\begin{equation}\label{eq:cammotion_cond}
	\left( \frac{w}{2} - v_x(t) < \tau_{cx} \right) \vee \left( \frac{h}{2} - v_y(t) < \tau_{cy} \right) \text{,}
\end{equation}
where $(v_x(t),v_y(t))$ correspond to horizontal and vertical extents of cluster $\mathcal{C}^k$ at frame $t$, while $w$ and $h$ correspond to the width and height of the image, respectively. The thresholds, $(\tau_{cx},\tau_{cy})$, to determine if the trajectories occupy the majority of the frame is defined as a ratio with respect to the resolution of the image (e.g., $(\tau_{cx},\tau_{cy})=(\frac{w}{15},\frac{h}{15})$).\\

Once the cluster of trajectories corresponding to camera motion has been identified, global shift can be estimated by calculating the median displacement of the trajectories on a frame-by-frame basis. That is, for displacement vector $\Delta \mathbf{x}^k_{(t,\Delta t)}$ of trajectory $\mathcal{T}^k$ between frames $t$ and $t-\Delta t$, global shift at frame $t$ is defined as
\begin{equation}
	(x_c(t),y_c(t)) = \left( \text{med}_{x^k(t)} \{\Delta x^k(t)\}, \text{med}_{y^k(t)} \{\Delta y^k(t)\} \right) \text{.}
\end{equation} 

Finally, with the global shift estimated, motion caused by camera motion can be removed by subtracting the estimated shift from each trajectory.
\end{subsubsection}

\begin{subsubsection}{Estimating Global Expansion/Contraction}
The displacement vectors in frames with global expansion (or contraction) motion (i.e., arising from camera zoom in (or zoom out)) exhibit unique patterns; see Figure~\ref{fig:zoom}. To handle frames with global expansion (or contraction), pattern of expansion (or contraction) must be detected, the scale of expansion (or contraction) imposed by the camera operator must be removed, then shifted to match the focus of expansion (or contraction). In the following, each of the steps necessary to remove global expansion (or contraction) will be detailed by describing the case where the focus of expansion (or contraction) is at the centre of image first, followed by an arbitrary focus of expansion (or contraction) location. \\

\begin{figure}[!h]
	\begin{center}
	\begin{subfigure}[b]{0.22\linewidth}
		\includegraphics[width=\linewidth]{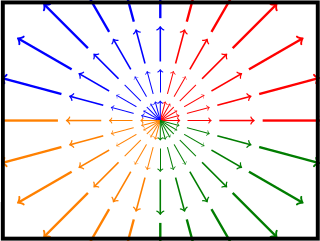}
		\caption{Expansion \label{fig:zoom_in} }
	\end{subfigure}
	~
	\begin{subfigure}[b]{0.22\linewidth}
		\includegraphics[width=\linewidth]{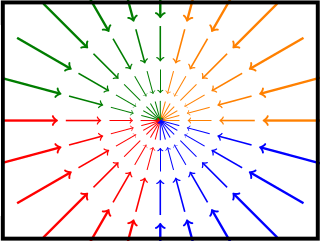}
		\caption{Contraction \label{fig:zoom_out} }
	\end{subfigure}
	~
	\begin{subfigure}[b]{0.5\linewidth}
		\begin{center}
			\includegraphics[width=0.44\linewidth]{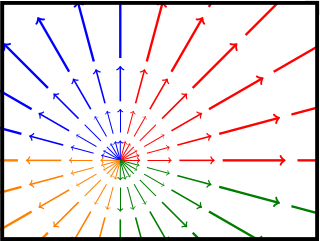}
			\includegraphics[width=0.44\linewidth]{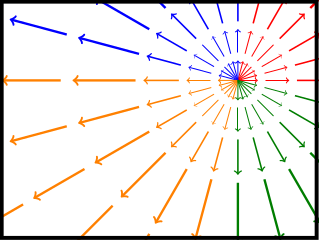}
		\end{center}
		\vspace{-10pt}
		\caption{Expansion - off centre \label{fig:zoom_offcentre}}
		\vspace{-5pt}
	\end{subfigure}
	\end{center}
	\caption[Illustration of trajectories in presence of linear expansion or contraction]{Illustration of trajectories in presence of linear expansion or contraction. The displacement vectors of each trajectory are grouped based on the quadrant that it belongs to (red for displacement vectors in quadrant $Q_I$, blue for $Q_{II}$, orange for $Q_{III}$ and green for $Q_{IV}$). A frame that contains expansion (i.e., zoom in) exhibits the following characteristic: $Q_{II}$ is to the left of $Q_I$, $Q_{III}$ is to the left of $Q_{IV}$, $Q_{II}$ is above $Q_{III}$ and $Q_I$ is above $Q_{IV}$. Conversely, a frame that contains contraction (i.e., zoom out) exhibits the following pattern: $Q_{IV}$ is to the left of $Q_{III}$, $Q_I$ is to the left of $Q_{II}$, $Q_{IV}$ is above $Q_{I}$ and $Q_{III}$ is above $Q_{II}$. \label{fig:zoom}}
\end{figure}

To identify the pattern indicative of expansion (or contraction), all displacement vectors present at frame $t$ must be assigned to a quadrant based on its orientation then subsequently consider the spatial arrangements of the assigned vectors. A displacement vector, $\Delta\mathbf{x}^k_{(t,\Delta t)}$, for trajectory $\mathcal{T}^k$ present at frame $t$ with a temporal difference of $\Delta t$ is assigned to a quadrant according to
\begin{equation}
\Delta\mathbf{x}^k(t) \in 
\begin{cases}
	Q_I, & \text{if } 0^\circ \leq \arctan{\left( \frac{\Delta y^k_{(t, \Delta t)}}{\Delta x^k_{(t, \Delta t)}} \right)} < 90^\circ \\
	Q_{II}, & \text{if } 90^\circ \leq \arctan{\left( \frac{\Delta y^k_{(t, \Delta t)}}{\Delta x^k_{(t, \Delta t)}} \right)} < 180^\circ \\
	Q_{III}, & \text{if } 180^\circ \leq \arctan{\left( \frac{\Delta y^k_{(t, \Delta t)}}{\Delta x^k_{(t, \Delta t)}} \right)} < 270^\circ \\
	Q_{IV}, & \text{if } 270^\circ \leq \arctan{\left( \frac{\Delta y^k_{(t, \Delta t)}}{\Delta x^k_{(t, \Delta t)}} \right)} < 360^\circ
\end{cases}
\text{.}
\end{equation}
The centre of each quadrant is calculated according to
\begin{equation}\label{eq:zoom_quad_centre}
	(x_q(t),y_q(t)) = \left( \frac{1}{|q|}\sum_{\forall \Delta x^k_{(t, \Delta t)} \in q}{\Delta x^k(t)} , \frac{1}{|q|}\sum_{\forall \Delta y^k_{(t, \Delta t)} \in q}{\Delta y^k(t)} \right)
\end{equation}
for $q\in\{Q_I,Q_{II},Q_{III},Q_{IV}\}$.
The pattern indicative of expansion (or contraction) is classified using the relative positions of the quadrant centres. A frame is considered to contain expansion if $Q_{II}$ is to the left of $Q_{I}$, $Q_{III}$ is to the left of $Q_{IV}$, $Q_{II}$ is above $Q_{III}$ and $Q_{I}$ is above $Q_{IV}$; see Figure~\ref{fig:zoom_in}. Conversely, a frame is considered to contain contraction if $Q_{IV}$ is to the left of $Q_{III}$, $Q_{I}$ is to the left of $Q_{II}$, $Q_{IV}$ is above $Q_{I}$ and $Q_{III}$ is above $Q_{II}$; see Figure~\ref{fig:zoom_out}. The state of expansion (or contraction) at frame $t$ can be determined according to
 \begin{equation}
 t_s =
	\begin{cases}
		1, & \text{if }
		\!\begin{aligned}
			&\left( x_{Q_{II}}(t) < x_{Q_{I}}(t) \right) \wedge \left( x_{Q_{III}}(t) < x_{Q_{IV}}(t) \right) \\
&\wedge \left( y_{Q_{III}}(t) < y_{Q_{II}}(t) \right) \wedge \left( y_{Q_{IV}}(t) < y_{Q_{I}}(t) \right) 
		\end{aligned}\\
		-1, & \text{if }
		\!\begin{aligned}
			&\left( x_{Q_{IV}}(t) < x_{Q_{III}}(t) \right) \wedge \left( x_{Q_{I}}(t) < x_{Q_{II}}(t) \right) \\
& \wedge \left( y_{Q_{IV}}(t) < y_{Q_{I}}(t) \right) \wedge \left( y_{Q_{III}}(t) < y_{Q_{II}}(t) \right)
		\end{aligned}\\
		0, & \text{otherwise}
	\end{cases}
	\text{,}
\end{equation}
where 
\begin{equation}
t_s=
\begin{cases}
	1 & \text{implies expansion} \\
	-1 & \text{implies contraction} \\
	0 & \text{neither expansion nor contraction} \\
\end{cases}
\text{.}
\end{equation}
\newline

The scale of expansion (or contraction), $\rho_t$, is determined by considering the magnitude of the displacement vector,
\begin{equation}
	\rho_t = \max_{\forall k |_t}{ \left\{ \| \Delta \mathbf{x}^k_{(t,\Delta t)} \| \right\} } \text{.}
\end{equation}
The largest magnitude is considered since it provides the most computationally stable measure of scale change between frames. For example, the error in an estimated scale change of 5 when the actual scale change is 5.1 is relatively small compared to an estimated scale change of 1 when the actual scale change is 0.9, even though the error is 0.1 in both cases. Here, reliance on a single (maximal) value is reasonable, because unreliable trajectories should already have been removed by the pruning in Section \ref{sec:traj_prune}. Consequently, the scale of expansion (or contraction) at frame $t$ is determined by comparing the ratio of the widths between frames (cf. Figure~\ref{fig:zoom_function}),
\begin{equation}
	s_t = \frac{w_t}{w'_{t+1}} =
	\begin{cases}
		\frac{w_{t}}{w_t + 2\sqrt{\rho_t}}, & \text{if } t_s = 1 \\
		\frac{w_{t}}{w_t-2\sqrt{\rho_t}}, & \text{if } t_s = -1\\
		1, & \text{if } t_s = 0\\
	\end{cases}
	\text{,}
\end{equation}
where $w_t$ is the width at frame $t$ and $w'_{t+1}$ is the estimated width at frame $t+1$. Note that the calculation for computing the scale of expansion (or contraction) is not limited to  widths, but can be found in a similar manner using heights.
Finally, the scale of expansion (or contraction), $s_t$, is used to remove the expansion (or contraction) imposed by the camera operator by scaling each of the tracked point, $\mathbf{x}^k$, of trajectory $\mathcal{T}^k$ at frame $t$ according to
\begin{equation}
	\hat{\mathbf{x}}^k(t) = s_t \mathbf{x}^k(t) \text{.}
\end{equation}

\begin{figure}
	\begin{center}
		\begin{subfigure}[b]{0.45\linewidth}
			\includegraphics[width=\linewidth]{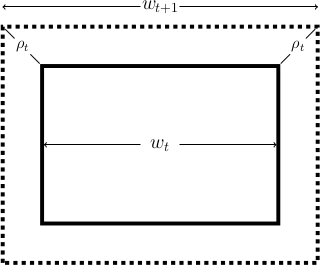}
			\caption{Expansion \label{fig:zoom_scale_in} }
		\end{subfigure}
		~
		\begin{subfigure}[b]{0.45\linewidth}
			\includegraphics[width=\linewidth]{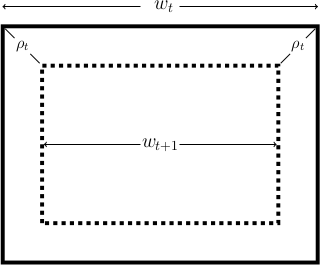}
			\caption{Contraction \label{fig:zoom_scale_out} }
		\end{subfigure}
	\end{center}
	\caption[Scale factor]{Scale factor. The scale factor between frames can be determined using the width and the maximum magnitude of the displacement vectors by comparing the width of the old frame (solid rectangle) to the new frame (dashed rectangle). \label{fig:zoom_function}}
\end{figure}

In the general case, where the focus of expansion (or contraction) is not at the centre of the image, as shown in Figure~\ref{fig:zoom_offcentre}, multiplying each frame by $s_t$ will incorrectly remove the expansion (or contraction) and further distort the imagery. Thus, the centre of the expansion (or contraction), $\mathbf{x}_z(t) = \left( x_z(t),y_z(t) \right)$, must be taken into account, which is computed using the median of the quadrant centres, \eqref{eq:zoom_quad_centre}, according to
\begin{equation}
	\mathbf{x}_z(t) = \left(\text{med}_{\forall q}{\{x_q(t)\}},\text{med}_{\forall q}{\{y_q(t)\}}\right) \text{.}
\end{equation}

Finally, once the scale change and the shift in zoom centre has been determined, expansion (or contraction) is reverted from the recovered trajectories by multiplying the recovered iDTs by the scale change, then shifting by the centre,
\begin{equation}
	\hat{\mathbf{x}}^k(t) = s_t \mathbf{x}^k(t) - \mathbf{x}_z(t)
	\text{.}
\end{equation} 
\end{subsubsection}

\begin{subsubsection}{Frames without Camera Motion}
	Presence of camera motion can be identified by detecting trajectories that are likely to correspond to camera motion. Conversely, absence of camera motion can be identified by detecting trajectories that are likely to correspond to static motion. Frames that are predominantly occupied by static trajectories (see Section \ref{sec:prune_static}) are likely to correspond to frames without camera motion. That is, suppose $\{(\acute{x}^k(t),\acute{y}^k(t)) | \acute{\mathcal{T}}^k\}$ denotes a set of points present at frame $t$ that belong to static trajectory $\acute{\mathcal{T}}^k$. Frame $t$ does not contain camera motion if
	\begin{equation}\label{eq:staticfr_cond}
		\left( \frac{w}{2} - \acute{v}_x(t) < \acute{\tau}_x \right) \wedge \left( \frac{h}{2} - \acute{v}_y < \acute{\tau}_y \right) \text{,}
	\end{equation}
	where $(\acute{\tau}_x,\acute{\tau}_y)$ correspond to horizontal and vertical vertices of the static trajectories that occupy the frame at frame $t$. The vertices of the static trajectories is estimated by taking the standard deviation, $(\acute{v}_x(t),\acute{v}_y(t)) = (2\acute{\sigma}_x(t),2\acute{\sigma}_y(t))$, of the points that are present at frame $t$. The thresholds, $(\acute{\tau_x},\acute{\tau_y})$, are defined in terms of the resolution of the video (e.g., $(\acute{\tau}_x,\acute{\tau}_y)=\left(\frac{w}{15},\frac{h}{15}\right)$). The \verb+AND+ condition is used in \eqref{eq:staticfr_cond} instead of \verb+OR+ as in \eqref{eq:cammotion_cond}, because \eqref{eq:staticfr_cond} will null frames that were originally identified to contain camera motion. Thus, a stronger condition, $\wedge$, is applied to ensure that frames with camera motion are not incorrectly identified as a static frame. These frames are further confirmed as static if the absolute of the average displacement of the static trajectories are less than some threshold. That is, if
	\begin{equation}
		\left( \left| \frac{1}{N(t)}\sum_{k=1}^{N(t)}{\Delta \acute{x}^k_{(t,1)}} \right| < \acute{\tau}_x \right)
		\wedge
		\left( \left| \frac{1}{N(t)}\sum_{k=1}^{N(t)}{\Delta \acute{y}^k_{(t,1)}} \right| < \acute{\tau}_y \right)
		\text{,}
	\end{equation}
	where $N(t)$ is the number of static trajectories that are present at frame $t$, and $(\acute{\tau}_x,\acute{\tau}_y)=(0.1,0.1)$ is some small pre-defined value, then frame $t$ is a static frame.
\end{subsubsection}

\begin{subsubsection}{Integrated Trajectory Stabilization}
	To stabilize a general set of input trajectories that may contain global shift (pan/tilt), linear expansion/contraction (zoom in/out), its combination or neither, a way to integrate global shift, expansion/contraction estimations and cancellations is considered, as follows. First, all frames are checked for absence of camera motion. Frames that meet the diagnostics of no camera motion are not processed further for trajectory stabilization. Second, all remaining frames are processed to stabilize the trajectories against global shift (pan/tilt). Third, the frames are processed to stabilize the trajectories against expansion/contraction (zoom in/out).
\end{subsubsection}
\end{subsection}

\begin{subsection}{Elongating trajectories}\label{sec:traj_cat}
	
To overcome the drifting effect when tracking points, trajectories estimated as iDTs are recommended to span $L=15$ or $20$ frames for videos with a frame rate of 30 fps \cite{Wang13}. This limitation yields short-term trajectories, which are often insufficient to obtain a reliable measurement of trajectory characteristics, such as oscillation amplitude. Thus, a more reliable set of trajectories for the biolocomotion detector are obtained by concatenating iDTs across frames to gain elongated trajectories. In particular, a set of trajectories that span a fixed number of frames is concatenated by considering (i) the spatiotemporal proximities and (ii) the appearance alikeness of the trajectories. 
That is, for trajectory $\mathcal{T}^k(t)$ that begins at frame $t$ and spans $L^k$ frames, the proximity criterion is considered by seeking trajectories that \textit{begin} at frame $s \in [t+L^k,t+L^k+\delta t]$ for some constant $\delta t$, whose spatial distance between the estimated point of $\mathcal{T}^k$ at frame $s$ and the first point of $\mathcal{T}^j$ is less than some pre-defined constant, $\tau_{sp}$, according to
\begin{equation}\label{eq:cattraj_space}
	d_{sp}\left( {\mathbf{\hat{x}}_{L^k+dt}^k}, \mathbf{x}_1^j \right) < \tau_{sp} \text{,}
\end{equation}
where ${\mathbf{\hat{x}}_{L^k+dt}^k}$ is an estimated point of $\mathcal{T}^k$ at frame $s=t+L^k+dt$ for $0\leq dt \leq \delta t$.
The spatial distance between two points, $\mathbf{x}_i^j$ and $\mathbf{x}_l^k$, is defined using the $L_2$-norm according to
\begin{equation}
d_{sp}(\mathbf{x}_i^j,\mathbf{x}_l^k) 
	= d_{sp}\left( (x_i^j,y_i^j),(x_l^k,y_l^k) \right) 
	= \sqrt{(x_l^k - x_i^j)^2 + (y_l^k - y_i^j)^2 } \text{.}
\end{equation}
The appearance similarity between two trajectories can be considered using HOGs \cite{Dalal05}. That is, if the distance between the corresponding HOGs, $H^k$ and $H^j$ for trajectories $\mathcal{T}^k$ and $\mathcal{T}^j$, respectively, is less than some pre-defined constant, $\tau_H$: 
\begin{equation}\label{eq:cattraj_hist}
	d\left( H^k,H^j \right) < \tau_H \text{,}
\end{equation}
then trajectories $\mathcal{T}^k$ and $\mathcal{T}^j$ are likely to correspond to the same point of an object.
The distance between HOGs, $d(H^k,H^j)$, is defined using the $L_2$-norm according to
\begin{equation}
	d(H^k,H^j) = \sqrt{ \sum_i{\left( H^j(i) - H^k(i) \right)^2} } \text{,}
\end{equation}
where $H^k = [H^k(i)] \in \mathbb{R}^{n}$.
Trajectory $\mathcal{T}^j$ that satisfies both \eqref{eq:cattraj_space} and \eqref{eq:cattraj_hist} is concatenated with trajectory $\mathcal{T}^k$. \\

Trajectory concatenation is not limited to the connection of two constant length trajectories, but can be applied iteratively to concatenate arbitrarily many trajectories provided \eqref{eq:cattraj_space} and \eqref{eq:cattraj_hist} are satisfied. Once two trajectories are concatenated, HOG is updated to the trajectory that occurred (temporally) later. That is, if $\mathcal{T}^{kj}$ denotes the concatenation of trajectories $\mathcal{T}^k$ and $\mathcal{T}^j$, where $\mathcal{T}^k$ occurred earlier than $\mathcal{T}^j$, then $H^{kj}=H^j$.\\

Currently, constant value of $\delta t = \frac{f}{6}$ for frame rate $f$ is used as the temporal threshold, $\tau_{sp} = 2.5 * S$ for sampling width $S$ as the threshold for the spatial distance, $\tau_H = 0.75$ as the threshold for appearance alikeness, and $n = 96 = 8 \times 2 \times 2 \times 3$ for the dimension of the HOG descriptor.

\end{subsection}
	\end{section}
	
	\begin{section}{Alternative oscillation measures}\label{app:amp}
		In addition to the frequency-based calculation presented in the main report (Section~\ref{sec:E_osc}) to obtain the oscillation measure, $\mathcal{M}_\Omega$, the amplitude of a detrended trajectory can also be calculated using integrals and differentials. \\

The amplitude of a trajectory can be computed via \textit{integrals} by considering the average of the maximum value and the absolute of the minimum value of the detrended trajectory if the overall integral is close to zero; see Figure~\ref{fig:amplitude}a. That is, for a detrended trajectory $\tilde{\mathcal{T}}^k$, its amplitude is defined as the average of the maximum and minimum values of $\mathcal{\tilde{T}}^k$, according to
\begin{equation}\label{eq:amp_int}
	a_\Sigma^k = \exp{\left( -\frac{\left(\sum_l \tilde{y}^k_l \right)^2}{2} \right)}{\left( \frac{\max_l{\tilde{y}^k_l}+|\min_l{\tilde{y}^k_l}|}{2} \right)} \text{.}
\end{equation}
The exponential function in \eqref{eq:amp_int} ensures that the amplitude is more recognized as the overall sum is close to zero, hence the use of an exponential weighting function. \\

Alternatively, the amplitude of a trajectory can be evaluated using \textit{differentials} by considering the average of vertical values where its slope is horizontal; see Figure~\ref{fig:amplitude}b. That is, the amplitude of a detrended trajectory $\tilde{\mathcal{T}}^k$ is defined as a weighted sum of vertical values whose first-order derivative is near zero according to
\begin{equation}
	a_\Delta^k = w_c \frac{ \sum_l w_s \tilde{y}_l^k }{ \sum_l w_s } \text{,}
\end{equation}
where $w_s$ is a weight that accounts for the slope and $w_c$ is a weight that accounts for the periodicity of $\tilde{\mathcal{T}}^k$. These weights are defined using Gaussian-like functions, as follows. \\

To ensure a larger weight is assigned to values with near horizontal slopes (i.e., $(\tilde{\mathcal{T}}^k)' \approx 0$), a Gaussian-like function centred at 0 with a height of 1 is used to define $w_s$ according to
\begin{equation}
w_s = \exp{\left(-\frac{ [(\tilde{y}^k_l)']^2 }{ 2(0.1)^2 } \right)} \text{.}
\end{equation}
$w_s$ assigns a value close to 1 if the slope of $\tilde{\mathcal{T}}^k$ is near zero. Empirically, a width of 0.1 was found effective for accepting how horizontal $\tilde{\mathcal{T}}^k$ is. \\

To ensure periodicity of $\tilde{\mathcal{T}}^k$, $w_c$ is a function defined to favour even number of concavities. Thus, $w_c$ is defined as the sum of second-derivative of $\tilde{\mathcal{T}}^k$ using a Gaussian-like function centred at 0 with a height of 1 according to
\begin{equation}
w_c = \exp{ \left(-\frac{[\sum_l w_s (\tilde{y}^k_l)'')]^2}{2}\right) }\text{.}
\end{equation}
$w_c$ assigns a value close to 1 when there are even number of concavities (i.e., the sum of the second derivative of $\tilde{\mathcal{T}}^k$ is approximately $0$).
Empirically, a width of 1 was found effective for determining the concavity evenness of $\tilde{\mathcal{T}}^k$. \\

\begin{figure}
	\begin{center}
		\begin{tabular}{c c}
			\includegraphics[width=0.4\textwidth]{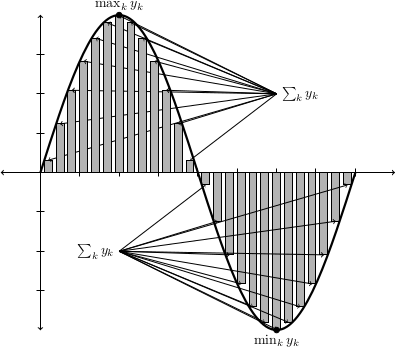} 
			& \includegraphics[width=0.48\textwidth]{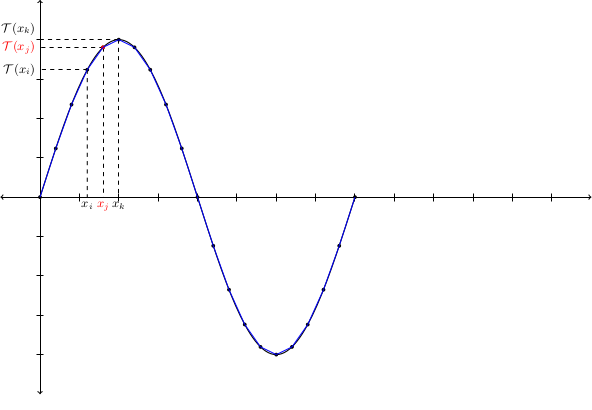} \\
			(a) Integral-based & (b) Differential-based \\
		\end{tabular}
	\end{center}
	\caption[Alternative measures of amplitude]{Alternative measures of amplitude. See accompanying text in this appendix for details. \label{fig:amplitude}}
\end{figure}

Based on preliminary experiments, the \mbox{frequency-}based approach was selected over the \mbox{integral-} and differential-based approaches for the work considered in this report.
	\end{section}
	
	\begin{section}{Additional empirical results}\label{app:results}
		In addition to the quantitative results presented in the main report for the proposed biolocomotion detection algorithm (Section \ref{sec:results_my_ablation}), this appendix presents more detailed results on the performance of individual algorithm components and their combinations on the BOLD dataset. Figure~\ref{fig:pr_ablation_full} presents PR curves for all algorithm component combinations. Table~\ref{tab:pr_ablation_full} shows precision values as detection threshold varies for all algorithm component combinations. 

\newgeometry{top=1.25in,bottom=1.25in}
\begin{figure}[!h]
	\begin{center}
	\begin{tabular}{c}
		\includegraphics[width=0.63\textwidth]{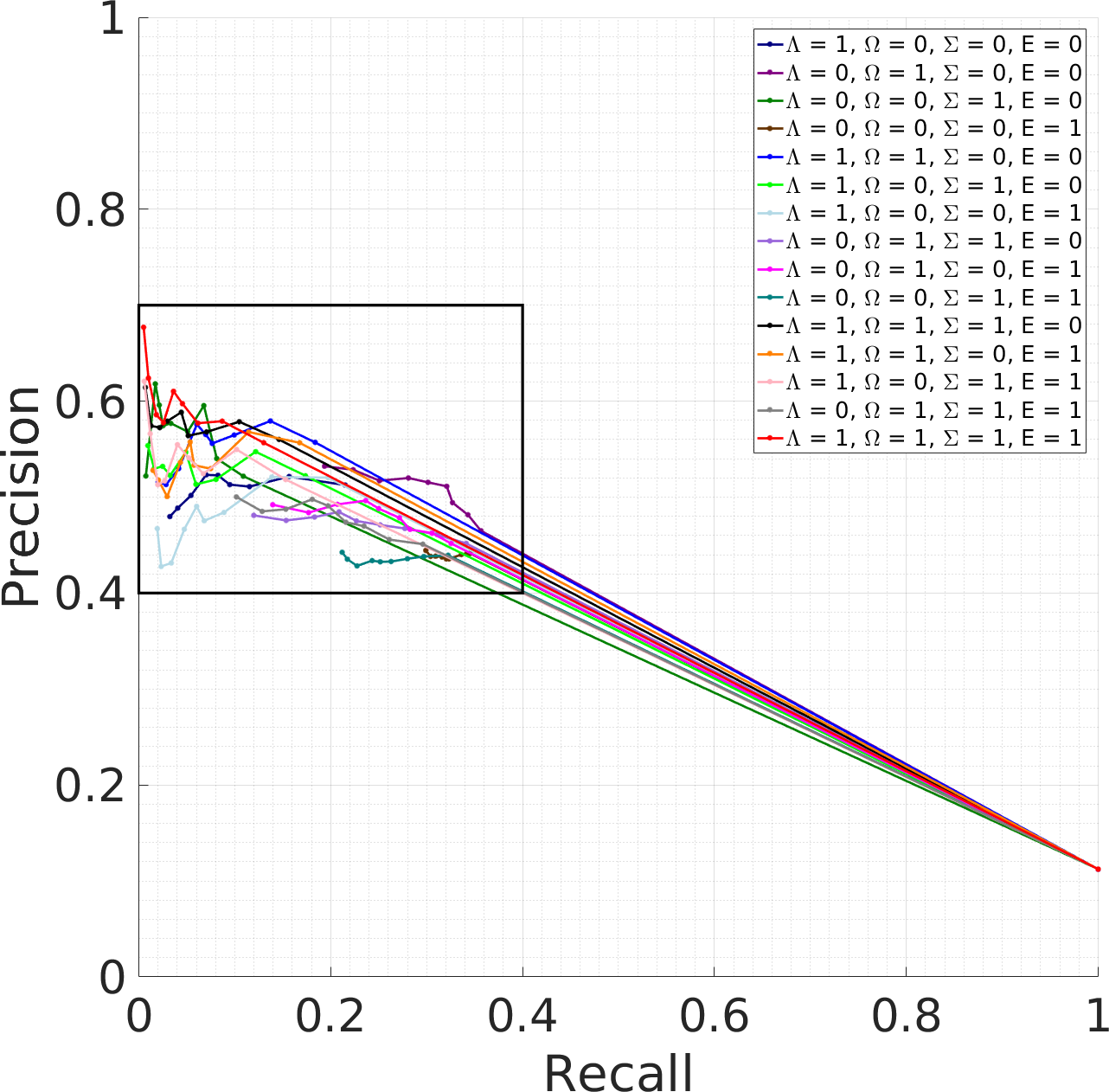} \\
		(a) PR curve in full scale. \vspace{3mm}\\
		\includegraphics[width=0.63\textwidth]{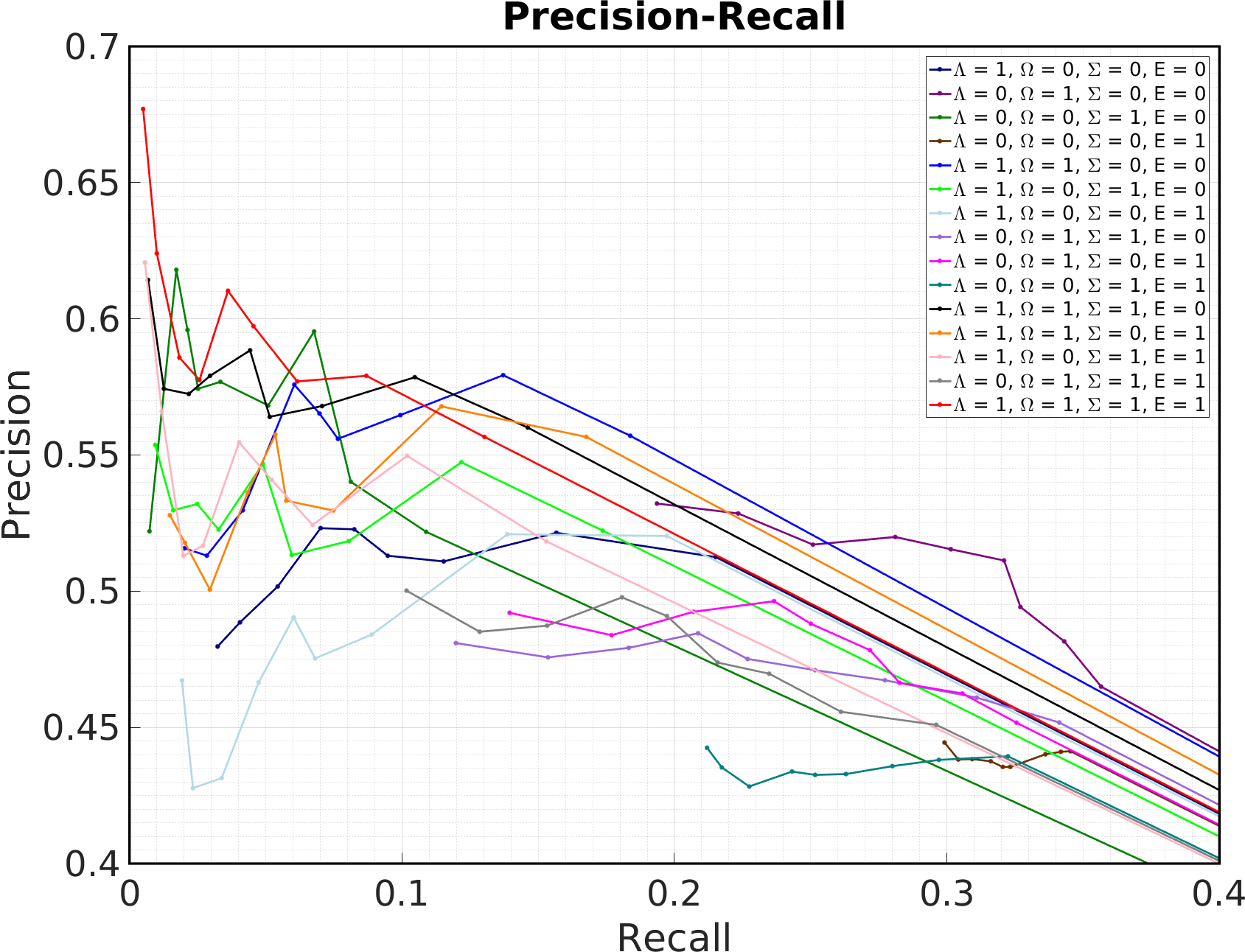}\\
		(b) PR curve for recall $\in [0,0.4]$ and precision $\in [0.4,0.7]$.\\
	\end{tabular}
	\end{center}
	\caption[PR curve for the proposed algorithm]{Precision-Recall (PR) curves of the proposed algorithm. `1' indicates enabled and `0' indicates disabled components of locomotion ($\Lambda$), oscillation ($\Omega$), asymmetry ($\Sigma$), and extrinsic motion dissimilarity ($E$). (a) a full PR curve is provided with both precision and recall in the ranges $[0,1]$ and (b) the same PR curve, as (a), restricted to recall values within $[0,0.4]$ and precision within $[0.4,0.7]$ as the majority of the points are focused in a select region of the PR curve. \label{fig:pr_ablation_full}}
\end{figure}
\restoregeometry
	
\begin{table}[!h]
	\resizebox{\linewidth}{!}{
	\begin{tabular}{| c | c | c | c || c | c | c | c | c | c | c | c | c | c | c |}
		\hline
		\multicolumn{4}{|c||}{\textbf{Components}} & \multicolumn{10}{c|}{$\tau_B$}\\
		\hline
		$\Lambda$ & $\Omega$ & $\Sigma$ & $E$ & 0 & 0.1 & 0.2 & 0.3 & 0.4 & 0.5 & 0.6 & 0.7 & 0.8 & 0.9 \\
		\hline
		1 & 0 & 0 & 0 & 0.1123 &0.5125 &0.5214 &0.5109 &0.5130 &0.5227 &0.5231 &0.5017 &0.4886 &0.4797 \\
		0 & 1 & 0 & 0 & 0.1121 &0.4650 &0.4816 &0.4942 &0.5113 &0.5154 &0.5199 &0.5171 &0.5285 &0.5322 \\
		0 & 0 & 1 & 0 & 0.1123 &0.5218 &0.5402 &0.5953 &0.5682 &0.5769 &0.5743 &0.5959 &0.6180 &0.5220 \\
		0 & 0 & 0 & 1 & 0.1123 &0.4412 &0.4410 &0.4401 &0.4355 &0.4355 &0.4376 &0.4384 &0.4382 &0.4444\\
		1 & 1 & 0 & 0 & 0.1123 &0.5571 &0.5793 &0.5647 &0.5559 &0.5652 &0.5758 &0.5297 &0.5131 &0.5157 \\
		1 & 0 & 1 & 0 & 0.1123 &0.5222 &0.5473 &0.5184 &0.5133 &0.5467 &0.5226 &0.5320 &0.5297 &0.5537 \\
		1 & 0 & 0 & 1 & 0.1123 &0.5203 &0.5209 &0.4840 &0.4753 &0.4904 &0.4665 &0.4314 &0.4277 &0.4672\\
		0 & 1 & 1 & 0 & 0.1123 &0.4518 &0.4609 &0.4673 &0.4710 &0.4751 &0.4846 &0.4792 &0.4757 &0.4810 \\
		0 & 1 & 0 & 1 & 0.1123 &0.4517 &0.4624 &0.4664 &0.4784 &0.4880 &0.4963 &0.4924 &0.4839 &0.4921\\
		0 & 0 & 1 & 1 & 0.1123 &0.4394 &0.4381 &0.4358 &0.4329 &0.4326 &0.4338 &0.4283 &0.4353 &0.4425\\
		1 & 1 & 1 & 0 & 0.1123 &0.5601 &0.5785 &0.5680 &0.5640 &0.5884 &0.5791 &0.5724 &0.5743 &0.6143 \\
		1 & 1 & 0 & 1 & 0.1123 &0.5566 &0.5678 &0.5296 &0.5332 &0.5575 &0.5362 &0.5006 &0.5177 &0.5279 \\
		1 & 0 & 1 & 1 & 0.1123 &0.5183 &0.5497 &0.5243 &0.5408 &0.5547 &0.5167 &0.5130 &0.5660 &0.6207 \\
		0 & 1 & 1 & 1 & 0.1123 &0.4509 &0.4557 &0.4697 &0.4738 &0.4909 &0.4977 &0.4874 &0.4851 &0.5002\\
		1 & 1 & 1 & 1 & 0.1123 &0.5566 &0.5791 &0.5770 &0.5973 &0.6103 &0.5775 &0.5858 &0.6240 &0.6770 \\
		\hline
	\end{tabular}
	}
	\caption[Precision values at specified threshold $\tau_B$]{Precision values at specified threshold $\tau_B$ for components: locomotion ($\Lambda$), oscillation ($\Omega$), asymmetry ($\Sigma$), and extrinsic motion dissimilarity ($E$). \label{tab:pr_ablation_full}}
\end{table}
	\end{section}
\end{appendices}

\newpage

\bibliographystyle{plain}
\bibliography{references}

\end{document}